\documentclass{article}

\usepackage{microtype}
\usepackage{graphicx}
\usepackage{booktabs} % for professional tables

\usepackage{amsmath,amssymb} % define this before the line numbering.
\usepackage{subfig}
\usepackage{tablefootnote}
\usepackage{wrapfig}
\usepackage{array}
\newcolumntype{L}[1]{>{\raggedright\let\newline\\\arraybackslash\hspace{0pt}}m{#1}}
\newcolumntype{C}[1]{>{\centering\let\newline\\\arraybackslash\hspace{0pt}}m{#1}}
\newcolumntype{R}[1]{>{\raggedleft\let\newline\\\arraybackslash\hspace{0pt}}m{#1}}
\usepackage{color}

\usepackage{hyperref}

\usepackage[accepted]{sysml2019}

\sysmltitlerunning{AdaScale: Towards Real-time Video Object Detection Using Adaptive Scaling}

\begin{document}

\twocolumn[
\sysmltitle{AdaScale: Towards Real-time Video Object Detection Using Adaptive Scaling}

\sysmlsetsymbol{equal}{*}

\begin{sysmlauthorlist}
\sysmlauthor{Ting-Wu Chin}{equal,cmu}
\sysmlauthor{Ruizhou Ding}{equal,cmu}
\sysmlauthor{Diana Marculescu}{cmu}
\end{sysmlauthorlist}

\sysmlaffiliation{cmu}{Department of ECE, Carnegie Mellon University, Pittsburgh}

\sysmlcorrespondingauthor{Ting-Wu Chin}{tingwuc@cmu.edu}
\sysmlcorrespondingauthor{Ruizhou Ding}{rding@cmu.edu}

\sysmlkeywords{Object Detection Acceleration, Convolution Neural Networks}

\vskip 0.3in

\begin{abstract}
  In vision-enabled autonomous systems such as robots and autonomous cars, video object detection plays a crucial role, and both its speed and accuracy are important factors to provide reliable operation. The key insight we show in this paper is that speed and accuracy are not necessarily a trade-off when it comes to image scaling. Our results show that re-scaling the image to a lower resolution will sometimes produce better accuracy. 
  Based on this observation, we propose a novel approach, dubbed AdaScale, which adaptively selects the input image scale that improves both accuracy and speed for video object detection. To this end, our results on ImageNet VID and mini YouTube-BoundingBoxes datasets demonstrate \textit{1.3 points} and \textit{2.7 points} mAP improvement with \textit{1.6$\times$ and 1.8$\times$ speedup}, respectively. Additionally,  we improve state-of-the-art video acceleration work by an extra \textit{1.25$\times$ speedup} with slightly better mAP on ImageNet VID dataset.
\end{abstract}
]

\printAffiliationsAndNotice{\sysmlEqualContribution}

\section{Introduction}

Video object detection acts as a fundamental building block for visual cognition in future autonomous agents such as autonomous cars, drones, and robots. Therefore, to build systems with reliable performance, it is critical for the detectors to be fast and accurate. Though object detection is well-studied for static images~\cite{dai16rfcn,girshick2015fast,he2014spatial,liu2016ssd,renNIPS15fasterrcnn}, there are unique challenges in the case of \textit{video} object detection, including motion blur caused by the moving objects, failure of camera focus~\cite{zhu2017flow}, and also real-time speed constraints when it comes to autonomous agents. Besides these \textit{challenges}, however, video object detection also brings new \textit{opportunities} to be exploited. Some of the prior work that focuses on video object detection tries to improve average precision by leveraging a unique characteristic of video~\cite{zhu2017flow,feichtenhofer2017detect,kang2017t}, which is the temporal consistency (consecutive frames have similar content). On the other hand, from a speed perspective, prior work~\cite{zhu2017deep,zhu2018, buckler2018} counts on the temporal consistency to reduce the computation needed for a standalone object detector. Similarly, we aim to leverage the temporal consistency, but to improve both speed and accuracy of the standalone object detectors with a novel technique called adaptive-scale testing, or \textit{AdaScale}. 

The scale of input image affects both the speed and accuracy of modern CNN-based object detectors~\cite{huang2017speed}. Prior work related to image scaling addresses two directions: (i) multi-scale testing for better accuracy, and (ii) down-sampling images for higher speed. Examples from the first category include re-sizing images to various scales (image pyramid) and pushing them through the CNN for feature extraction at various scales~\cite{dai16rfcn,girshick2015fast,he2014spatial}, as well as fusing feature maps from different layers generated by a single-scale input image~\cite{lin2017feature,cai2016unified,bell2016inside}. However, these approaches introduce extra computational overhead compared to object detectors with single-scale inputs. Examples from the second category include Pareto optimal search by tuning the input image scale~\cite{lin2017focal,liu2016ssd,redmon2016yolo9000,huang2017speed} and dynamically re-sizing the image according to the input image~\cite{chin18approx}. However, results for these approaches demonstrate that higher speed comes at the cost of lower accuracy when it comes to image scaling.

\begin{figure*}[t]
    \subfloat[][600]{\begin{tabular}[b]{c}
    \includegraphics[width=.22\linewidth]{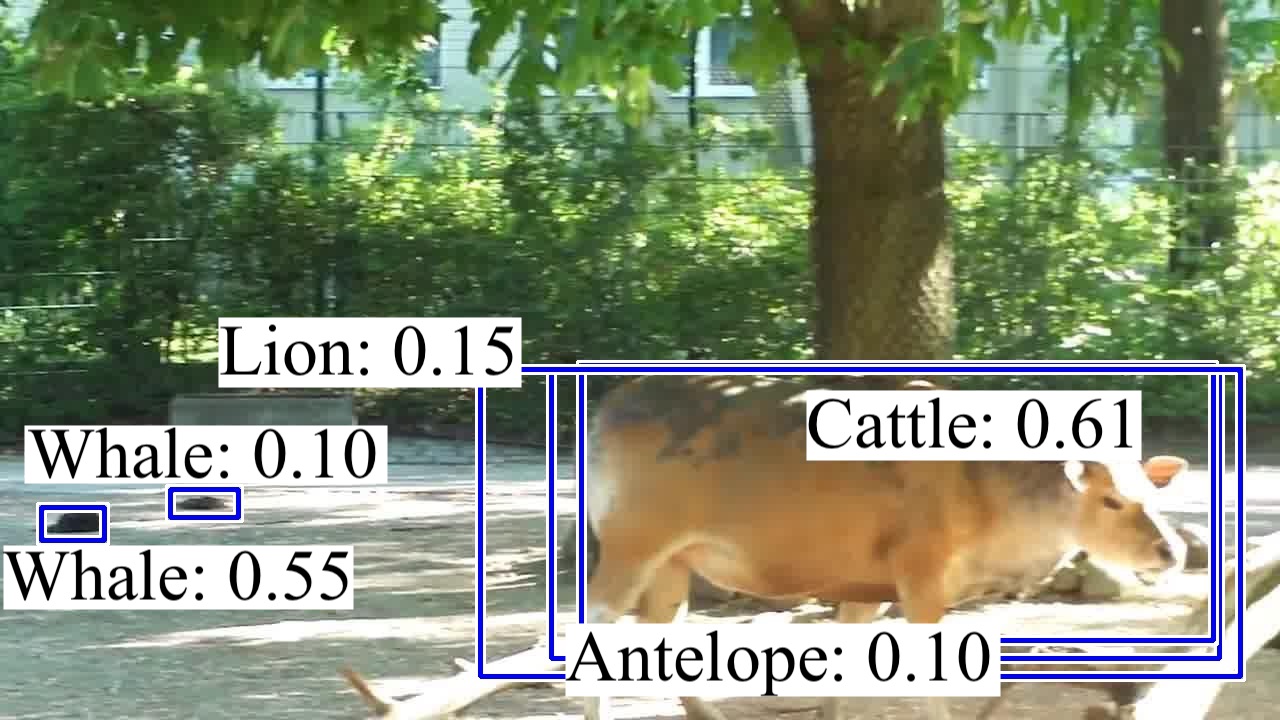} \\ \includegraphics[width=.22\linewidth]{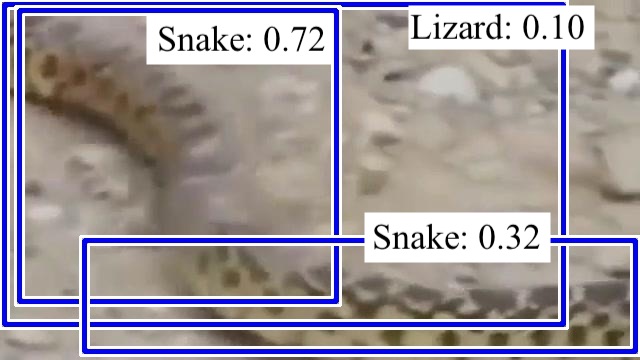}\end{tabular}\label{fig:example:1}}
    \subfloat[][240]{\begin{tabular}[b]{c}
    \includegraphics[width=.22\linewidth]{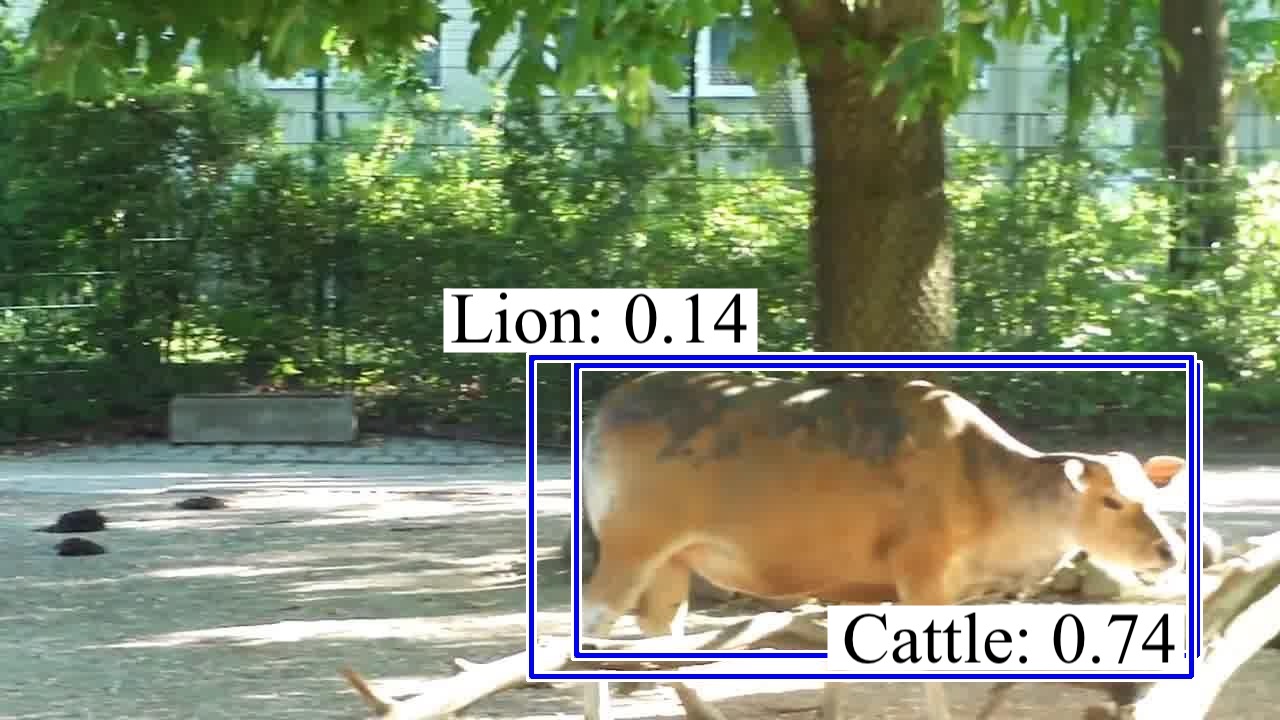} \\ \includegraphics[width=.22\linewidth]{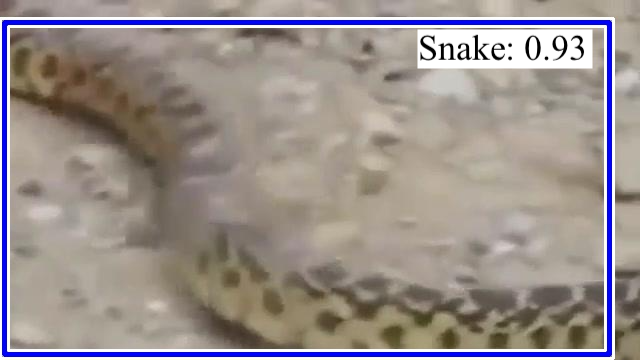}\end{tabular}\label{fig:example:2}}
    \subfloat[][600]{\begin{tabular}[b]{c}
    \includegraphics[width=.22\linewidth]{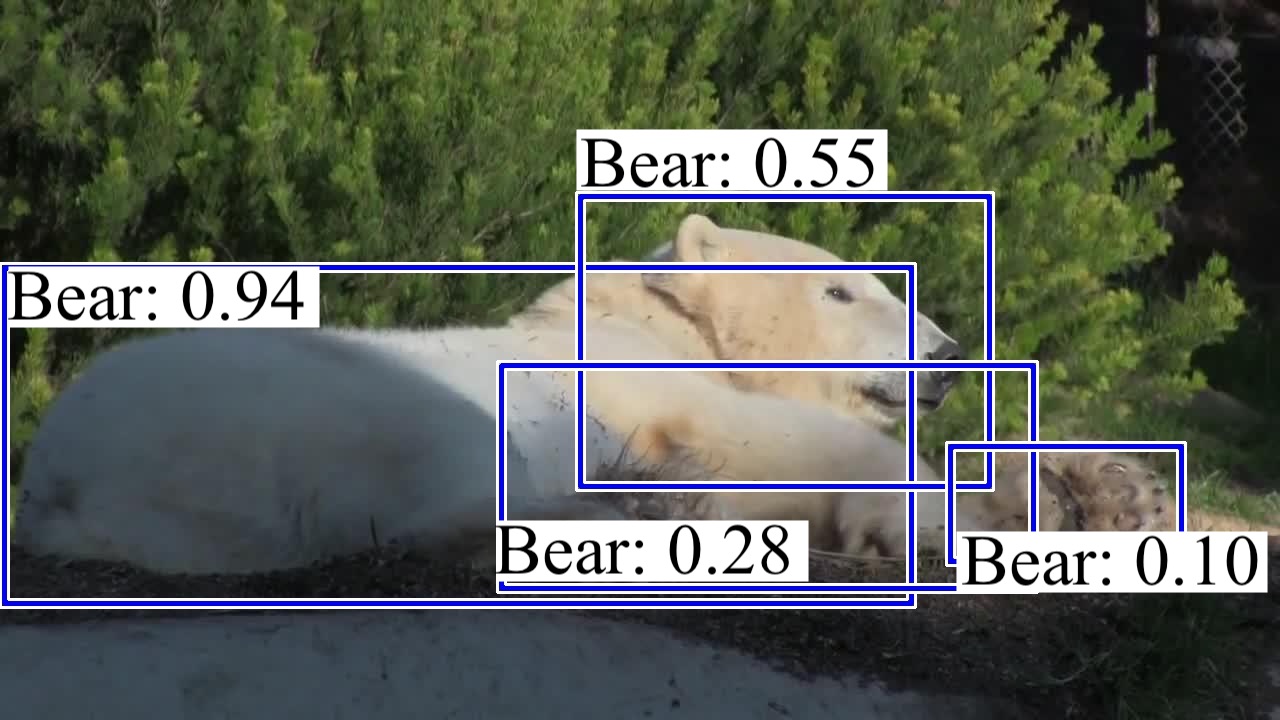} \\ \includegraphics[width=.22\linewidth]{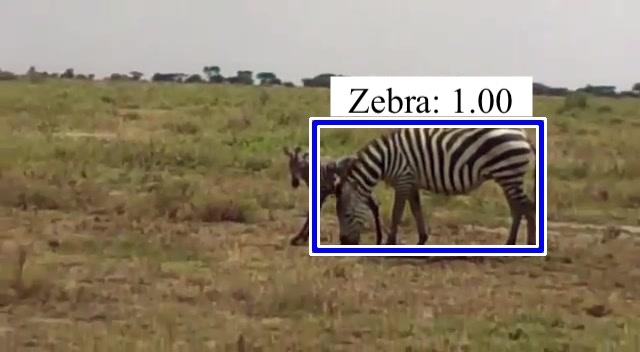}\end{tabular}\label{fig:example:3}}
    \subfloat[][480]{\begin{tabular}[b]{c}
    \includegraphics[width=.22\linewidth]{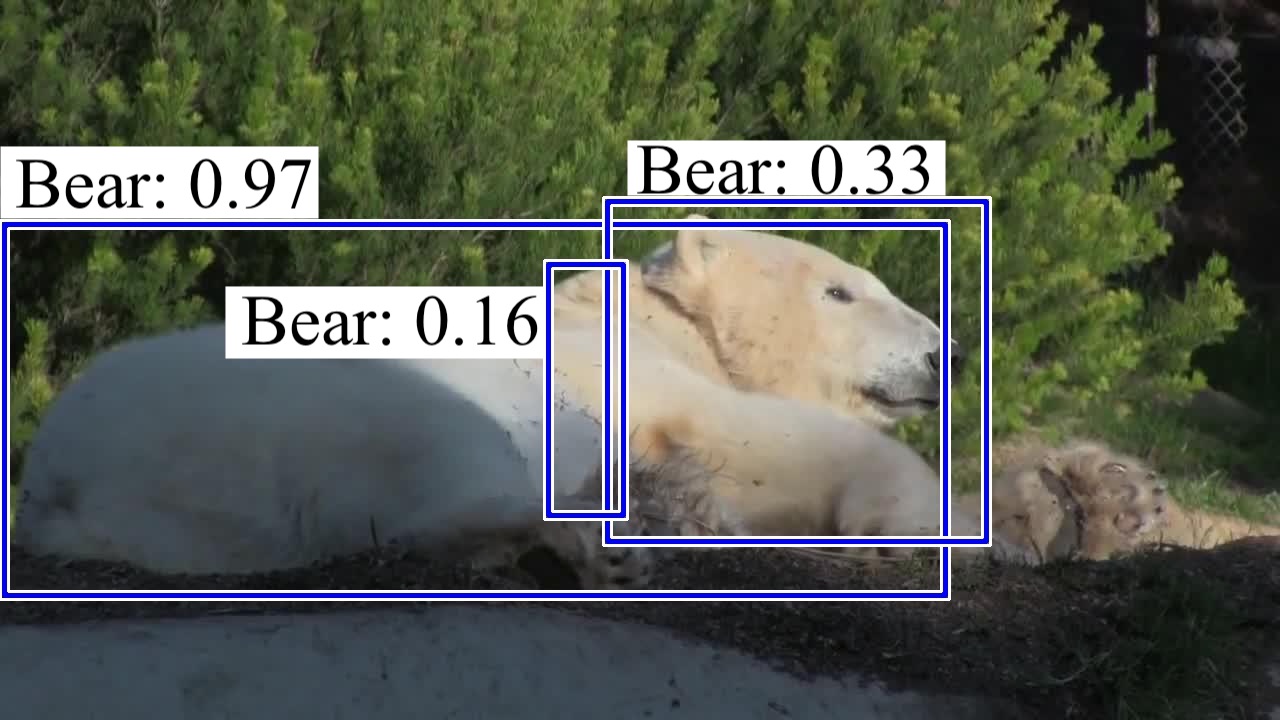} \\ \includegraphics[width=.22\linewidth]{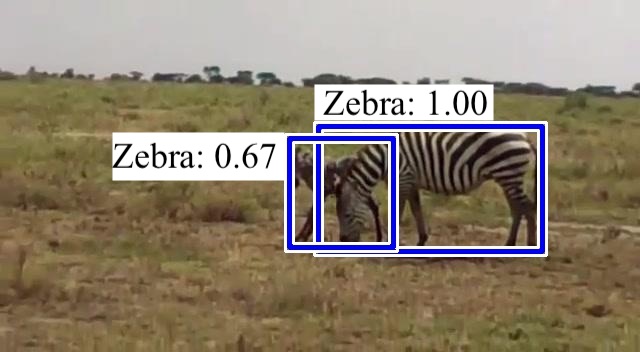}\end{tabular}\label{fig:example:4}}
    \caption{Examples where down-sampled images have better detection results. Blue boxes are the detection results, and the numbers are the confidence. The detector is trained on a single scale (pixels of the shortest side) of 600. Column (a) and (c) are tested at scale 600. Column (b) is tested at scale 240 and column (d) is tested at scale 480.}
    \vspace{-6pt}
    \label{fig:examples}
\end{figure*}

In contrast with prior work, we find that down-sampling images is sometimes beneficial in terms of accuracy. Specifically, there are two sources of improvement brought by image down-sampling: (i) Reducing the number of false positives that may be introduced by focusing on unnecessary details. (ii) Increasing the number of true positives by scaling the objects that are too large to a size at which the object detector is more confident. Fig.~\ref{fig:examples} shows images that are better when down-sampled in our experiments using Region-based Fully Convolutional Network (R-FCN)~\cite{dai16rfcn} object detector on ImageNet VID dataset.

Motivated by this, our goal is to re-size the images to their ``best" scale aiming for both higher speed and accuracy. In this work, we propose \textit{AdaScale} to boost both the accuracy and the speed of the standalone object detector. Specifically, we use the current frame to predict the optimal scale for the next frame. Our results on ImageNet VID and mini YouTube-BB datasets demonstrate 1.3 points and 2.7 points mAP improvement with 1.6$\times$ and 1.8$\times$ speedup, respectively. Moreover, by combining with the state-of-the-art video acceleration work~\cite{zhu2017deep}, we improve its speed by an an extra 25\% with a slight mAP increase on ImageNet VID dataset.

\label{sec:intro}

\section{Related Work}
\label{sec:related}

Our work focuses on applying adaptive scaling to video object detection in order to improve both speed and accuracy of object detectors. In the sequel we discuss prior work in scale-related object detection and video object detection.

\subsection{Image Scale for Object Detection}

We discuss two categories of scale-related object detection work: (i) single-shot detection by exploiting feature maps from various layers of the CNN with inherently different scales, and (ii) multi-shot detection with input images at multiple scales.

\textit{Single-Shot:} In this category, object detectors are designed to take an input image once and detect objects at various scales. That is, this category of prior work treats deep CNNs as scale-invariant. Prior work~\cite{bell2016inside} uses features from different layers in the CNN and merge them with normalization and scaling. Similar idea is also adopted by other work~\cite{liu2016ssd,cai2016unified,lin2017feature,zhou2017cad}. From a different viewpoint, prior art~\cite{liu2017recurrent} proposes to use a recurrent network to approximate feature maps produced by images at different scales. Though single-shot approaches have shown great promise in better detecting various scales, the scale-invariant design philosophy generally requires a large model capacity~\cite{kanazawa2014locally,liu2017recurrent}. We note that, without perfect scale-invariance, different image scales will result in different accuracy, and prior art often uses a fixed single scale, \emph{e.g.} 600 pixels on the smallest side of the image. Hence, this line of work could be further improved in terms of speed and accuracy when augmented to adaptive scaling.

\textit{Multi-shot:} This refers to scaling a single input image to various scales, forwarding each scaled image through the object detector, and merging the obtained results. Some work~\cite{he2014spatial,girshick2015fast,ren2017object,he2016deep,dai16rfcn} forwards multiple scales of images to obtain feature maps with various scales. More recently, prior work~\cite{dai2017deformable} leverages multiple scales of images to infer multiple detection results and merge them using Non-Maximum Suppression. While multi-shot object detection alleviates the problem of imperfect scale-invariance, it incurs significant extra computation overhead, \textit{i.e.}, up to 4$\times$~\cite{girshick2015fast}.

Our work is aiming to alleviate the imperfect scale-invariance by selecting the best scale for each image, and hence, improves the accuracy compared to single-shot methods. Moreover, to improve the speed in the meantime, we consider down-sampling rather than up sampling. We note that our method could possibly be extended to multi-shot version, \emph{i.e.}, adaptively select multiple scales for a given image, and we leave it for the future work.

\subsection{Video Object Detection}

We discuss prior work that aims at improving speed and/or accuracy of video object detection.

\textit{Speed:} Optical flow was proposed to reduce detection overhead by prior work~\cite{zhu2017deep}. Similar to our idea, some prior art~\cite{chin18approx} proposes to adaptively scale the image to improve the detection speed. However, both works improve speed at the expense of accuracy loss.
\textit{Accuracy:} Prior work~\cite{kang2017t} proposes to leverage contextual and temporal information across the video while some work \cite{zhu2017flow} uses the idea from Deep Feature Flow (DFF)~\cite{zhu2017deep} to incorporate temporal information across consecutive frames. Another study~\cite{feichtenhofer2017detect} proposes to integrate detection with tracking into an end-to-end trainable deep CNN.
\textit{Both}: Some prior work~\cite{zhu2017towards} extends \cite{zhu2017flow} and \cite{zhu2017deep} to use both feature aggregation and propagation. Additionally, they propose to regress a quality metric of the optical flow to decide when and how to propagate the features.

Compared to the aforementioned related work, the main contribution that sets our work apart is that our work focuses on fixing the problem where existing object detectors use fixed single scale for each of the image while the object detectors are not scale-invariant, which is different from most of the prior work that focuses on exploiting the relationship of the detection results among the neighboring frames~\cite{kang2017t,zhu2017deep,zhu2017flow,feichtenhofer2017detect,zhu2017towards}. Moreover, we show that we are complementary to state-of-the-art video object detection acceleration technique~\cite{zhu2017deep}.

\section{Adaptive Scaling}
\label{sec:adascale}

\begin{figure}[h]
\centering
\includegraphics[width=1\linewidth]{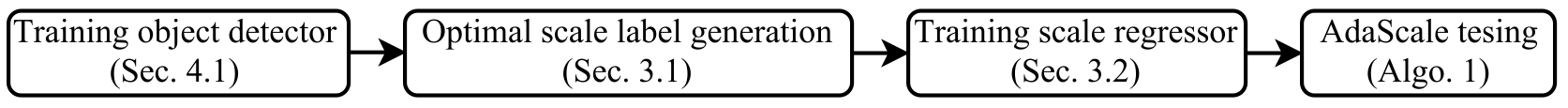}
\caption{The AdaScale methodology.}
\vspace{-6pt}
\label{fig:guideline}
\end{figure}

Fig.~\ref{fig:guideline} provides an overview for AdaScale methodology. It includes fine-tuning the object detector, using the resulting detector to generate the optimal scale labels, training the scale regressor with the generated labels, and the deployment of AdaScale in video object detection. We discuss each component in detail in the following sections.
\vspace{-5pt}

\vspace{-12pt}

\subsection{Optimal Scale}
\label{sec:metric}
To define the optimal scale (pixels of the shortest side) of a given image, we need to first define a finite set of scales $S$ (\textit{e.g.}, in our case $S = \{600, 480, 360, 240\}$) and we must have a metric that evaluates the quality of the detection results at these different scales. Naively, we can use the commonly used mean average precision (mAP) to compare different scales, and define the scale with the largest mAP as the optimal scale. However, the mAP evaluated for a single image is sparse due to limited number of ground truths per image. Hence, we opt to count on the loss function that is used to train the object detector as the metric to compare results at different scales. In general, the loss function for an object detector used in training often includes the bounding box regression loss and classification loss~\cite{girshick2015fast,renNIPS15fasterrcnn,dai16rfcn}:
\begin{align}
    L(\bold{p}, u,\bold{t}, \bold{\hat{t}}) = L_{cls}(\bold{p}, u) + \lambda[u\ge 1]L_{reg}(\bold{t}, \bold{\hat{t}}), \label{eq:1}
\end{align}
where $\bold{p}$ is a vector of predicted probability for each pre-defined class, $u$ is the ground truth class label (0 means background), $\bold{\hat{t}}$ is a four-dimension vector that indicates the location information of the bounding box~\cite{girshick2015fast}, and $\bold{t}$ is also a four-dimension vector that represents the ground truth location of the bounding box. Noted that $[u\ge 1]$ indicates that regression loss only applies to the bounding box whose ground truth label is not background. Generally~\cite{dai16rfcn,lin2017focal}, a predicted bounding box is assigned to foreground when there is at least one ground truth bounding box that has over 0.5 Jaccard overlap (intersection over union)~\cite{erhan2014scalable} with it; otherwise, it is assigned to background.
However, since this loss function naturally assumes that the regression loss for background is 0, directly using it to \textit{assess} different image scales will favor the image scale with fewer foreground bounding boxes.

\begin{figure*}[t]
\centering
\begin{minipage}{.65\textwidth}
  \centering
  \includegraphics[width=1\linewidth]{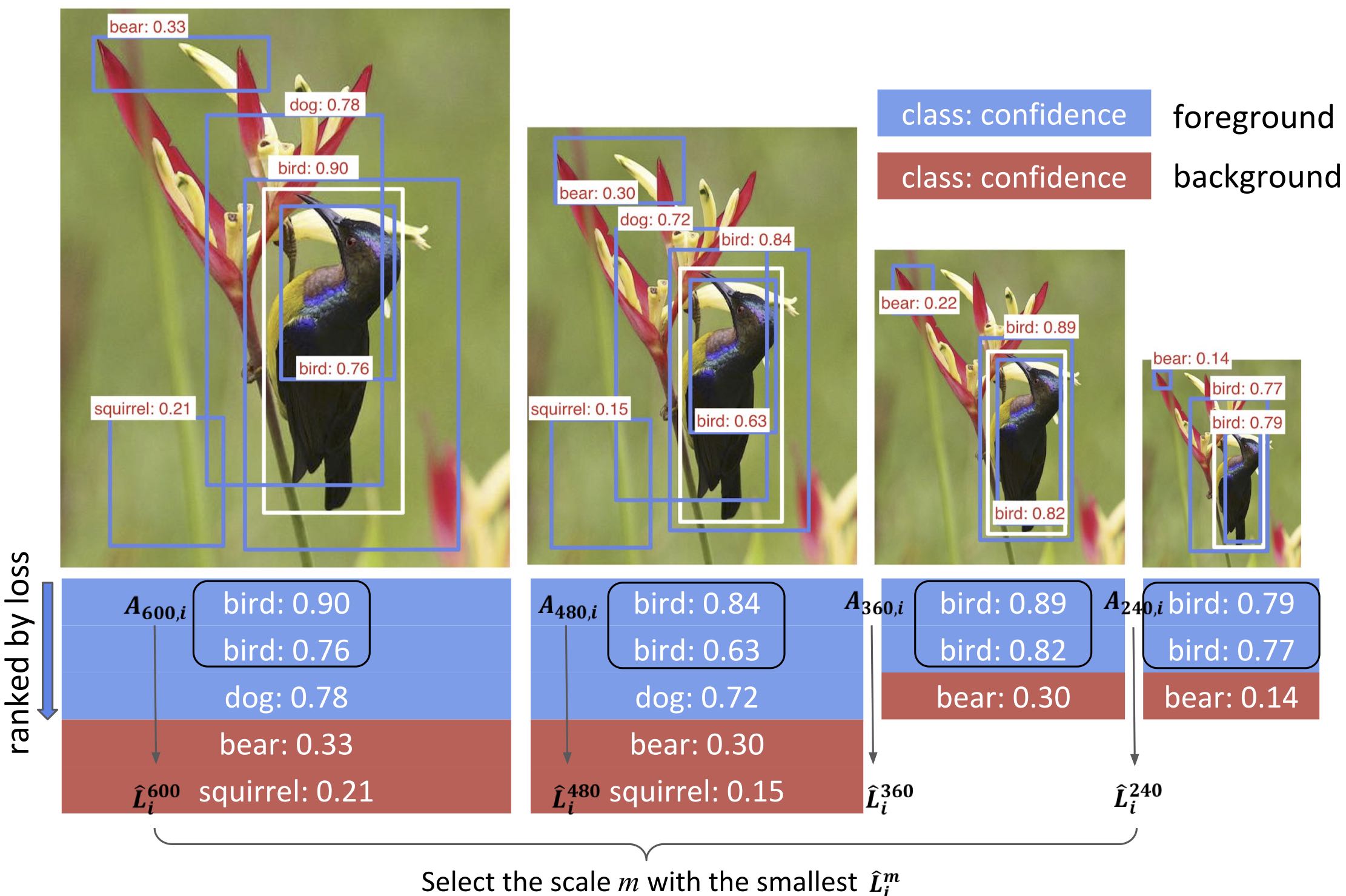}
  \caption{Optimal scale determination. First, the same number of predicted foregrounds from four scales are selected as $A_{m,i}$. Then, the scale with the lowest loss $\hat{L^{m}_{i}}$ is selected as the optimal scale.}
  \label{fig:loss}
\end{minipage}\hfill%
\begin{minipage}{.3\textwidth}
  \centering
  \includegraphics[width=1\linewidth]{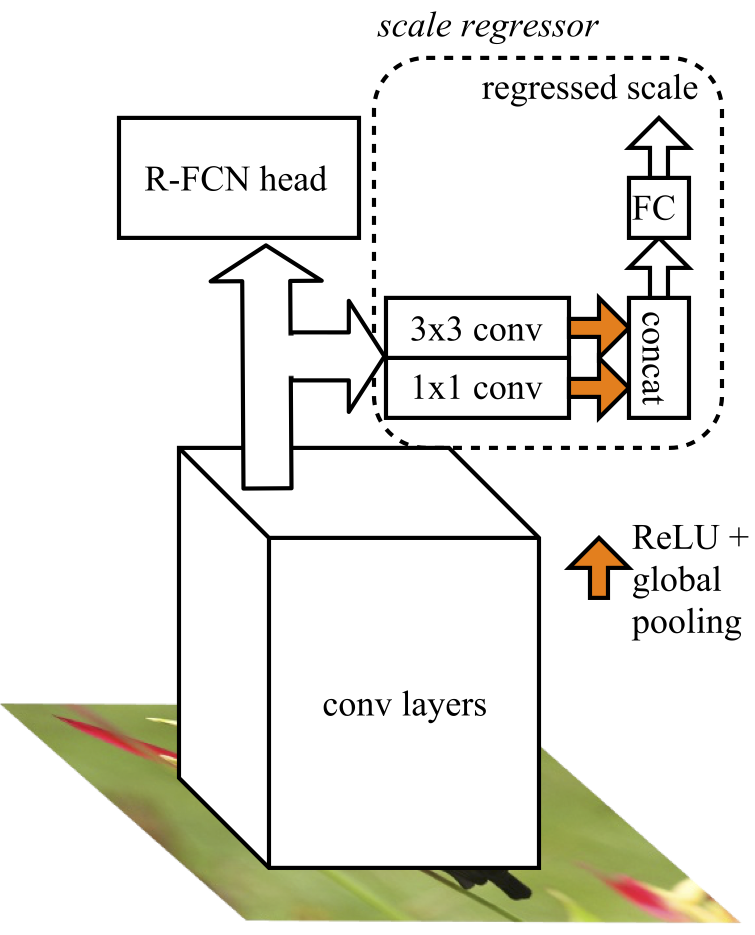}
  \caption{The scale regressor module.}
  \label{fig:regressor}
\end{minipage}
\end{figure*}

Hence, to deal with this, we devise a new metric that focuses only on the same number of foreground bounding boxes to compare different image scales. To explain our proposed metric, we denote $L^{m}_{i,a}, m\in S$ as the loss of predicted bounding box $a$ of image $i$ at scale $m$ using (\ref{eq:1}), and denote $\hat{L^{m}_{i}}, m\in S$ for image $i$ at scale $m$, as our proposed metric. To obtain $\hat{L^{m}_{i}}$, we first compute the number of predicted foreground bounding boxes, $n_{m,i}$, for image $i$ at each scale $m \in S$, then let $n_{min,i} = \min_{m}(n_{m,i})$. Concretely, the proposed metric can be computed as: $\hat{L^{m}_{i}} = \sum_{a\in A_{m,i}}L^{m}_{i,a}$,
where $A_{m,i}$ is a set of predicted foreground bounding boxes of image $i$ at scale $m$ and $|A_{m,i}| = n_{min,i}$. To obtain $A_{m,i}$, for each scale, we sort the predicted foreground bounding boxes of image $i$ with respect to $L^{m}_{i,a}$ in ascending order and pick the first $n_{min,i}$ predictions into the set $A_{m,i}$. The visual illustration of the process is shown in Fig.~\ref{fig:loss}. With the proposed metric, we define optimal scale $m_{opt, i}$ for image $i$ as:
\begin{align}
    m_{opt, i} = argmin_{m} \hat{L^{m}_{i}}.
    \vspace{-6pt}
    \label{eq:3}
\end{align}

\subsection{Scale Regressor}
\label{sec:regressor}
Now that we understand which scale is better for a given image, we may be able to predict the optimal scale for the image. Intuitively, if the object is large or has simple texture, it is likely that we would down-sample the image to let the object detector focus on the salient objects rather than the distracting details. On the other hand, if the object is small or there are many salient objects, the image should remain in a large scale. Since R-FCN head~\cite{dai16rfcn} counts on the deep features (\textit{i.e.}, the last convolutional layer of the backbone feature extractor) to regress bounding box locations, we think that the channels of the deep features already contain size information. As a result, we build a scale regressor using deep features to predict the optimal scale, as shown in Fig.~\ref{fig:regressor}. Specifically, we use a 1x1 convolutional layer to capture the size information from different feature maps. Additionally, we use a parallel 3x3 convolutional layer to capture the complexity of each 3x3 patch in the feature maps.  After the non-linear unit, we use global pooling that acts as a voting process. Lastly, we combine the two streams with a fully connected layer to regress the output scale. To be precise, we define the deep features as $X \in R^{C\times H\times W}$, where $C$ is number of channels, $H$ and $W$ are height and width of the deep feature maps. We define our regressor as $g: R^{C\times H\times W} \to R$. It is important to note that we do not regress the optimal scale $m_{opt}$ directly since what matters is the content instead of the image size itself. Hence, we regress a relative scale so that the module learns to react (up-sample, down-sample, or stay the same) given the current content of the image. Specifically, the target of the regressed scale for image $i$ is defined as:
\begin{align}
    t(m_{i}, m_{opt, i}) =2\times \frac{m_{opt, i}/m_{i} - m_{min}/m_{max}}{m_{max}/m_{min} - m_{min}/m_{max}} - 1,
    \label{eq:4}
\end{align}
where $m_i$ is the scale of the image $i$, $m_{min}$ is the minimum defined scale, \textit{e.g.}, 128, while $m_{max}$ is the maximum defined scale, \textit{e.g.}, 600. That is, we are regressing to normalized, \textit{i.e.}, [-1, 1], relative scales. To generate labels for the regressor, we calculate (\ref{eq:3}) over the training data to obtain $m_{opt, i}\ \ \forall i \in D_{train}$, where $D_{train}$ is the training data. As commonly used in regression problems, we adopt mean square error (\ref{eq:5}) as the loss function to train the regressor:
\begin{align}
    L_{scalereg} = \frac{1}{|D_{train}|} \sum_{i\in D_{train}} (g(X_{i}) - t(m_{i}, m_{opt, i}))^2.
    \label{eq:5}
\end{align}

To incorporate adaptive scaling, or AdaScale, in the video setting, we impose a temporal consistency assumption. More precisely, we assume that the optimal scales for the two consecutive frames are similar; our results empirically justify this assumption. Algorithm~\ref{algo:adascale} shows an example of leveraging AdaScale for video object detection, which is elaborated in section~\ref{sec:train}.
\vspace{-6pt}
\begin{algorithm}
  \caption{Pseudo-code for using AdaScale in the testing phase.}\label{algo:adascale}
  \begin{algorithmic}
  \STATE {\bfseries Input:} {detector, video, $S$: pre-defined scale set}
  \STATE image = video.next\_frame();
  \STATE targetScale = 600; $//$ \textit{Initialize image scale}
  \WHILE {image}
    \STATE image = resize(image, targetScale); \\
    \STATE base\_size = minimum(image.height, image.width); \\
    \STATE $//$ \textit{Regress $t$ of Eq.~(\ref{eq:4})} \\
    \STATE bboxes, scores, targetScale = detector.detect(image); \\
    \STATE $//$ \textit{Invert Eq.~(\ref{eq:4})} \\
    \STATE targetScale = decode(targetScale, base\_size, $S$); \\
    \STATE targetScale.clip\_(min($S$), max($S$)).round\_(); \\
    \STATE image = video.next\_frame();
  \ENDWHILE
  \end{algorithmic}
\end{algorithm}

\section{Experiments}
\label{sec:exp}
\subsection{Setup}
All of our experiments are done using Nvidia GTX 1080 Ti. We base our implementation on the code released by prior work~\cite{zhu2017deep}, where MXNet~\cite{chen2015mxnet} is used as the deep learning framework. We conduct our experiments mainly on the ImageNet VID dataset~\cite{russakovsky2015imagenet}, which contains 3862 and 555 training and validation video snippets, respectively. We use a pre-trained R-FCN model~\cite{zhu2017deep}, which is trained on both ImageNet DET and ImageNet VID training set. For DET dataset, only the 30 categories that overlap with the VID dataset are selected for training. The evaluation of ImageNet VID is performed on validation set, which follows prior work~\cite{zhu2017deep}. In addition to ImageNet VID, we also evaluate our performance on the recently released YouTube-BB dataset~\cite{real2017youtube}, which contains 23 categories and around 380,000 video segments. Due to resource and time limitation, we randomly sample 100 segments per category and cut 20 frames per segment to form our mini training set. We also sample 10 segments per category for the validation set to form our mini testing set. To train the model for mini Youtube-BB dataset, we use the model trained on ImageNet VID and DET as a pre-trained model to further fine-tune on mini Youtube-BB.

\subsection{Training and Testing}
\label{sec:train}
\textit{Object Detector:}
First, to avoid the object detector to be biased toward a single scale, we fine-tune the R-FCN model pre-trained at scale 600, for four epochs using multi-scale training~\cite{girshick2015fast}. The hyperparameters used follow prior work~\cite{zhu2017deep}. Specifically, we use a learning rate of 0.00025 and divide it by 10 after 1.3 and 2.6 epochs, respectively. We use two GPUs with a single image per GPU. Therefore, the training batch size is two. In a addition, we pick the scale (the shortest side of the image) from the set $S_{train} = \{600, 480, 360, 240\}$, and use the maximum bound for the longer side as 2000. Our re-sizing protocol follows Fast R-CNN~\cite{girshick2015fast}. In the following sections, we will refer to the shortest side size as the \textit{image scale}. All the detection results in this work use Non-Maximum Suppression (NMS) with threshold 0.3~\cite{dai16rfcn}. For each image, the top-300 confident bounding boxes after NMS are selected as the final output.

\textit{Scale Regressor:}
With the multi-scale trained object detector, we generate the scale label for each frame in the training data with a set of pre-defined scales using the proposed metric in section~\ref{sec:metric}. To enable adaptive scaling, the regressor needs to learn to scale up or down according to the current content. To best train the regressor, we should scale the image to every possible scales for the regressor to learn the dynamics. That is, when training the regressor, the input image scale is randomly drawn from a uniform distribution of the pre-defined scale set $S_{reg}$. In practice, we find $S_{reg} = \{600, 480, 360, 240, 128\}$ is enough to cover the dynamics between 600 and 128. Note that we pick 128 since it is the scale of smallest pre-defined bounding box or anchor used in the Region Proposal Network~\cite{renNIPS15fasterrcnn} inside R-FCN and we want to push the image to an as small as possible scale for the largest potential speed improvement. With the generated label, we then train our scale regressor using the training data and freeze the weights of the entire network, except for the scale regressor module. We train the scale regressor for two epochs with an initial learning rate of $10^{-4}$ and divide by 10 after 1.3 epoch. For the testing phase, as shown in Algorithm~\ref{algo:adascale}, we begin every video snippet by re-sizing the first frame to 600. Then, we use the decoded regressed scale for the next frame. As for decoding the regressed scale, we first count on the inverse of (\ref{eq:4}) to obtain a scale in floating point. Then, we round it to an integer, and clip it to the range $[S_{min}, S_{max}]$.

\subsection{Evaluation}
\label{sec:eval}
\begin{table*}[t]
\centering
\caption{Evaluation of the proposed method. We denote methods by their approach of training and testing, \textit{e.g.}, MS/SS stands for multi-scale (MS) training and single-scale (SS) testing. Blue text and red text indicate $\ge$ 1 AP improvement and degradation compared to SS/SS, respectively.}
\label{table:eval}
\vspace{5pt}
\subfloat[ImageNet VID]{%
\resizebox{\textwidth}{!}{%
\begin{tabular}{c|C{0.8cm}C{0.8cm}C{0.8cm}C{0.8cm}C{0.8cm}C{0.8cm}C{0.8cm}C{0.8cm}C{0.8cm}C{0.8cm}C{0.8cm}C{0.8cm}C{0.8cm}C{0.8cm}C{0.8cm}}
Method &\rotatebox{60}{airplane}& \rotatebox{60}{antelope} & \rotatebox{60}{bear} & \rotatebox{60}{bike} & \rotatebox{60}{bird} & \rotatebox{60}{bus} & \rotatebox{60}{car} & \rotatebox{60}{cattle} & \rotatebox{60}{dog} & \rotatebox{60}{cat} & \rotatebox{60}{elephant} & \rotatebox{60}{fox} & \rotatebox{60}{g.panda} & \rotatebox{60}{hamster} & \rotatebox{60}{horse} \\
\hline
SS/SS & 88.9 & 84.5 & 86.0 & 65.8 & 72.2 & 76.1 & 58.3 & 71.0 & 69.4 & 76.0 & 76.4 & 87.2 & 81.6 & 89.8 & 69.6\\
\hline
MS/SS & 88.5 & 86.2 & 75.3 & 62.8 & 74.2 & 74.5 & 55.4 & 71.2 & 72.1 & 75.8 & 77.1 & 87.5 & 82.1 & 87.5 & 74.4\\
\hline
~~~MS/AdaScale~~~ & 88.2 & \textcolor{blue}{87.0} & \textcolor{red}{80.2} & \textcolor{blue}{67.4} & \textcolor{blue}{73.7} & 75.3 & 57.8 & \textcolor{blue}{73.4} & \textcolor{blue}{74.1} & \textcolor{blue}{81.7} & \textcolor{blue}{77.7} & \textcolor{blue}{89.1} & 81.5 & \textcolor{blue}{93.5} & \textcolor{blue}{75.6}\\
\hline

\end{tabular}}}

\resizebox{\textwidth}{!}{%
\begin{tabular}{C{0.8cm}C{0.8cm}C{0.8cm}C{0.8cm}C{0.8cm}C{0.8cm}C{0.8cm}C{0.8cm}C{0.8cm}C{0.8cm}C{0.8cm}C{0.8cm}C{0.8cm}C{0.8cm}C{0.8cm}|c|c}
\rotatebox{60}{lion} & \rotatebox{60}{lizard} & \rotatebox{60}{monkey} & \rotatebox{60}{motorcycle} & \rotatebox{60}{rabbit} & \rotatebox{60}{r.panda} & \rotatebox{60}{sheep} & \rotatebox{60}{snake} & \rotatebox{60}{squirrel} & \rotatebox{60}{tiger} & \rotatebox{60}{train} & \rotatebox{60}{turtle} & \rotatebox{60}{watercraft} & \rotatebox{60}{whale} & \rotatebox{60}{zebra} & \rotatebox{0}{mAP(\%)} & \rotatebox{0}{Runtime(ms)} \\
\hline
51.9 & 79.1 & 51.2 & 84.0 & 63.4 & 76.8 & 56.3 & 75.6 & 53.9 & 89.5 & 82.4 & 79.0 & 65.1 & 74.5 & 91.3 & 74.2 & 75\\
\hline
57.1 & 78.7 & 51.2 & 83.8 & 61.0 & 58.7 & 61.5 & 68.9 & 57.3 & 89.8 & 81.4 & 78.2 & 64.5 & 74.3 & 89.2 & 73.3 & 75\\
\hline
\textcolor{blue}{62.6} & 78.7 & 52.2 & 84.6 & 63.6 & \textcolor{red}{66.4} & \textcolor{blue}{62.2} & \textcolor{red}{73.0} & \textcolor{blue}{61.0} & \textcolor{blue}{90.7} & 82.3 & 79.7 & 65.6 & \textcolor{blue}{75.6} & 90.4 & \textcolor{blue}{75.5} & \textbf{47}\\
\hline\\
\\

\end{tabular}}

\subfloat[Mini YouTube-BB]{%
\resizebox{1\textwidth}{!}{%
\begin{tabular}{c|C{0.7cm}C{0.7cm}C{0.7cm}C{0.7cm}C{0.7cm}C{0.7cm}C{0.7cm}C{0.7cm}C{0.7cm}C{0.7cm}C{0.7cm}C{0.7cm}C{0.7cm}C{0.7cm}C{0.7cm}C{0.7cm}C{0.7cm}C{0.7cm}C{0.7cm}C{0.7cm}C{0.7cm}C{0.7cm}C{0.7cm}|c|c}
Method &\rotatebox{60}{person}& \rotatebox{60}{bird} & \rotatebox{60}{boat} & \rotatebox{60}{bike} & \rotatebox{60}{bus} & \rotatebox{60}{bear} & \rotatebox{60}{cow} & \rotatebox{60}{cat} & \rotatebox{60}{giraffe} & \rotatebox{60}{p.plant} & \rotatebox{60}{horse} & \rotatebox{60}{motorcycle} & \rotatebox{60}{knife} & \rotatebox{60}{airplane} & \rotatebox{60}{skateboard} & \rotatebox{60}{train} & \rotatebox{60}{truck} & \rotatebox{60}{zebra} & \rotatebox{60}{toilet} & \rotatebox{60}{dog} & \rotatebox{60}{elephant} & \rotatebox{60}{umbrella} & \rotatebox{60}{car} & \rotatebox{0}{mAP(\%)} & \rotatebox{0}{Runtime(ms)} \\
\hline
SS/SS & 24.9 & 45.3 & 39.3 & 49.1 & 83.1 & 67.8 & 71.8 & 86.5 & 83.7 & 55.0 & 74.4 & 51.8 & 65.1 & 89.9 & 54.2 & 86.7 & 87.1 & 88.5 & 79.7 & 53.5 & 82.8 & 61.1 & 83.5 & 68.0 & 75\\
\hline
MS/SS & 22.4 & 49.4 & 42.0 & 61.7 & 84.2 & 71.0 & 71.3 & 85.1 & 85.9 & 49.5 & 69.3 & 52.1 & 62.1 & 88.8 & 56.1 & 88.1 & 86.8 & 89.2 & 83.1 & 52.5 & 79.9 & 61.5 & 83.4 & 68.5 & 75\\
\hline
~~~MS/AdaScale~~~ & \textcolor{blue}{26.2} & \textcolor{blue}{53.2} & \textcolor{blue}{41.9} & \textcolor{blue}{63.6} & 83.4 & \textcolor{blue}{72.6} & 72.0 & \textcolor{blue}{87.6} & \textcolor{blue}{86.8} & \textcolor{blue}{57.8} & \textcolor{blue}{75.4} & \textcolor{blue}{59.0} & \textcolor{blue}{70.4} & 89.7 & \textcolor{red}{52.5} & 86.7 & 87.2 & 89.0 & \textcolor{blue}{83.8} & 53.3 & 81.4 & \textcolor{blue}{66.4} & \textcolor{blue}{85.7} & \textcolor{blue}{70.7} & \textbf{41}\\
\hline
\end{tabular}}}
\end{table*}

To evaluate the proposed AdaScale, we progressively compare the three methods: (i) SS/SS - a detector trained and tested at 600, which is usually adopted by prior art~\cite{renNIPS15fasterrcnn,dai16rfcn,zhu2017deep,zhu2017flow,feichtenhofer2017detect}, (ii) MS/SS - a detector trained at $S_{train}$ and tested at 600, and (iii) MS/AdaScale - a detector trained at $S_{train}$ and tested on an adaptively changing scale between 128 and 600, given the range of $S_{reg}$. Note that the scale for MS/AdaScale can be any integer value within this range since it is predicted by the scale regressor. The evaluation results are shown in Table~\ref{table:eval}. From this point on, for the sake of simplicity, we base our analysis on ImageNet VID only. The analysis holds for mini YouTube-BB as well.

\textit{Accuracy:} Compared to the baseline SS/SS, MS/AdaScale increases mAP by 1.3 points. For better visualization, blue numbers in Table~\ref{table:eval} indicate $\ge 1$ AP improvement while red numbers represent $\ge 1$ AP degradation. Our approach achieves $\ge 1$ AP improvements in half of the categories with only three categories having $\ge 1$ AP degradation. 
In general, multi-scale training can enrich the training data and achieve better generalization of the model. However, this is not always the case. For categories like \textit{red panda} and \textit{bear}, there is a huge AP degradation for all the multi-scale training-based approaches. We find that multi-scale training could potentially lead to some confusion for certain categories. We leave the in-depth study of this phenomena to future work. 

\begin{figure}[t]
    \subfloat[][Lion]{\includegraphics[width=.33\linewidth]{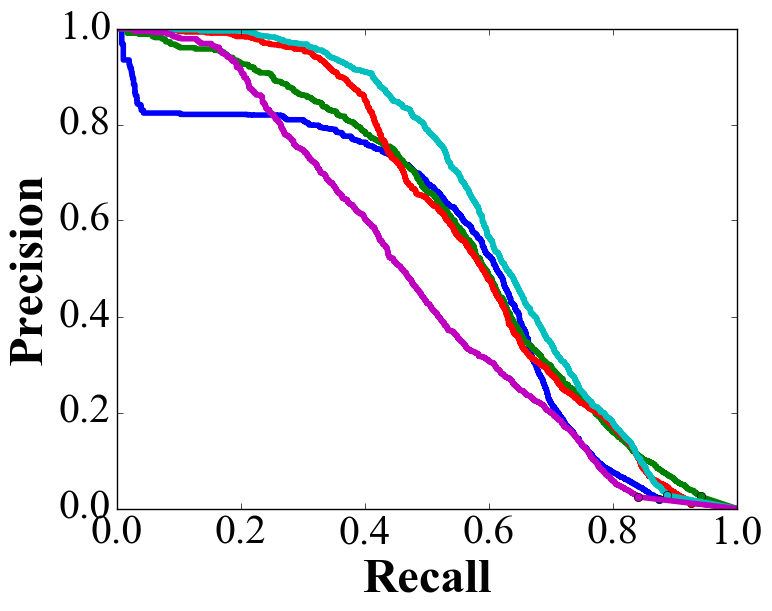}\label{fig:prcurve_lion}}
    \subfloat[][Squirrel]{\includegraphics[width=.33\linewidth]{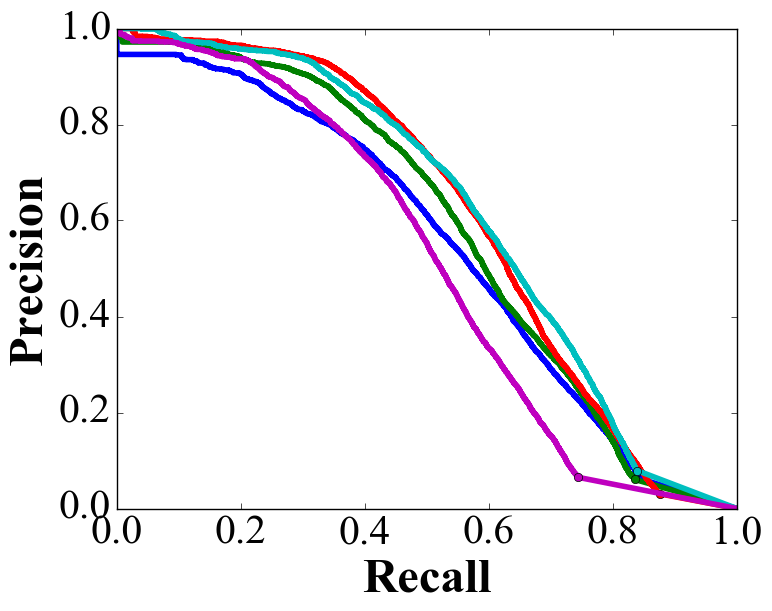}\label{fig:prcurve_squirrel}}
    \subfloat[][Horse]{\includegraphics[width=.33\linewidth]{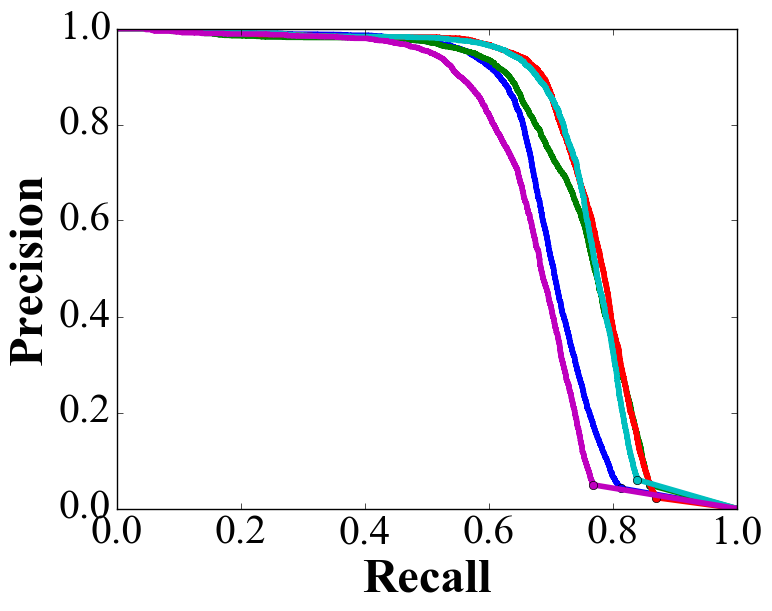}\label{fig:prcurve_horse}}\\
    \subfloat[][Airplane]{\includegraphics[width=.33\linewidth]{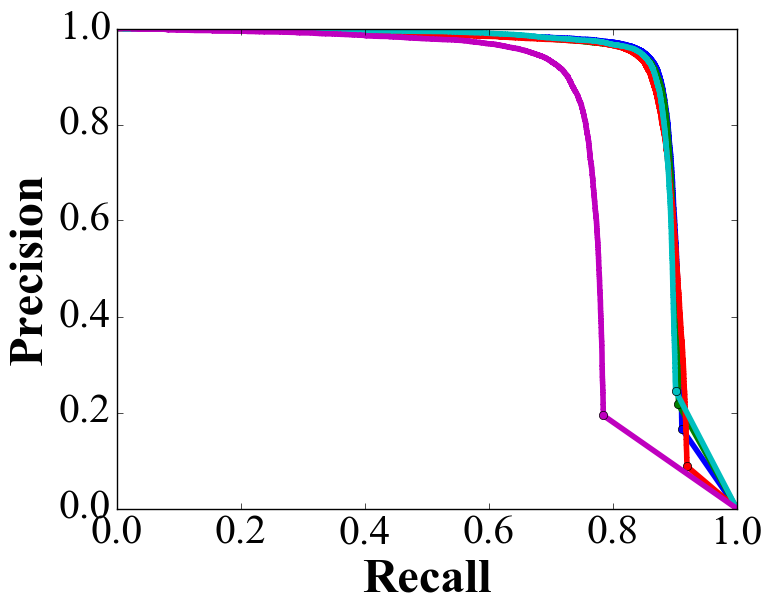}\label{fig:prcurve_airplane}}
    \subfloat[][Red panda]{\includegraphics[width=.33\linewidth]{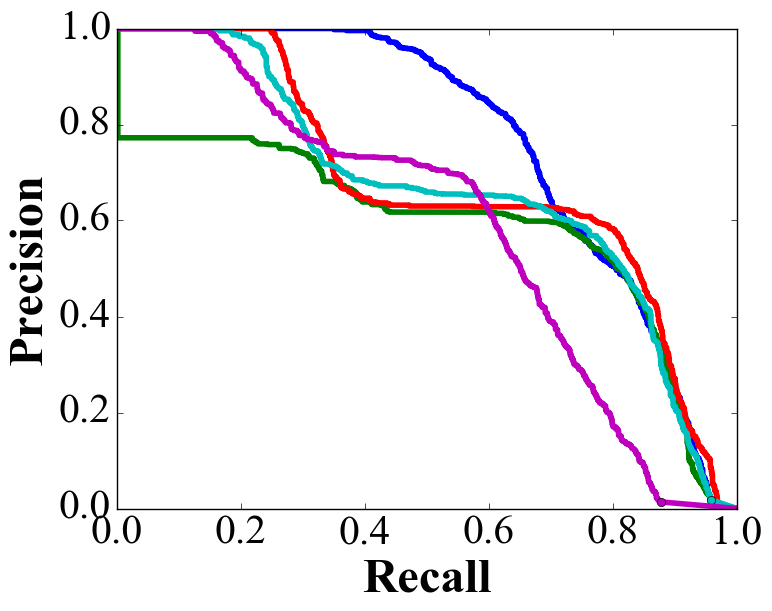}\label{fig:prcurve_redpanda}}
    \subfloat[][Bear]{\includegraphics[width=.33\linewidth]{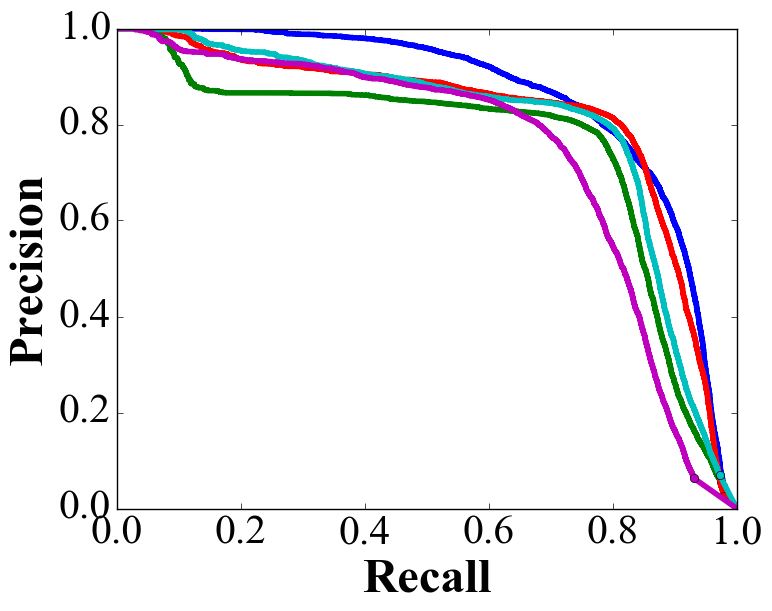}\label{fig:prcurve_bear}}
    \begin{center}
    \includegraphics[width=0.7\linewidth]{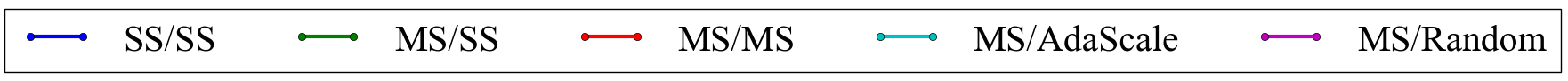}\label{fig:prcurve_legend}
    \vspace{-8pt}
    \end{center}
    \caption{Precision-Recall curves for categories that MS/AdaScale has (a)(b)(c) better performance, (d) on-par performance, and (e)(f) worse performance compared to SS/SS.}
    \label{fig:prcurve}
\end{figure}

We further dive into the precision-recall curve to understand the dynamics of precision and recall for all the methods. Fig.~\ref{fig:prcurve} shows that precision-recall curves for three most improved categories~(a)-(c), one on-par category~(d), and two most degraded categories~(e)-(f). To give a more comprehensive analysis, we add multi-scale training and multi-scale testing (MS/MS) here for comparison. In addition, we also compare with multi-scale training and random testing scenario, which selects one of the five scales in $S_{reg}$ randomly at test time. Compared to random scaling, MS/AdaScale clearly learns the dynamics of when and how to scale to be able to have consistently higher average precision. Additionally, we can tell from the figure that irrespective of getting better or worse compared to SS/SS, MS/AdaScale follows the curve of MS/MS closely.

\textit{Speed:}
Our scale regressor incurs only 2ms of overhead, which is 3\% of the runtime of R-FCN. To see the speed improvement brought by the MS/AdaScale, Fig.~\ref{fig:histogram}(a) shows the size distribution produced by the scale regressor on ImageNet VID validation dataset and we conduct speed sensitivity analysis on scale set $S_{train}$ in section~\ref{sec:ablation}. We note that, to profile the runtime, we warm up the GPU memory in order to remove the impact of memory allocation overhead of MXNet~\cite{chen2015mxnet}.

\subsection{Higher Precision with AdaScale}

\begin{figure}[t]
    \subfloat[][Lion]{\includegraphics[width=.33\linewidth]{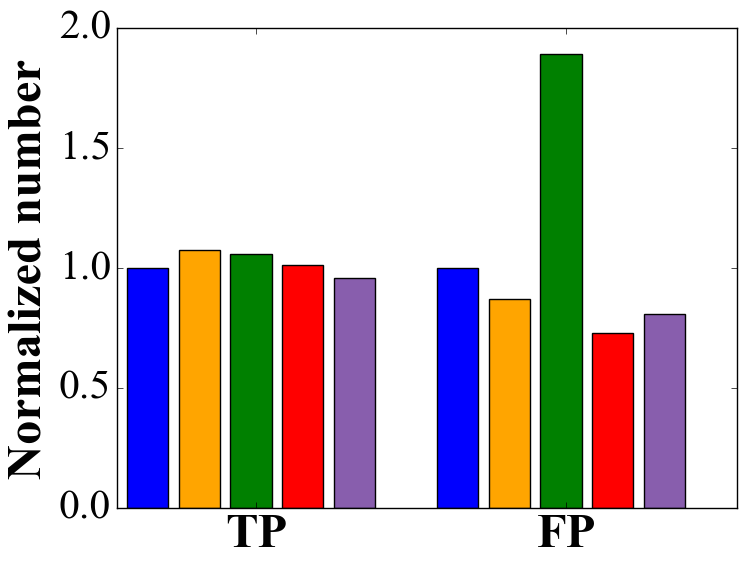}\label{fig:tpfp_lion}}
    \subfloat[][Squirrel]{\includegraphics[width=.33\linewidth]{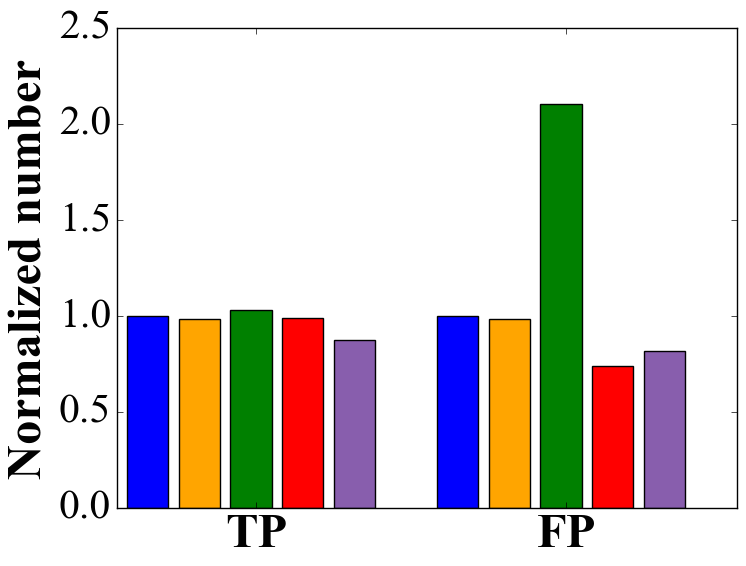}\label{fig:tpfp_squirrel}}
    \subfloat[][Horse]{\includegraphics[width=.33\linewidth]{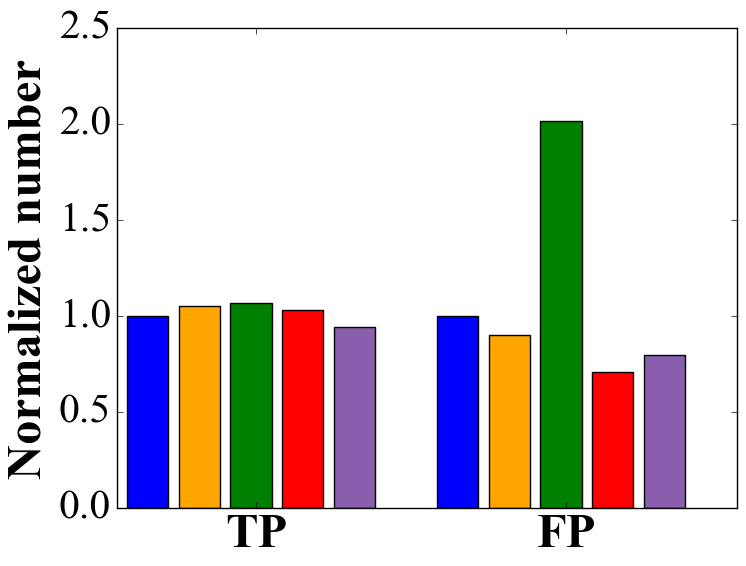}\label{fig:tpfp_horse}}\\
    \subfloat[][Airplane]{\includegraphics[width=.33\linewidth]{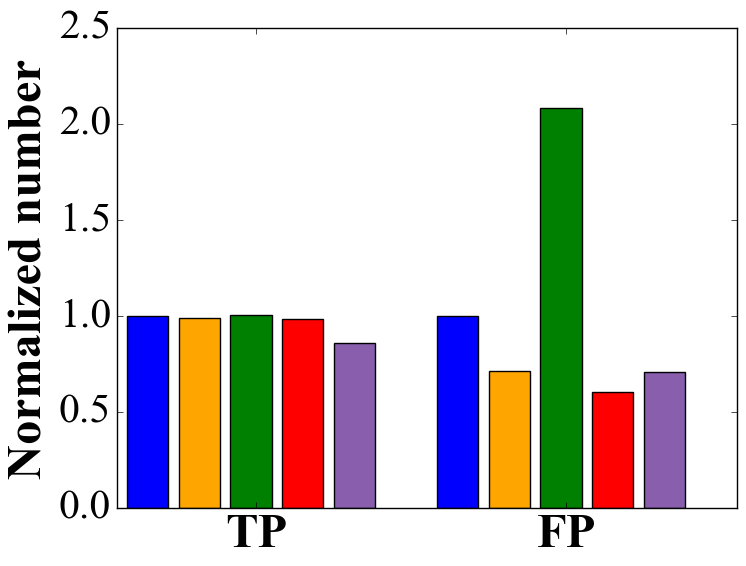}\label{fig:tpfp_airplane}}
    \subfloat[][Red Panda]{\includegraphics[width=.33\linewidth]{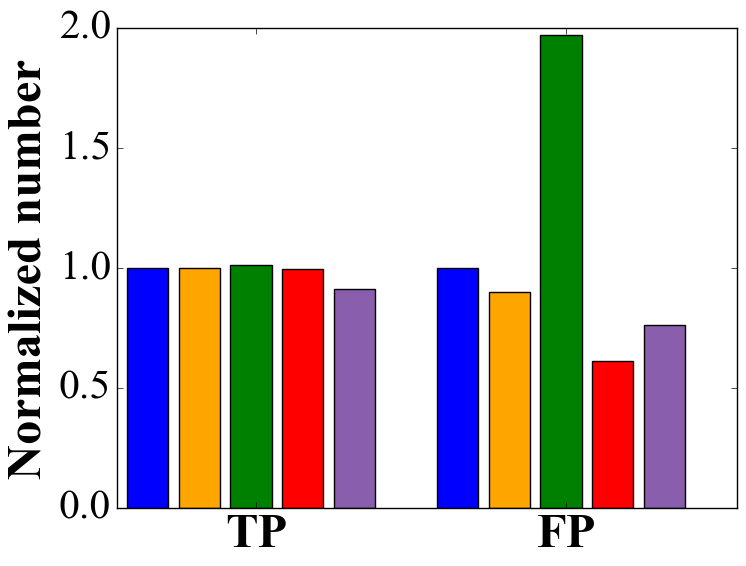}\label{fig:tpfp_redpanda}}
    \subfloat[][Bear]{\includegraphics[width=.33\linewidth]{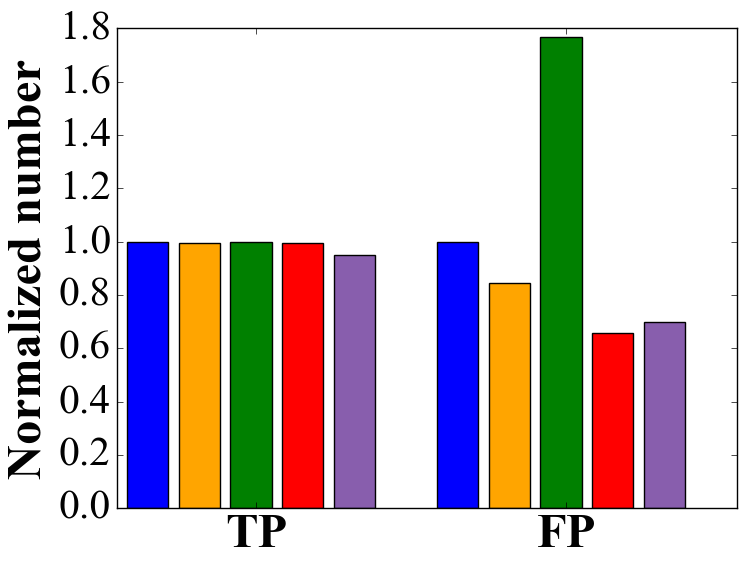}\label{fig:tpfp_bear}}
    \begin{center}
    %\vspace{-12pt}
    \includegraphics[width=0.7\linewidth]{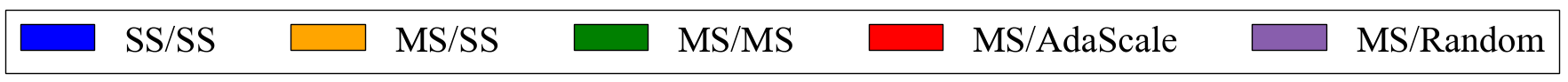}\label{fig:tpfp_legend}
    \vspace{-6pt}
    \end{center}
    \caption{Normalized true positives and false positives for different methods across all the images in validation set for three selected categories.}
    \label{fig:tpfp}
\end{figure}

We further dig into what our method actually improves - precision or recall. As mentioned earlier in section~\ref{sec:regressor}, adaptive scaling could possibly increase true positives by scaling the object into a better scale for the detector or reduce false positives by not focusing too much on unnecessary details. To conduct this analysis, we compute the number of true positives and false positives across all the images in the validation set for method SS/SS, MS/SS, MS/MS, MS/AdaScale, as well as MS/Random. Fig.~\ref{fig:tpfp} shows the number of true positives and the number of false positives normalized to method SS/SS (we present the results for all categories in Appendix). First, by comparing SS/SS and MS/SS, we can observe that multi-scale training is able to lower the number of false positives dramatically. This is reasonable since multi-scale training reduces the chance that the classifier counts on scale information as a discriminating feature. The results of MS/SS and MS/Random show that simply down-sampling images can also reduce false positive, but it reduces true positives as well. In addition to the false positive reduction brought by multi-scale training and image down-sampling, MS/AdaScale manages to reduce even more false positives, with true positives comparable to SS/SS. In general, MS/AdaScale is able to increase precision at a slight cost of recall degradation.

\begin{figure}[t]
    \centering
    \includegraphics[width=0.8\linewidth]{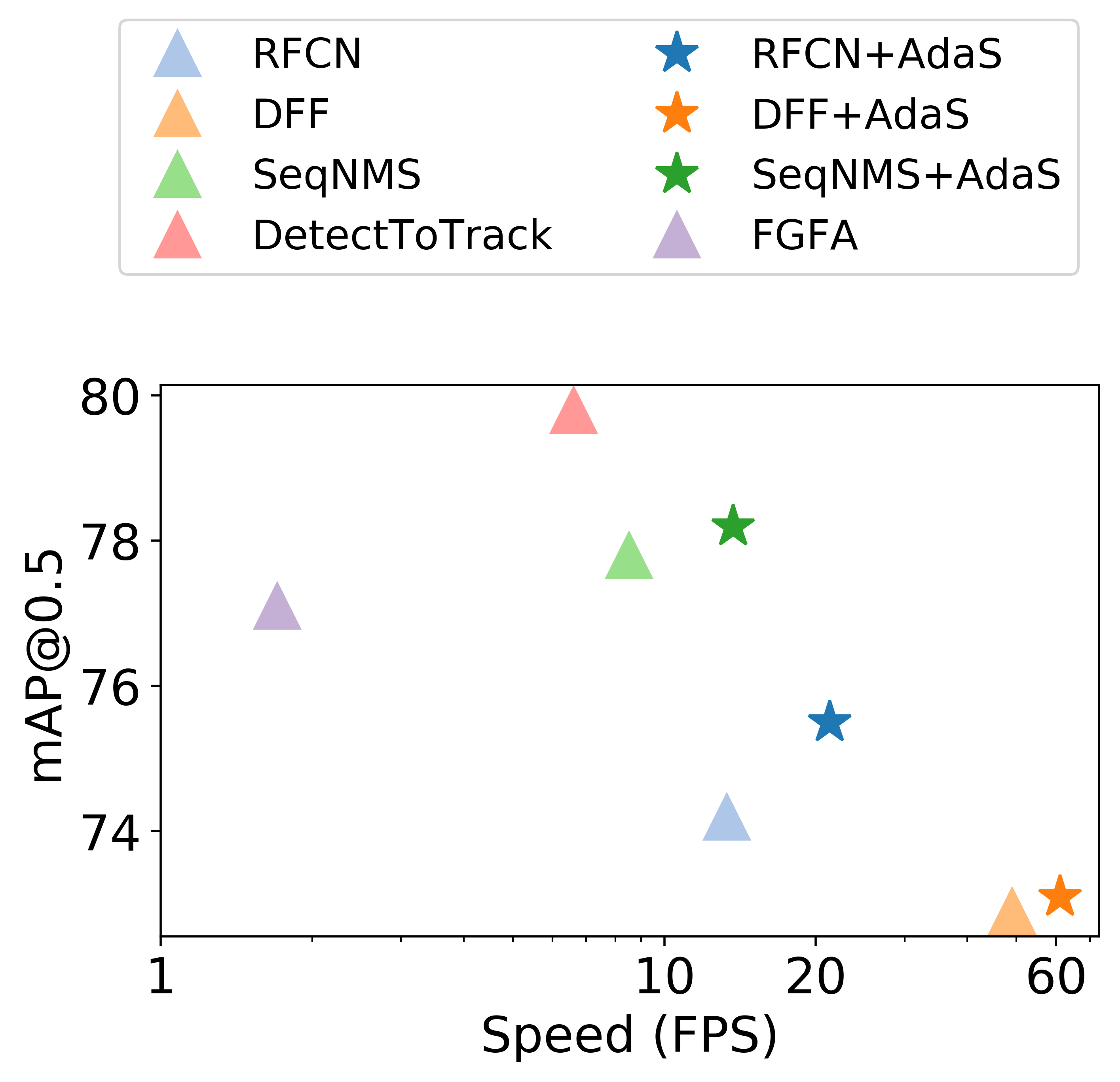}
    \vspace{-6pt}
    \caption{mAP and speed comparison with prior art on ImageNet VID dataset. Applying our AdaScale to RFCN~\cite{dai16rfcn}, DFF~\cite{zhu2017flow} and SeqNMS~\cite{han2016seq} can further improve both speed and accuracy.}
    \vspace{-6pt}
    \label{fig:prior}
\end{figure}

\subsection{Qualitative Results}

\begin{figure*}[t]
    \centering
    \subfloat[][SS/SS]{\begin{tabular}[b]{c}\includegraphics[width=.22\linewidth]{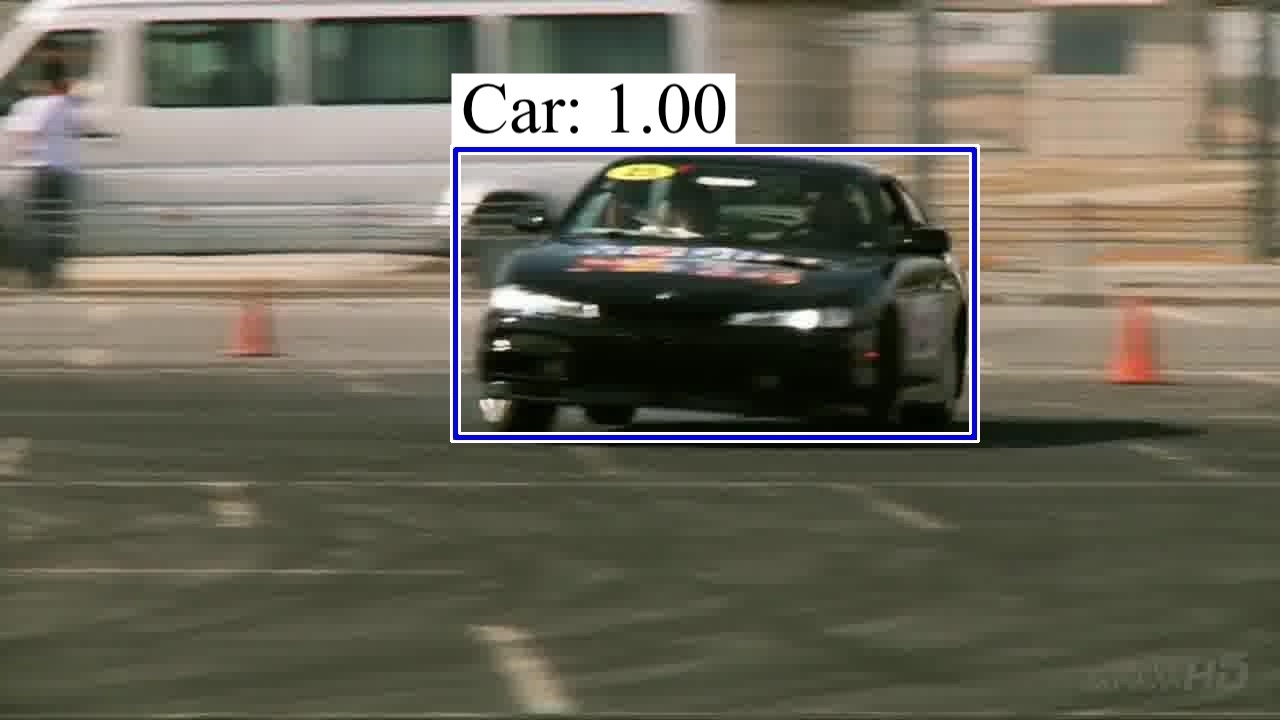}\\
    \includegraphics[width=.22\linewidth]{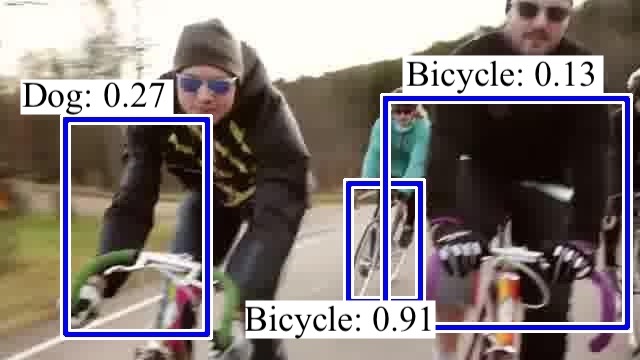}\\
    \includegraphics[width=.22\linewidth]{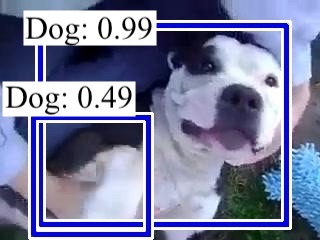}\\
    \includegraphics[width=.22\linewidth]{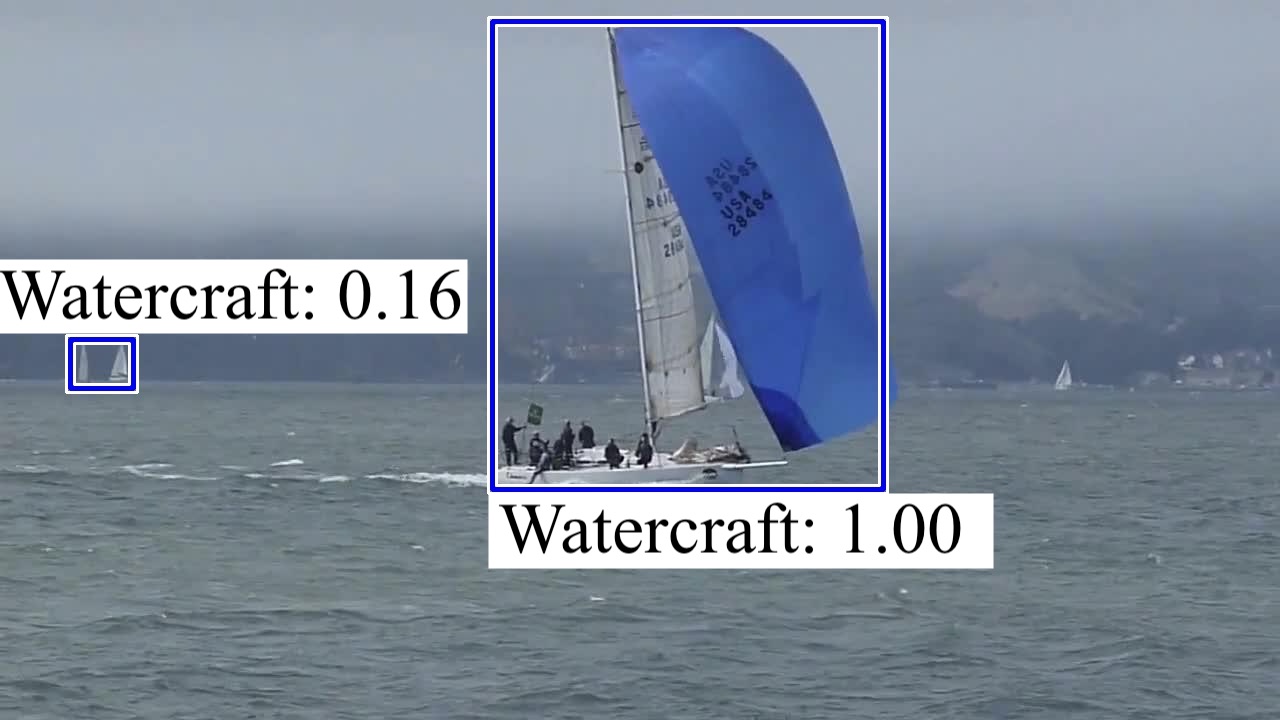}
    \end{tabular}\label{fig:qual:col1}}
    \subfloat[][MS/AdaScale]{\begin{tabular}[b]{c}\includegraphics[width=.22\linewidth]{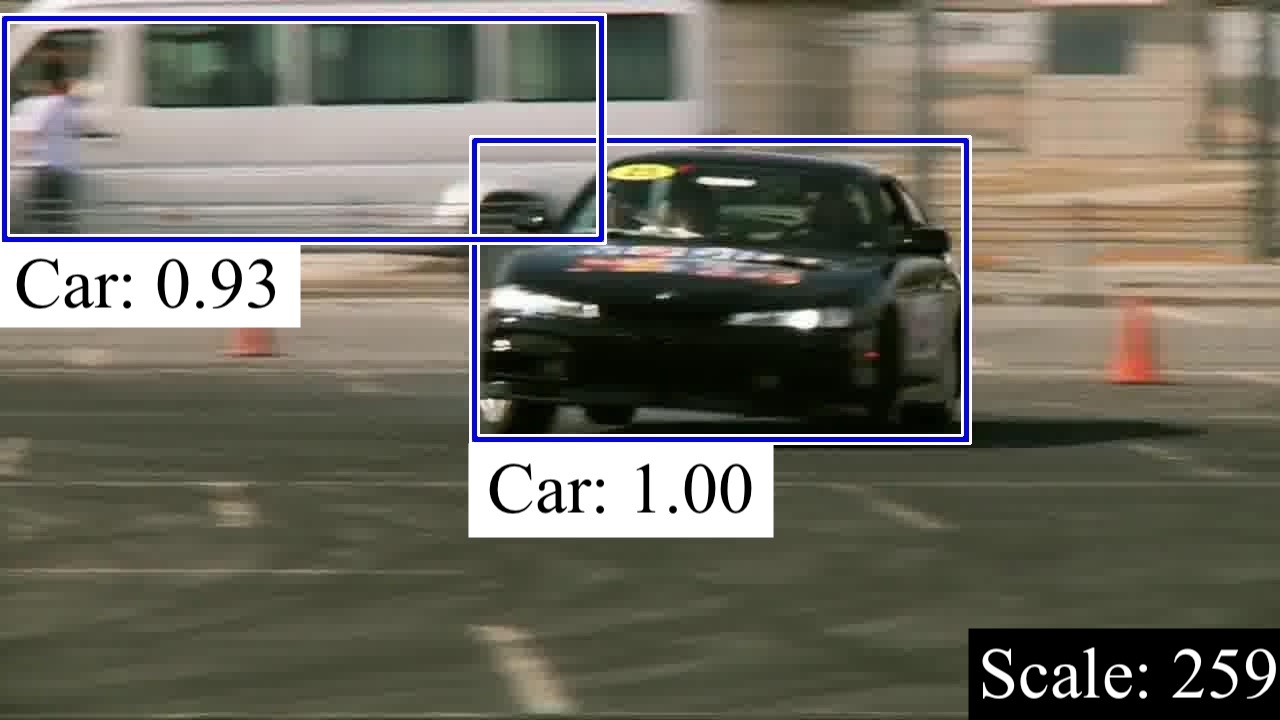}\\
    \includegraphics[width=.22\linewidth]{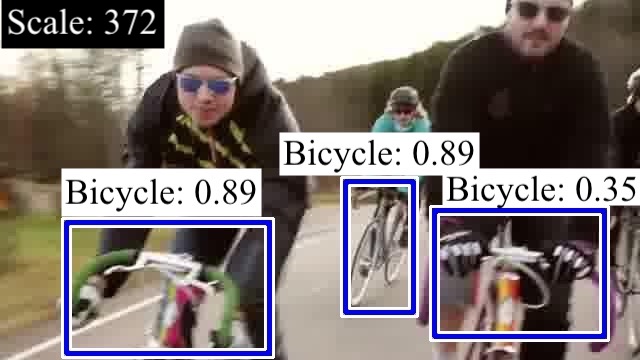}\\
    \includegraphics[width=.22\linewidth]{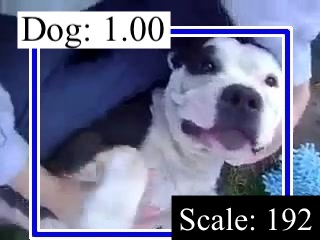}\\
    \includegraphics[width=.22\linewidth]{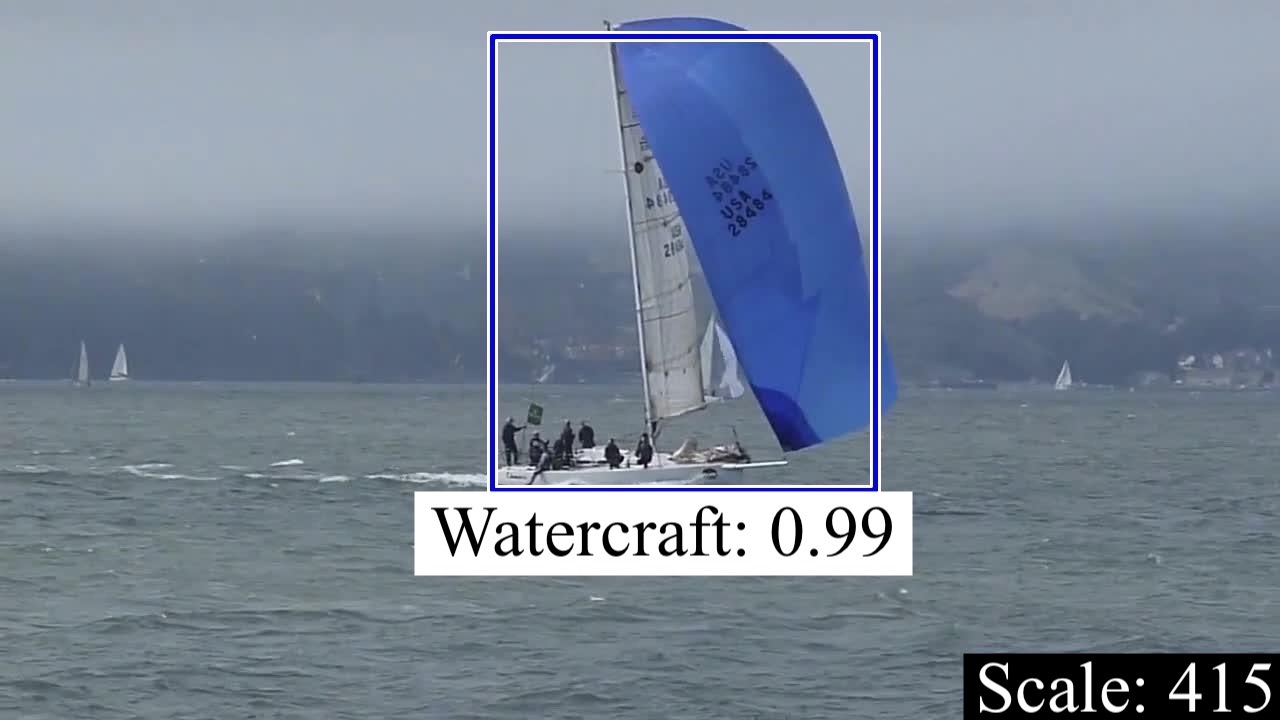}
    \end{tabular}\label{fig:qual:col2}}
    \subfloat[][SS/SS]{\begin{tabular}[b]{c}\includegraphics[width=.22\linewidth]{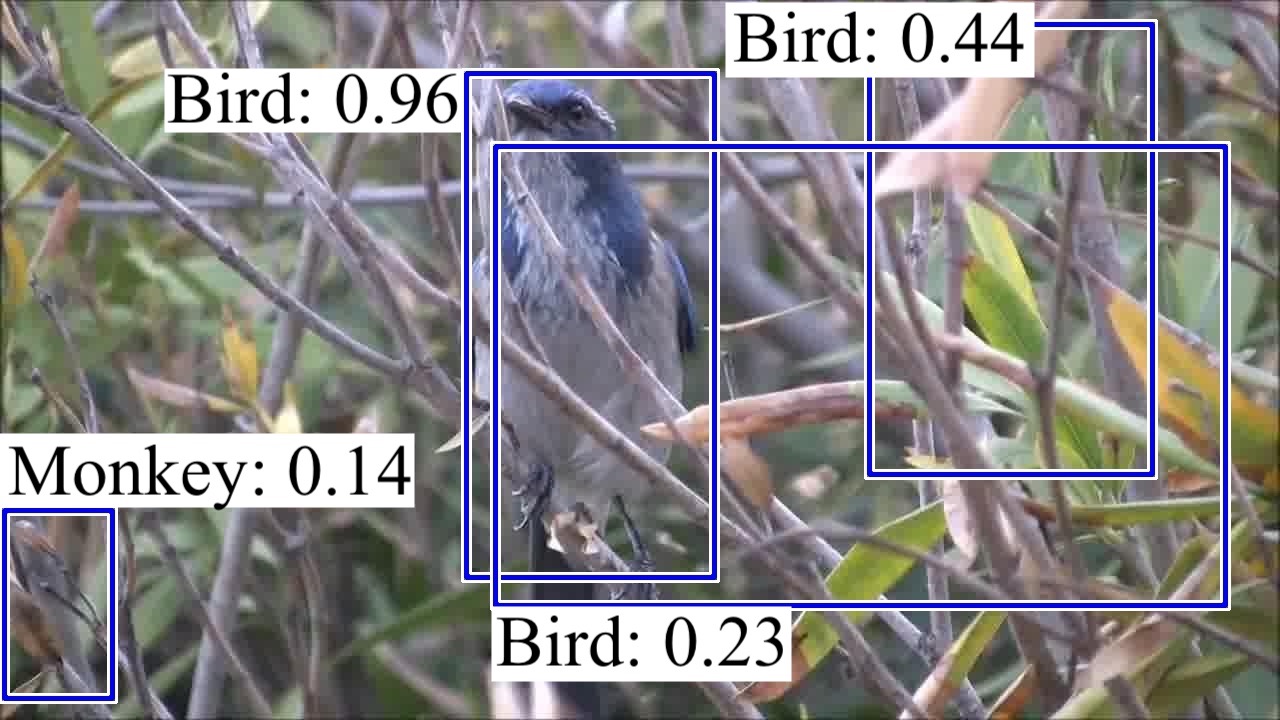}\\
    \includegraphics[width=.22\linewidth]{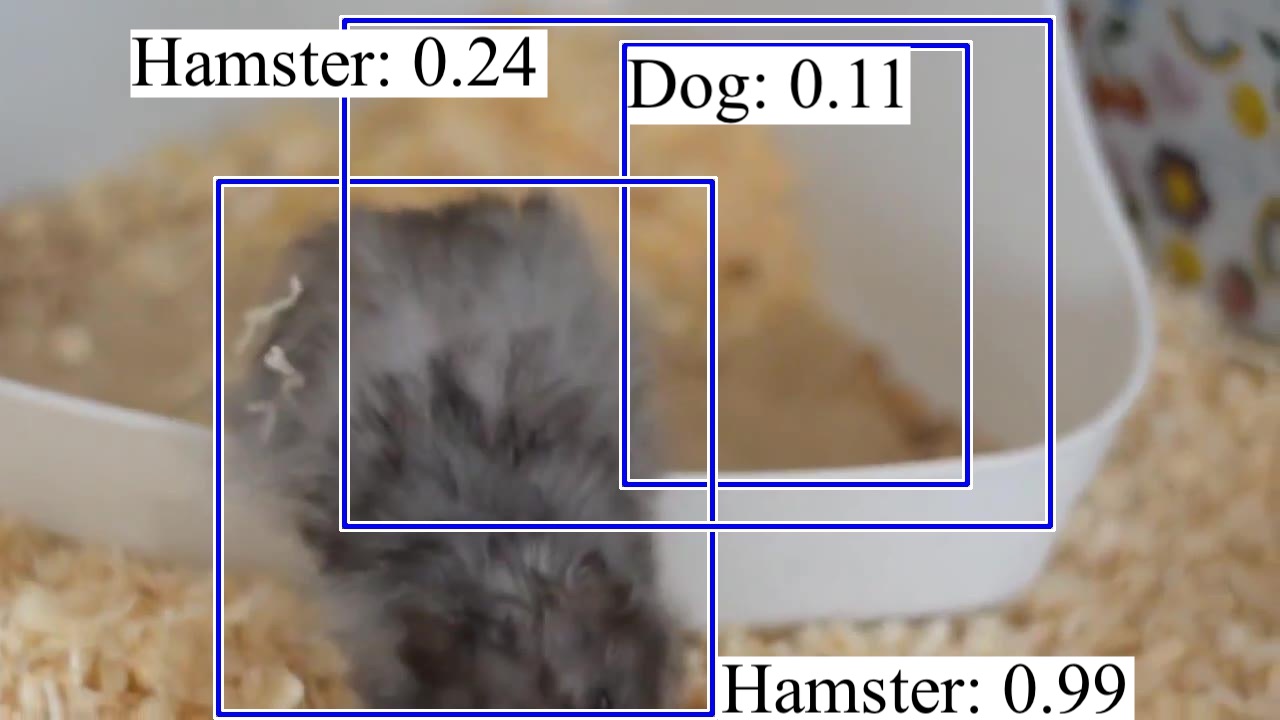}\\
    \includegraphics[width=.22\linewidth]{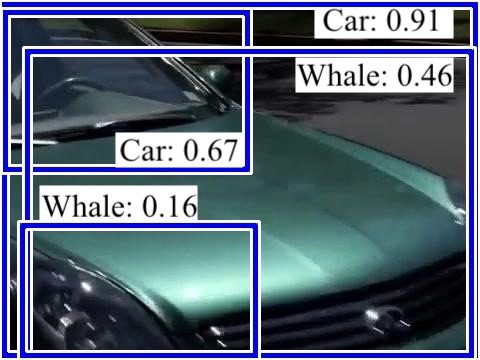}\\
    \includegraphics[width=.22\linewidth]{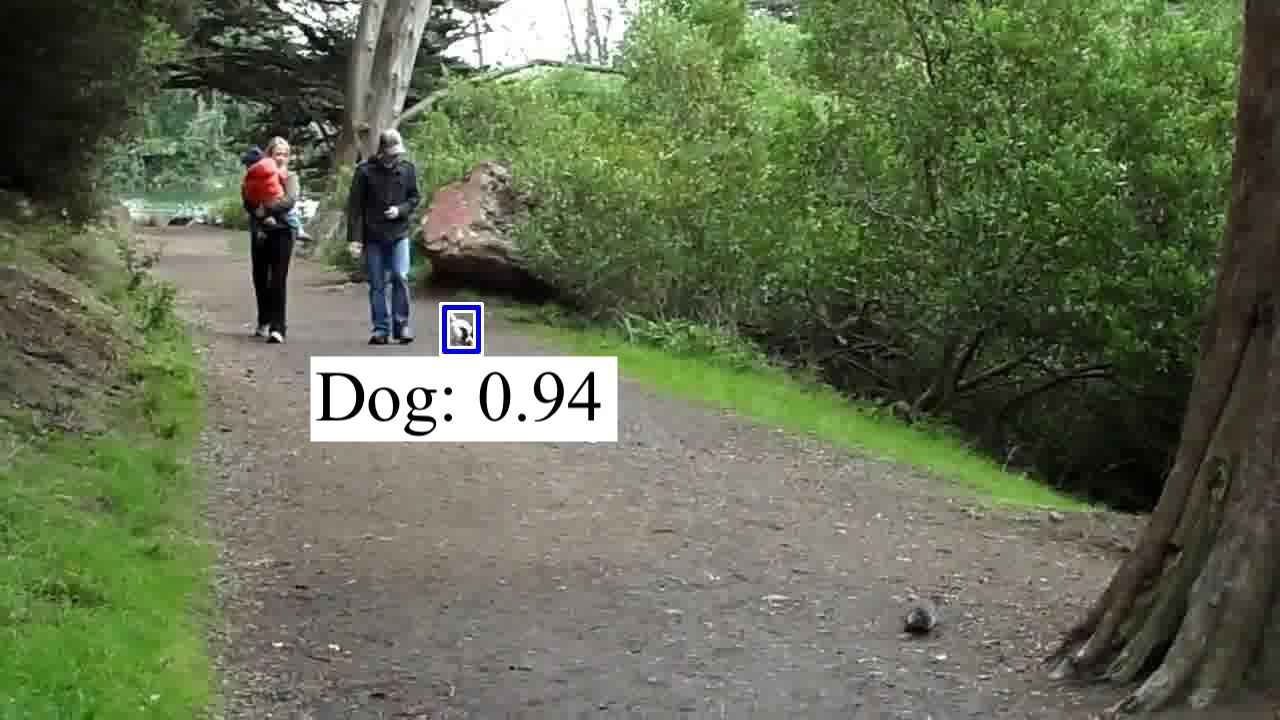}
    \end{tabular}\label{fig:qual:col3}}
    \subfloat[][MS/AdaScale]{\begin{tabular}[b]{c}\includegraphics[width=.22\linewidth]{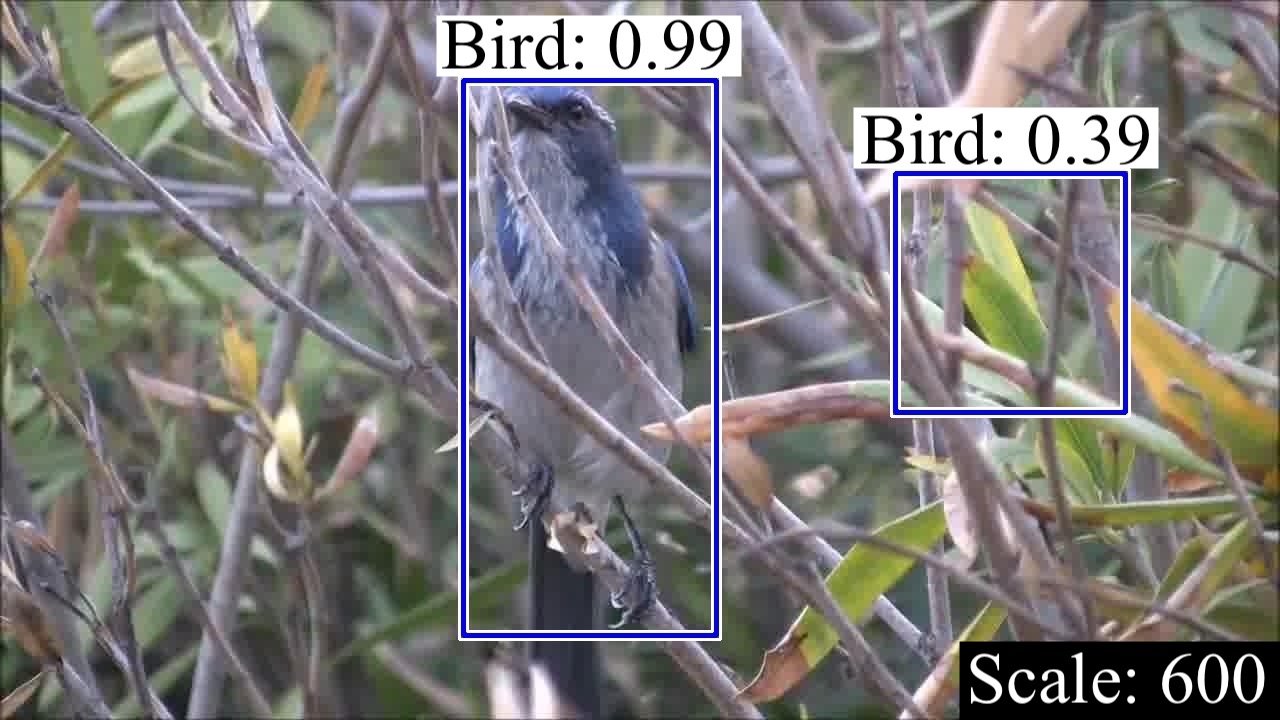}\\
    \includegraphics[width=.22\linewidth]{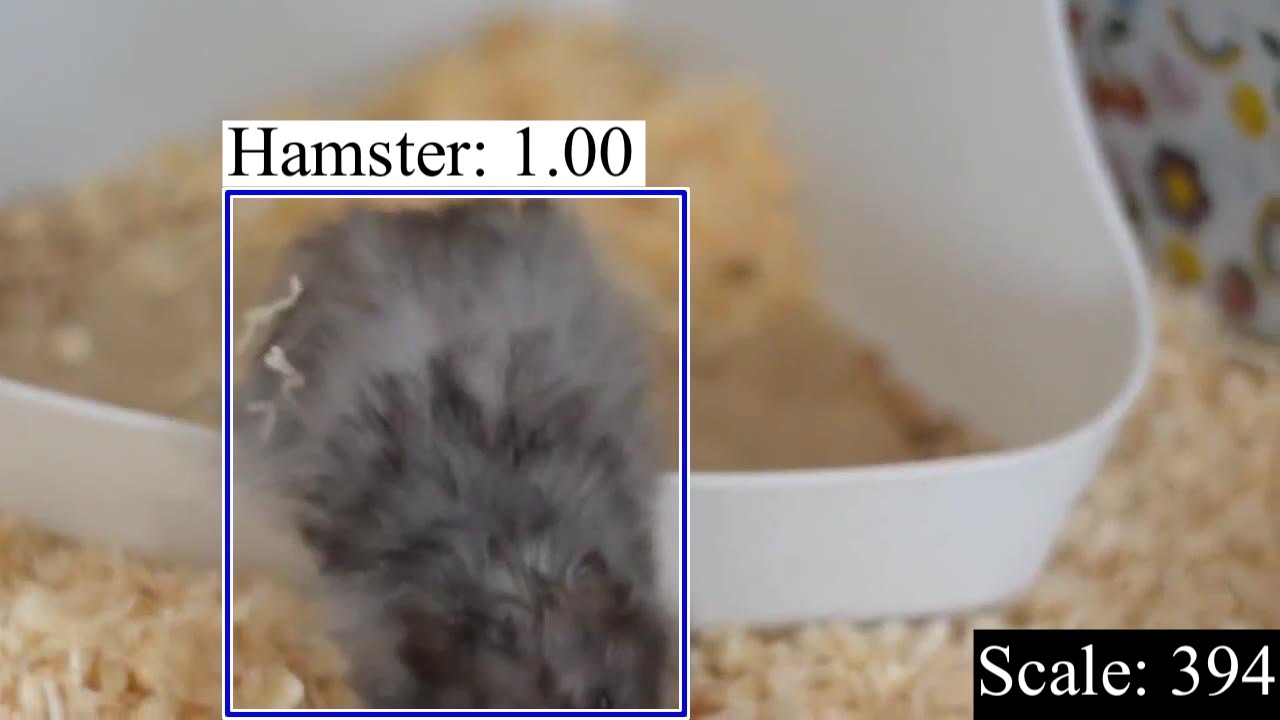}\\
    \includegraphics[width=.22\linewidth]{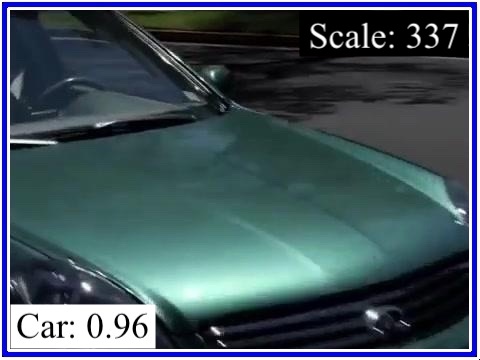}\\
    \includegraphics[width=.22\linewidth]{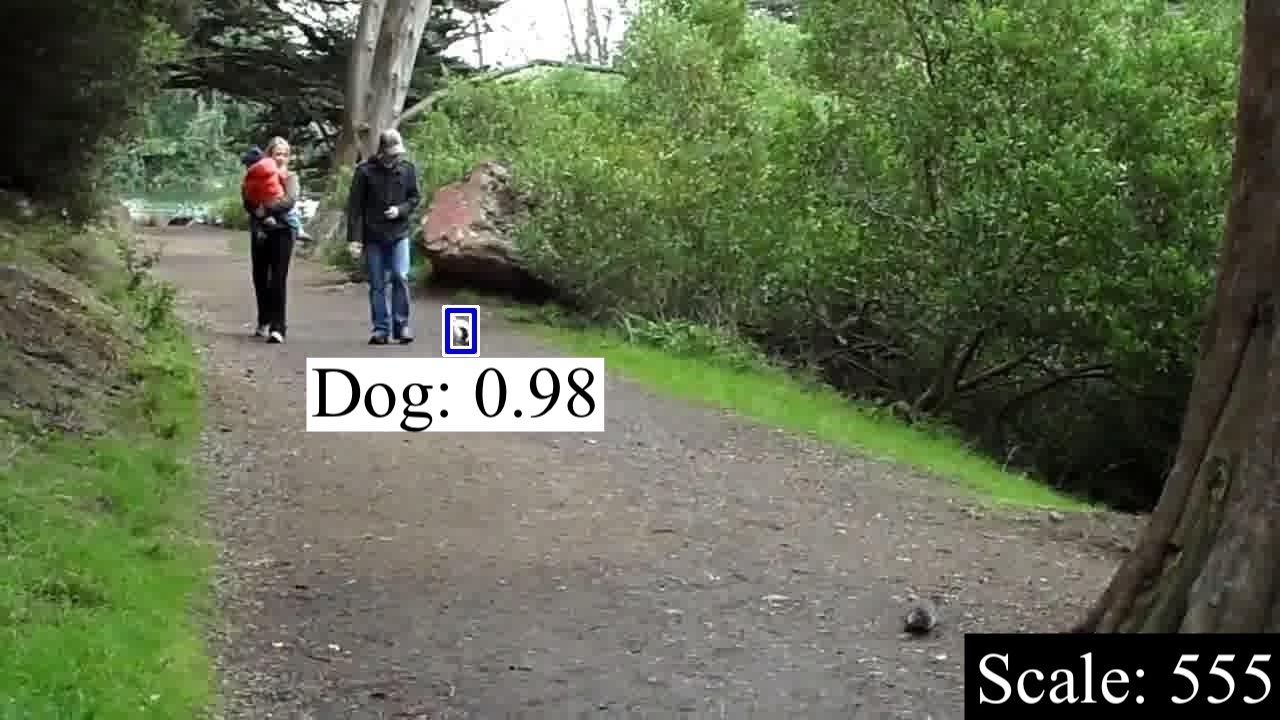}
    \end{tabular}\label{fig:qual:col4}}
    \caption{Comparing the results of SS/SS and MS/AdaScale qualitatively. Column (a) and (c) are results produced by SS/SS; column (b) and (d) are results produced by MS/AdaScale. The scales used in MS/AdaScale are labeled in black rectangle with white text.}
    \label{fig:qual}
\end{figure*}

In Fig.~\ref{fig:qual}, we show some example images for the detection results of both the baseline SS/SS and MS/AdaScale. First, we observe that the regressor learns to down-sample the image when there is a large object in the image. On the other hand, it stays in higher scales if there is a small object in the image. Also, we notice that the regressor learns to scale to the right size to avoid false positives and even correct predictions with false classes.

\begin{figure*}[ht]
    \begin{tabular}[b]{ccccc}
    \includegraphics[width=.17\linewidth]{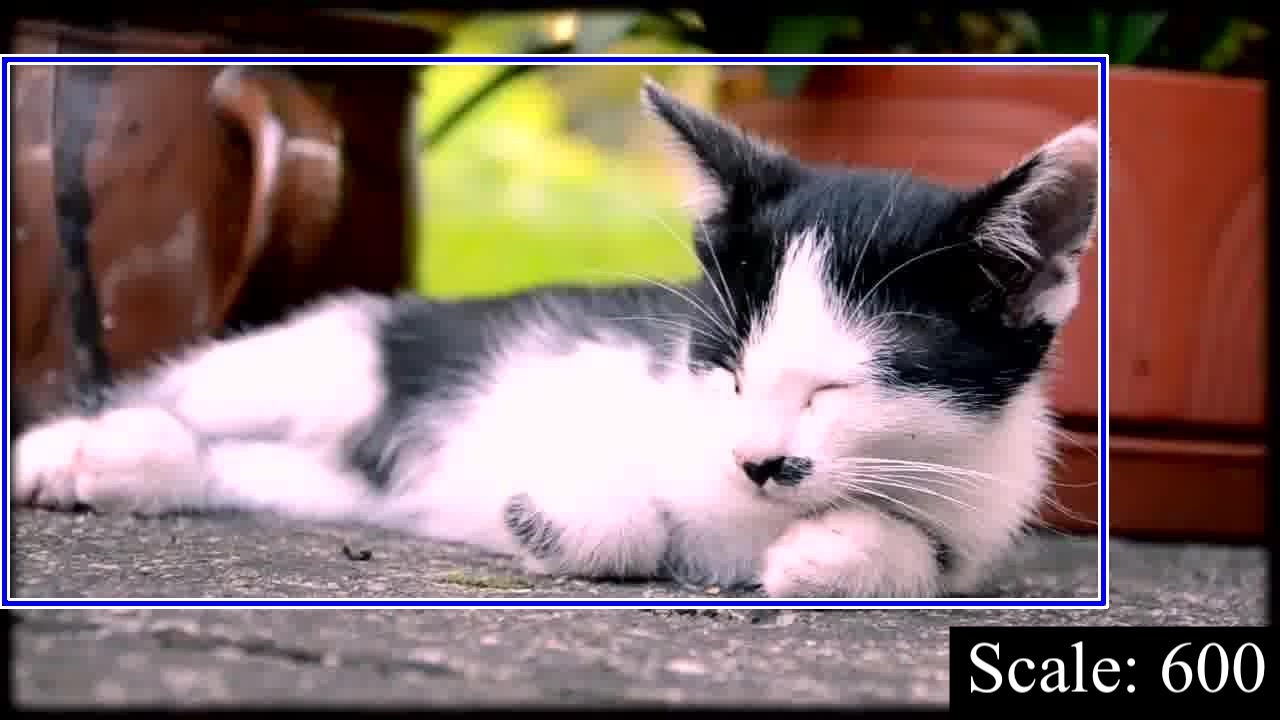} &
    \includegraphics[width=.17\linewidth]{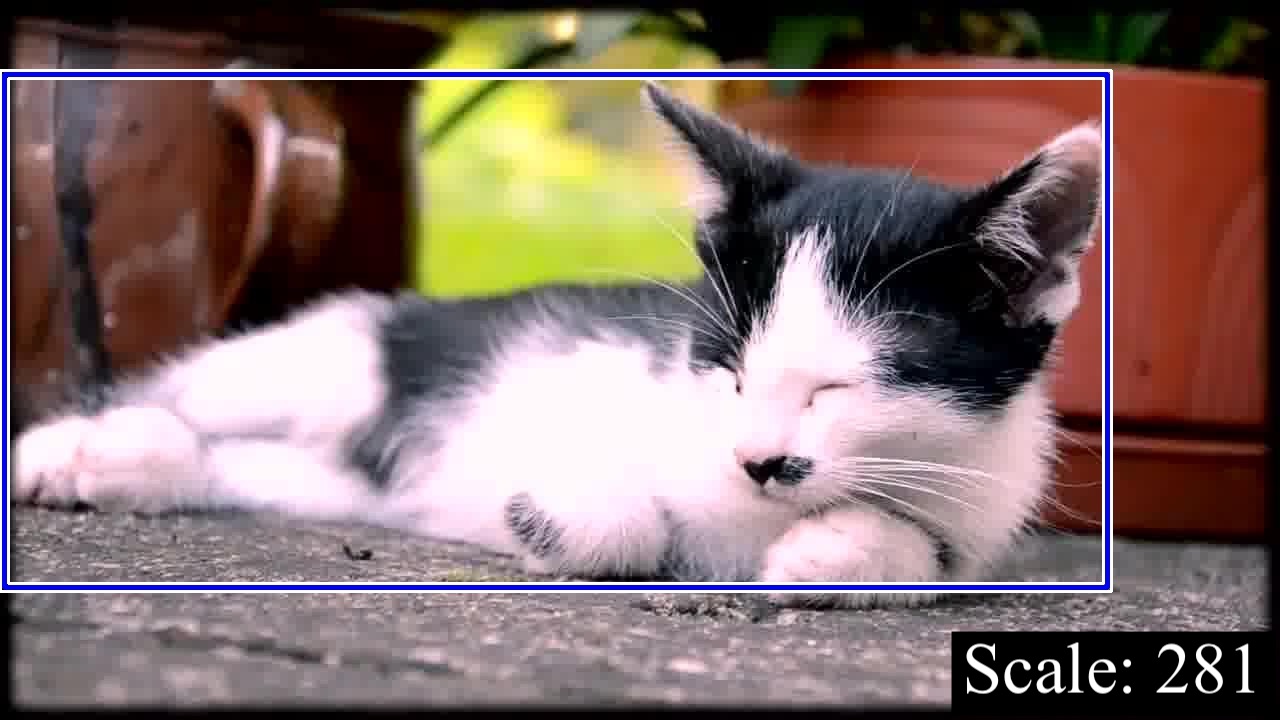} &
    \includegraphics[width=.17\linewidth]{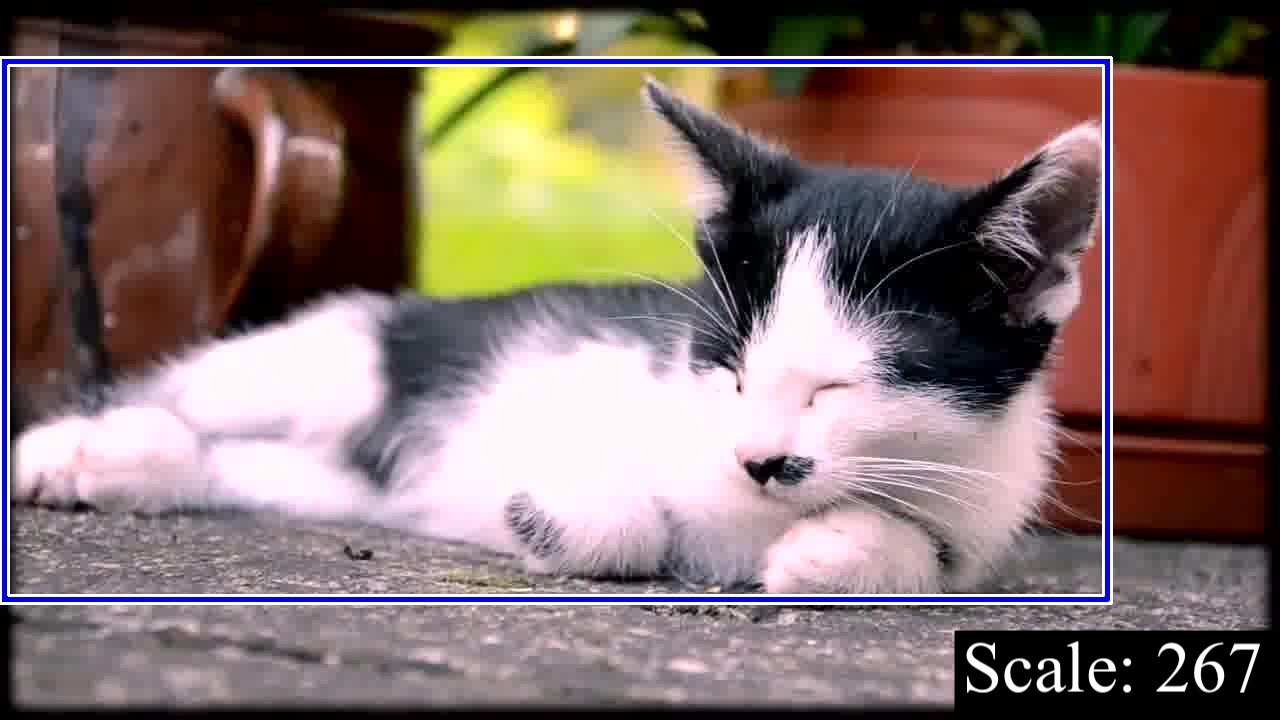} &
    \includegraphics[width=.17\linewidth]{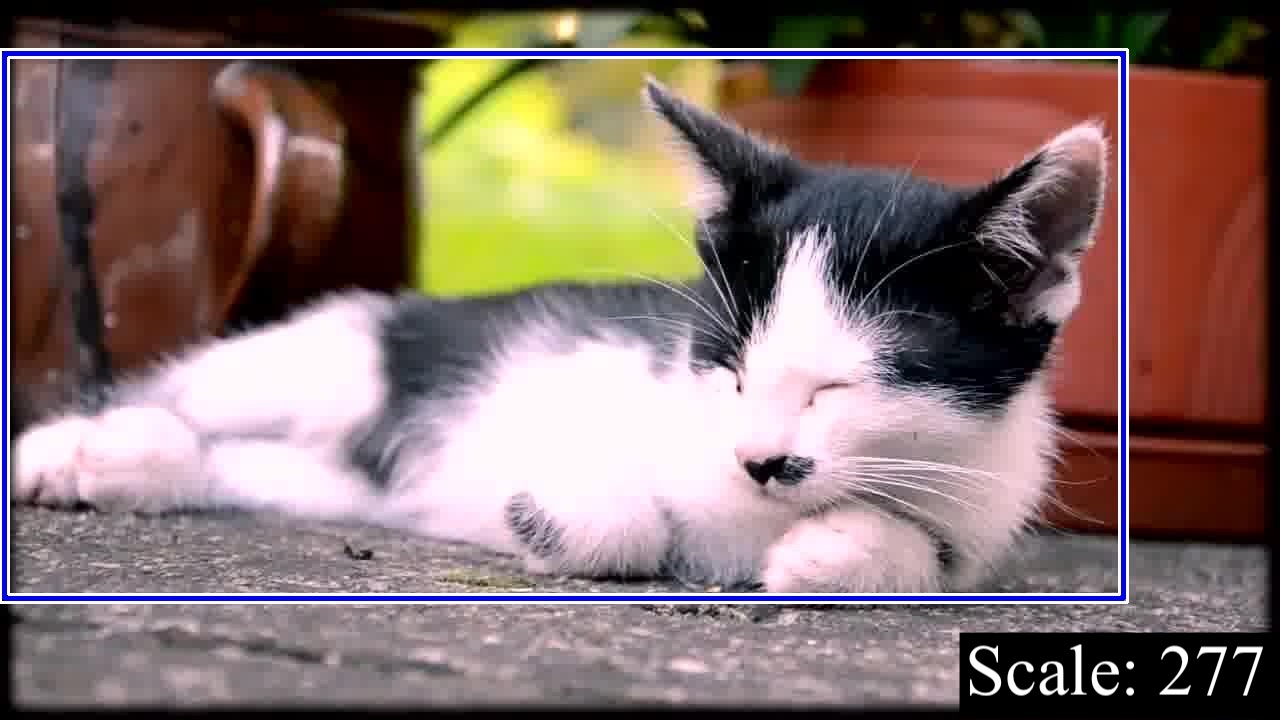} &
    \includegraphics[width=.17\linewidth]{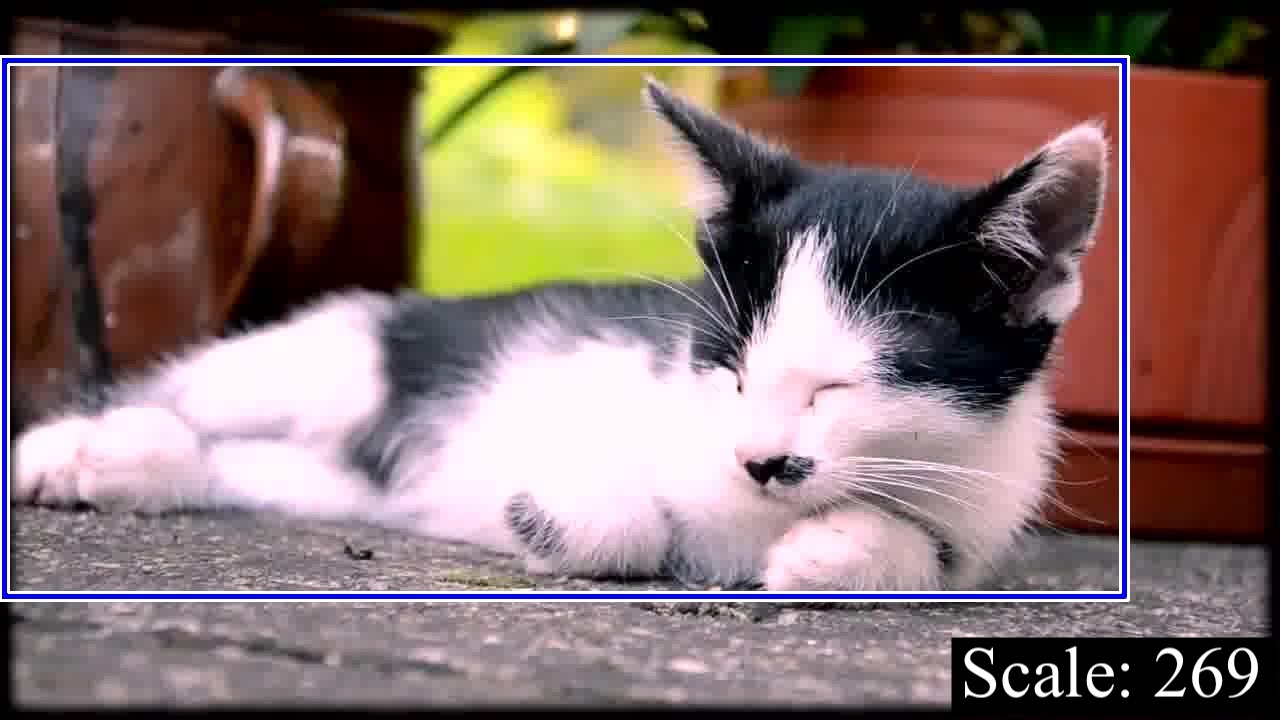} \\
    \includegraphics[width=.17\linewidth]{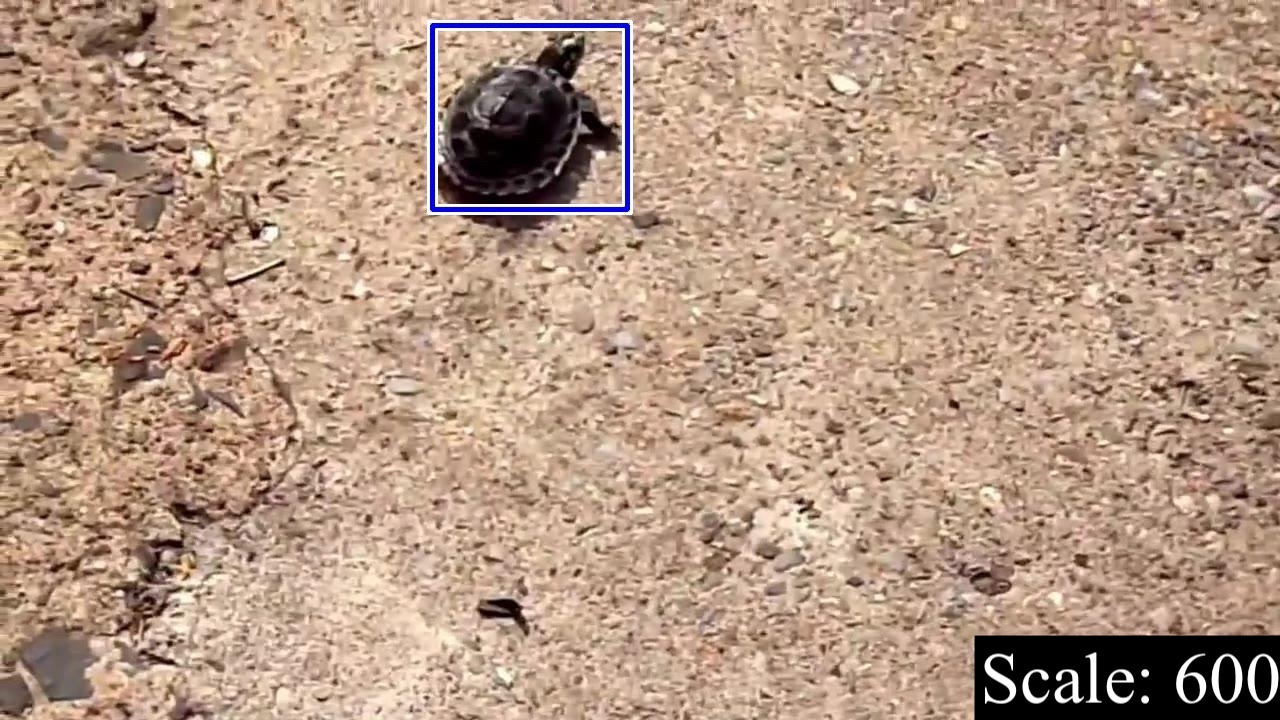} &
    \includegraphics[width=.17\linewidth]{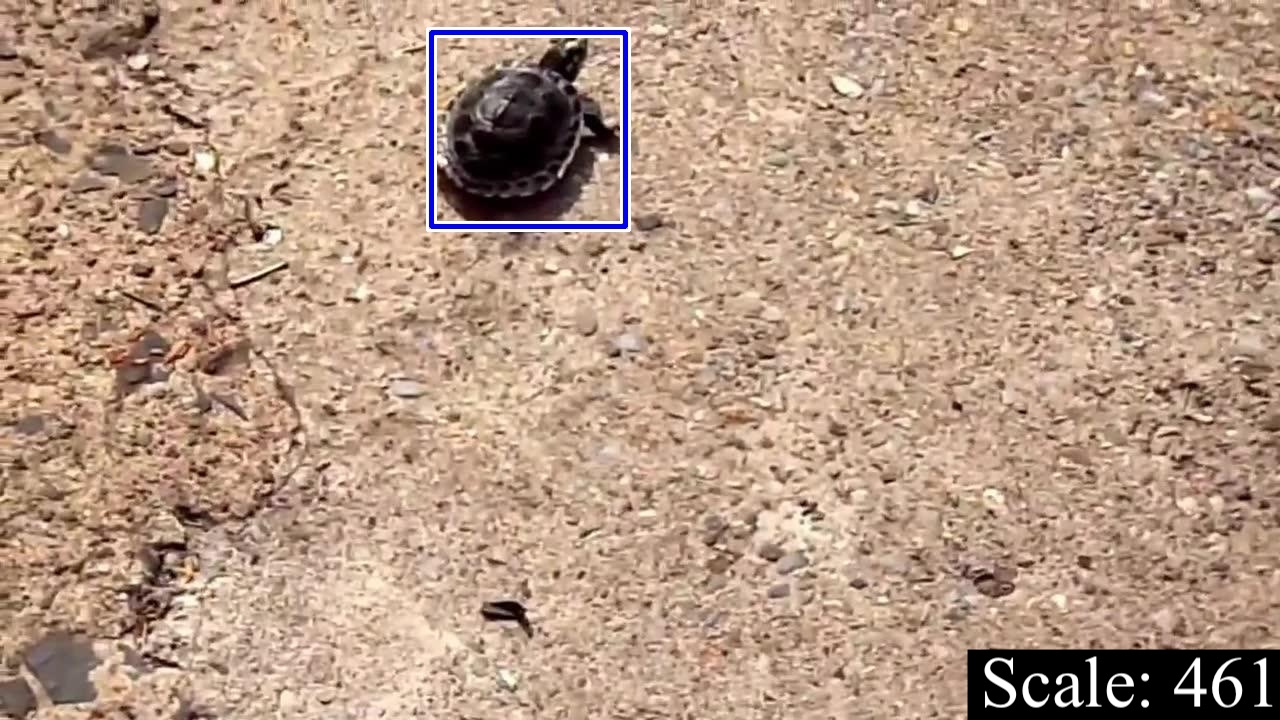} & 
    \includegraphics[width=.17\linewidth]{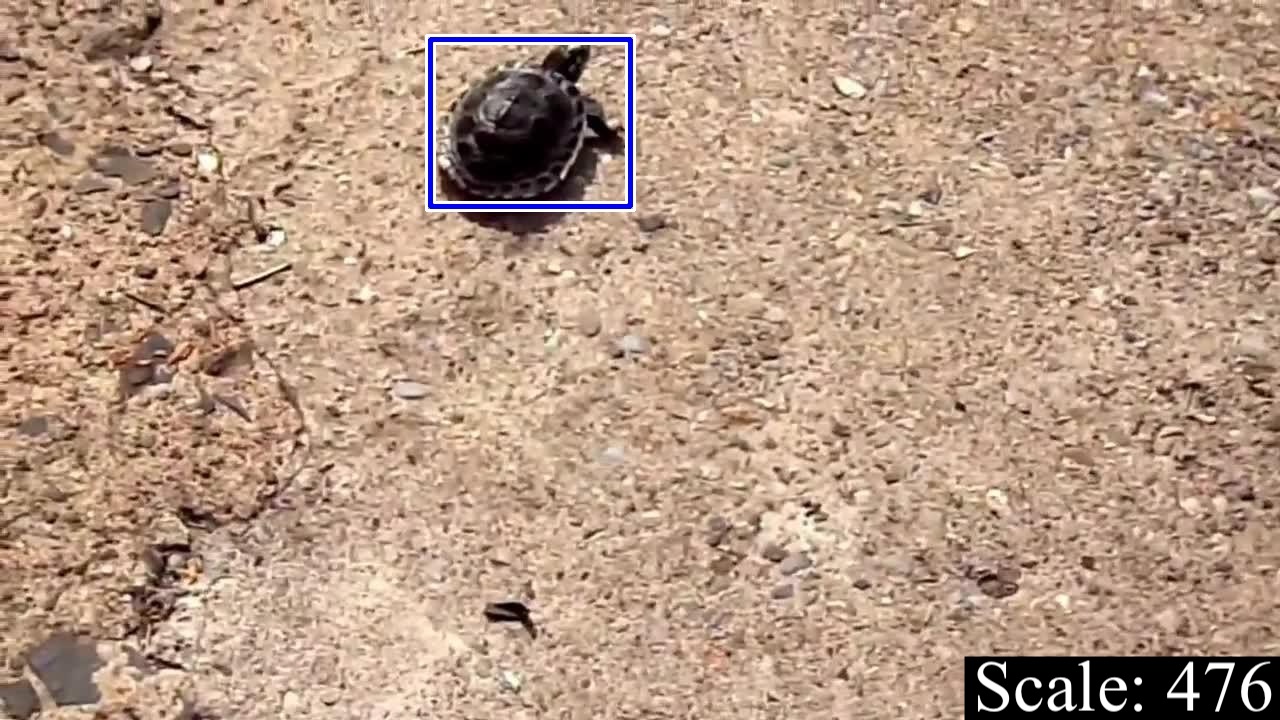} & 
    \includegraphics[width=.17\linewidth]{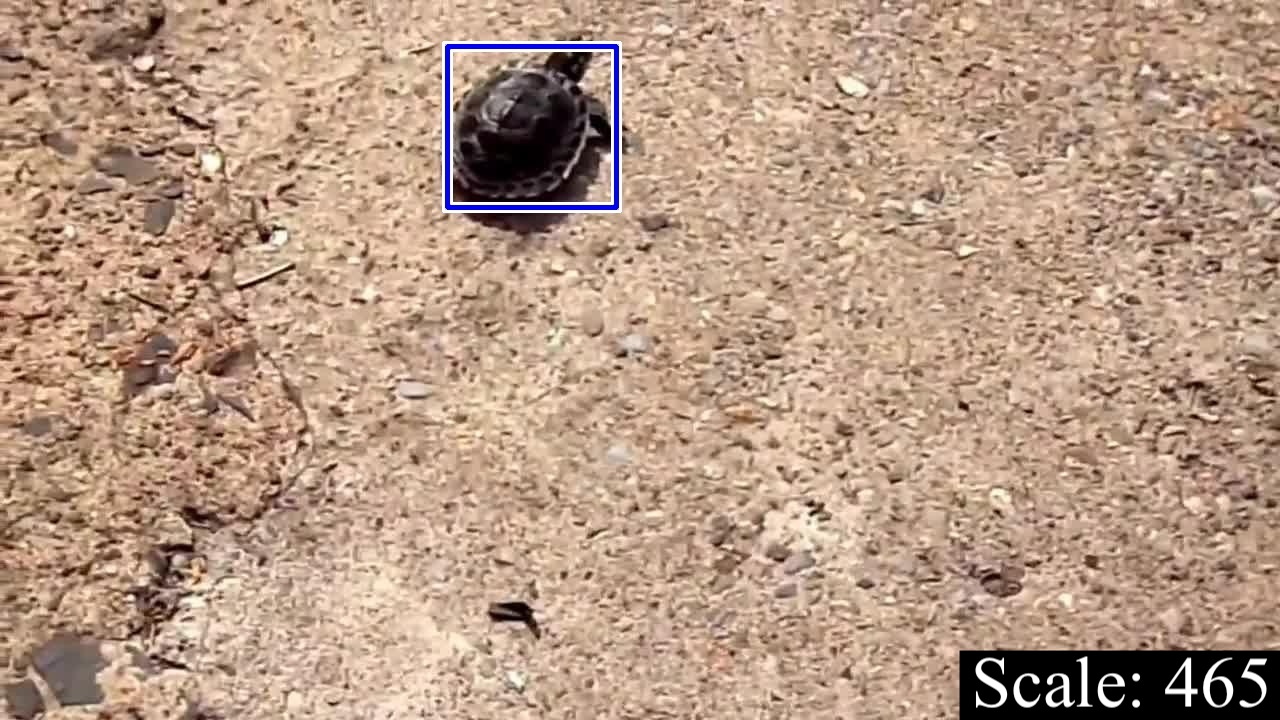} & 
    \includegraphics[width=.17\linewidth]{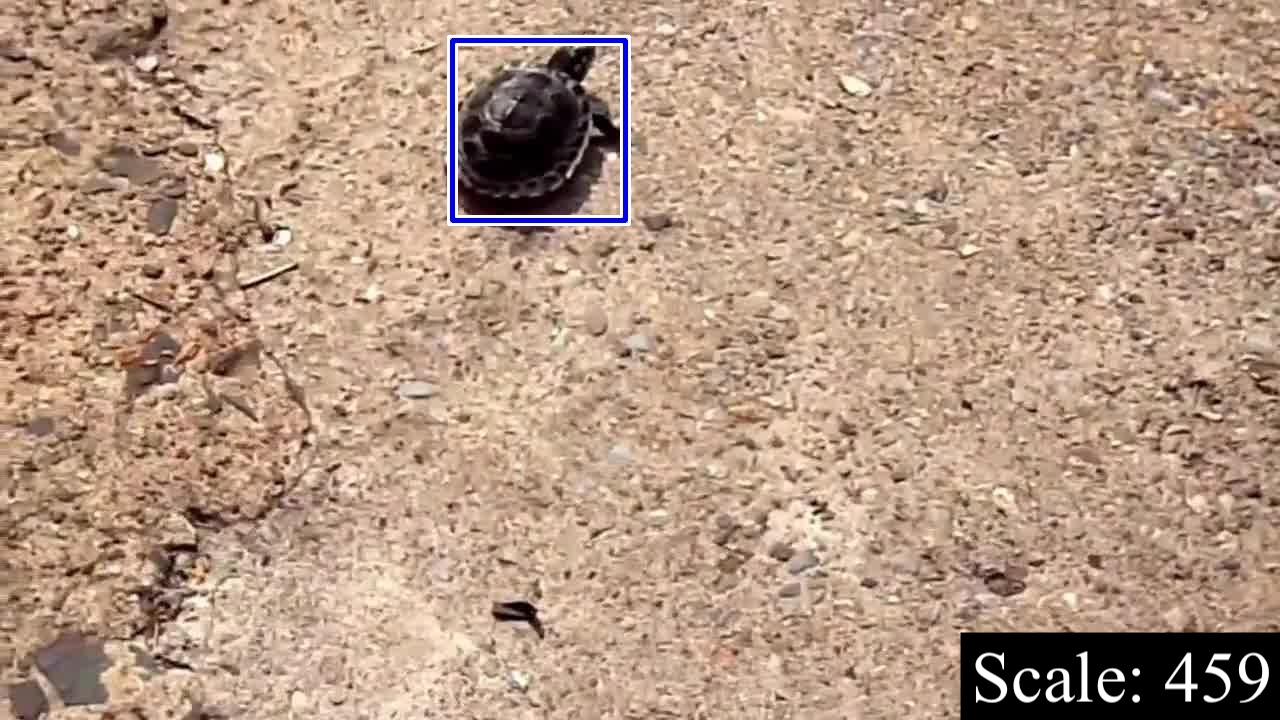} \\
    \includegraphics[width=.17\linewidth]{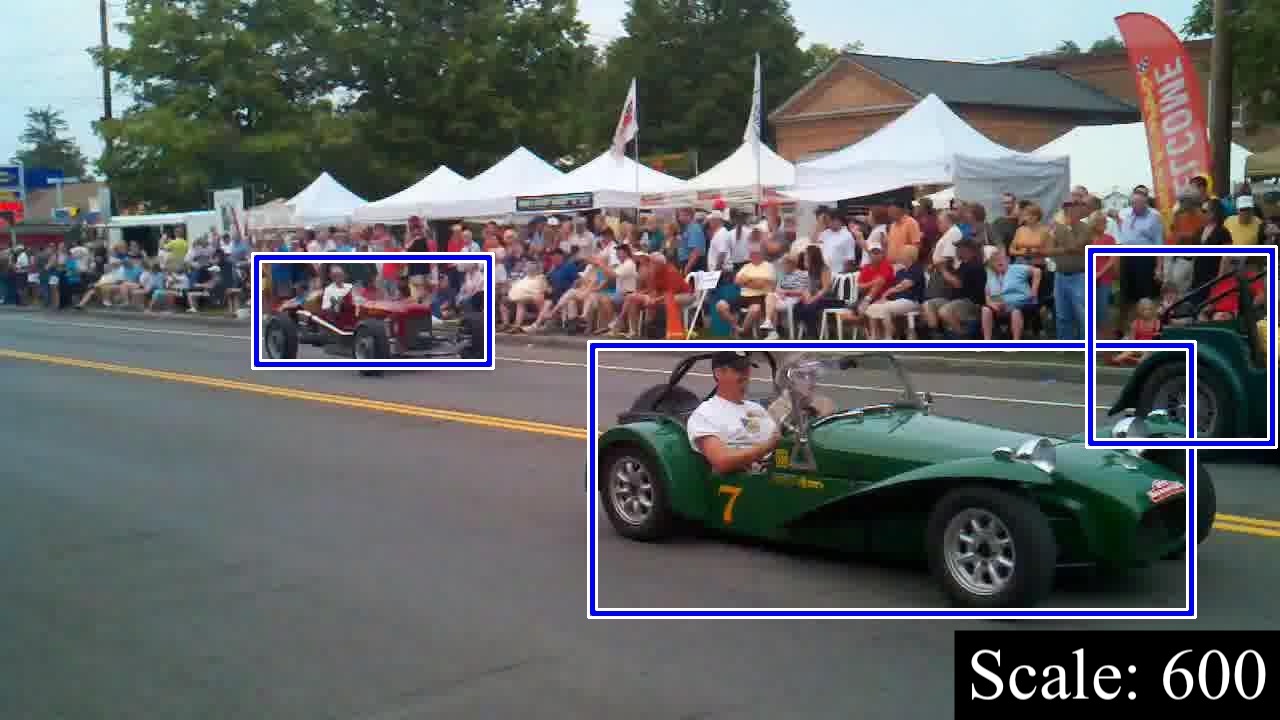} &
    \includegraphics[width=.17\linewidth]{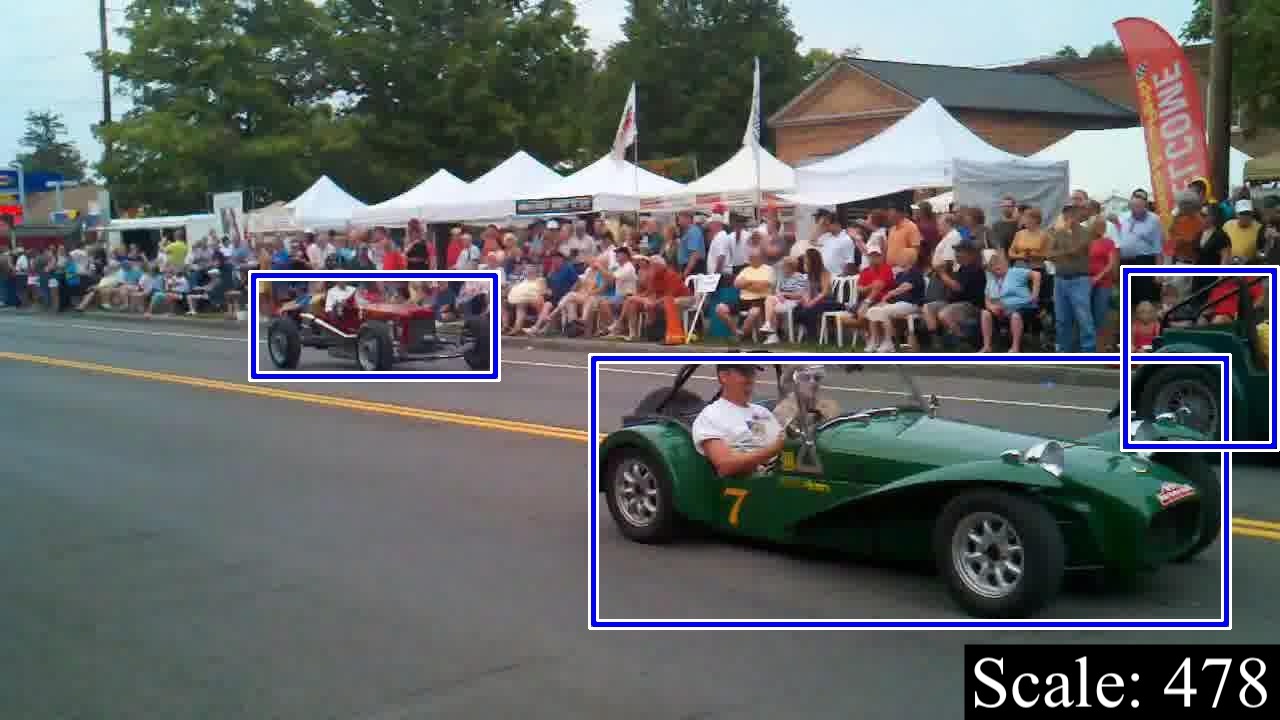} &
    \includegraphics[width=.17\linewidth]{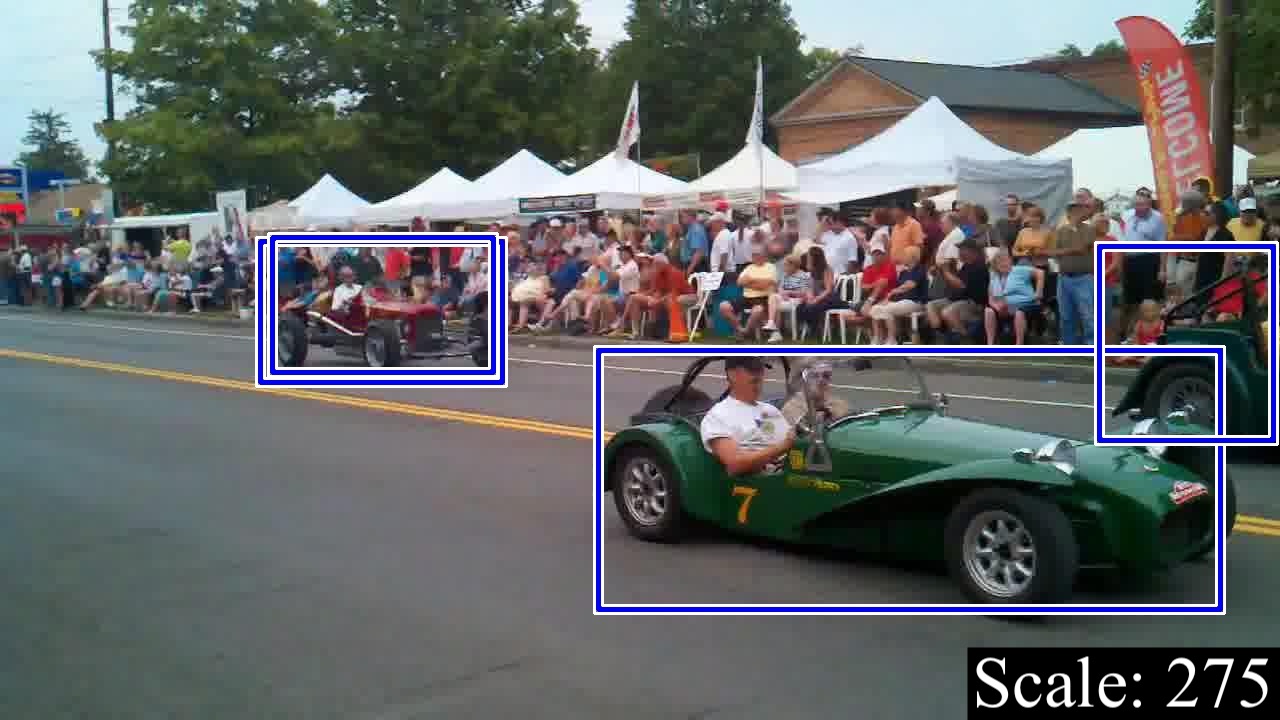} &
    \includegraphics[width=.17\linewidth]{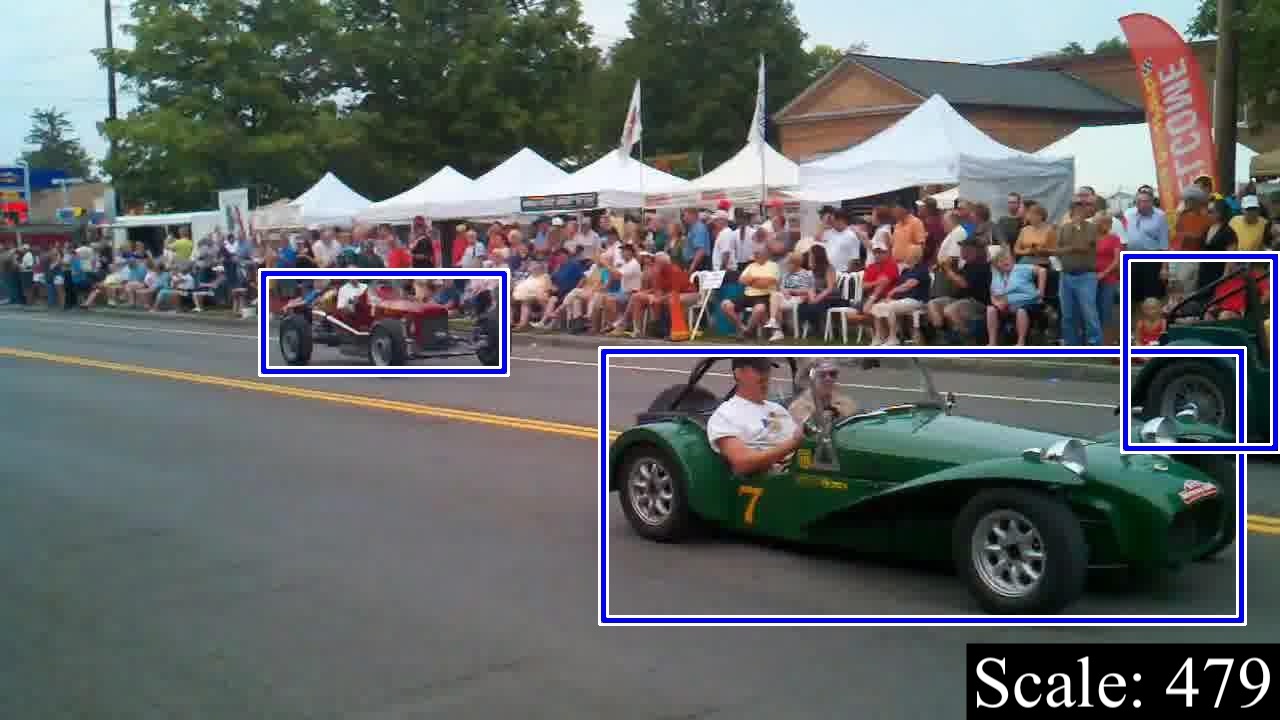} &
    \includegraphics[width=.17\linewidth]{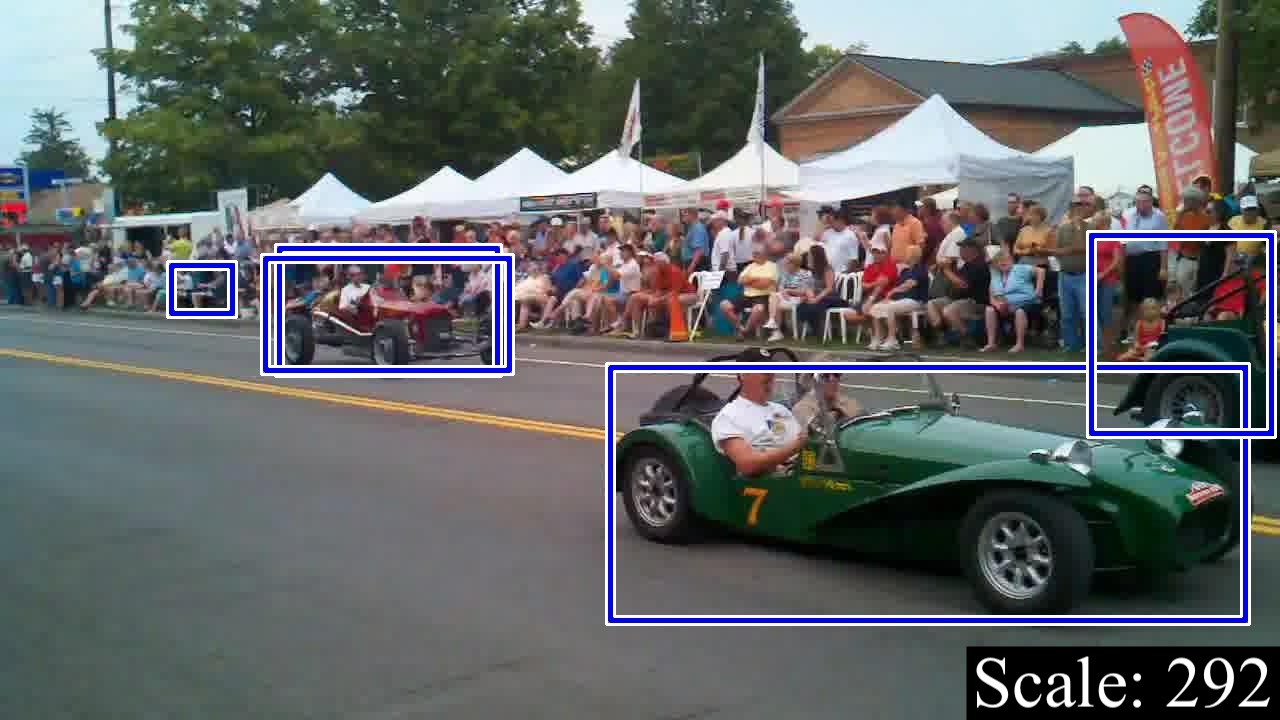} \\
    \end{tabular}
    \caption{The investigation of the dynamics of AdaScale. The scales of the images are labeled in bottom-right.}
    \label{fig:robust}
\end{figure*}

To understand AdaScale more in terms of the sequential decisions, Fig.~\ref{fig:robust} shows the  AdaScale dynamics of three clips. Specifically, it shows that (i) it stably down-samples images with a large object; (ii) it stably scales the images into larger scales when the object is small; and (iii) it jitters when there are multiple objects with varying sizes. The scale jittering in the third clip indicates that if there are size-varying multiple objects in the frame, it is harder to decide what constitutes a better size, which can also be observed in the watercraft of Fig.~\ref{fig:qual}. To enhance the current design, it is possible to apply AdaScale recursively on the attention of the given image, to obtain results from multiple regressed scales. We leave improvements of the current design to future work.

\subsection{Comparison with Prior Work}
\label{sec:combineprior}
To our best knowledge, our work is the first to exploit the use of images with smaller scales for improving both speed and accuracy, rather than treating them as a trade-off~\cite{huang2017speed,dai16rfcn,redmon2016yolo9000,chin18approx,lin2017focal}. For video object detection, our work is complementary to some of the prior work that tries to benefit from the detection results of multiple frames to improve accuracy or speed. 

In Fig.~\ref{fig:prior}, the baseline object detector is R-FCN~\cite{dai16rfcn} with 74.2 mAP and 13.3 frame-per-second (FPS). We run the prior work approaches~\cite{zhu2017flow,zhu2017deep,han2016seq,feichtenhofer2017detect} that provide source code for our experiment setup to profile both speed and mAP. Additionally, we combine our work with SeqNMS~\cite{han2016seq} and Deep Feature Flow (DFF)~\cite{zhu2017deep} to further push the Pareto frontier by maintaining the accuracy while speeding up testing by an additional 61\% and 25\%, respectively.

\subsection{Ablation Study}
\label{sec:ablation}

\begin{table*}[ht]
\begin{center}
\caption{mAP and runtime for different multi-scale training settings.}
\label{table:multi-scales}
\begin{tabular}{|c||c|c||c|c||c|c||c|c|}
\hline
$S_{train}$ & \multicolumn{2}{c||}{\{600,480,360,240\}}  & \multicolumn{2}{c||}{\{600,480,360\}} & \multicolumn{2}{c||}{\{600,360\}} &
\multicolumn{2}{c|}{\{600\}} \\
\hline
testing method & SS & Ada.
& SS & Ada.
& SS & Ada. 
& SS & Ada. \\
\hline
mAP (\%) & 73.3 & 75.5 & 73.3 & 74.8 & 73.4 & 74.8 & 74.2 & 74.2 \\
\hline
runtime (ms) & 75 & 47 & 75 & 55 & 75 & 57 & 75 & 68 \\
\hline
\end{tabular}
\end{center}
\end{table*}

\begin{figure*}[ht]
    \vspace{-12pt}
    \centering
    \subfloat[][\{600,480,360,240\}]{\includegraphics[width=.24\linewidth]{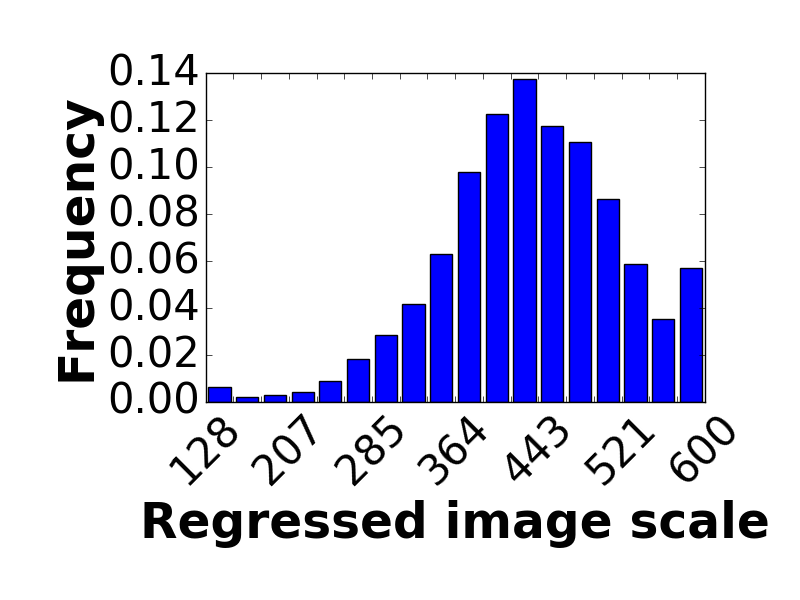}\label{fig:histogram:4}}
    \subfloat[][\{600,480,360\}]{\includegraphics[width=.24\linewidth]{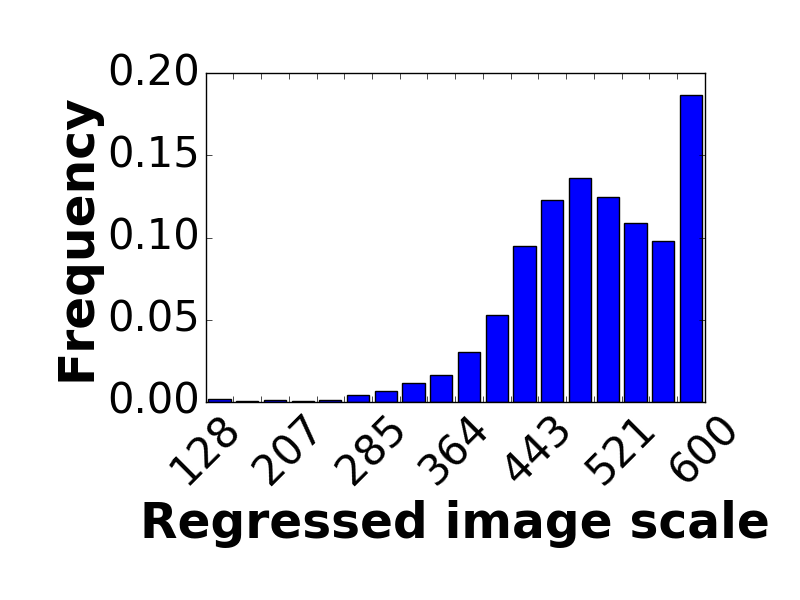}\label{fig:histogram:3}}
    \subfloat[][\{600,360\}]{\includegraphics[width=.24\linewidth]{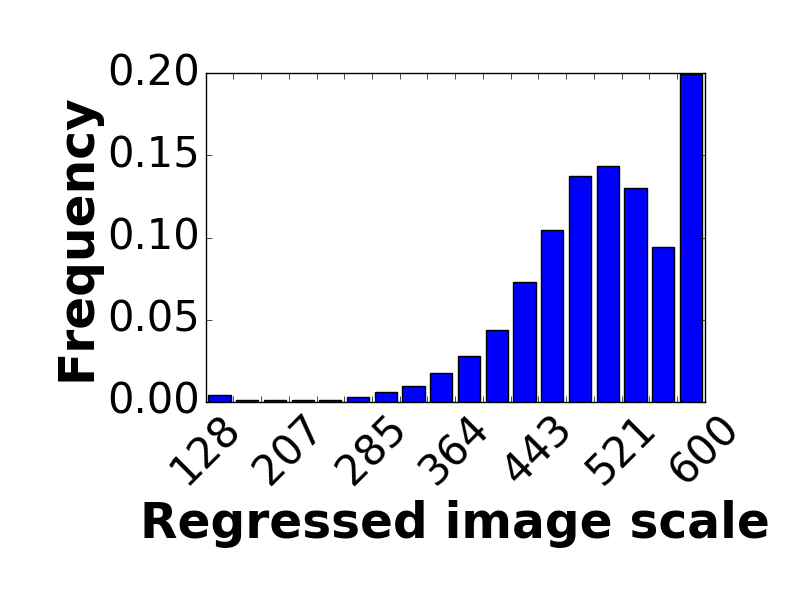}\label{fig:histogram:2}}
    \subfloat[][\{600\}]{\includegraphics[width=.24\linewidth]{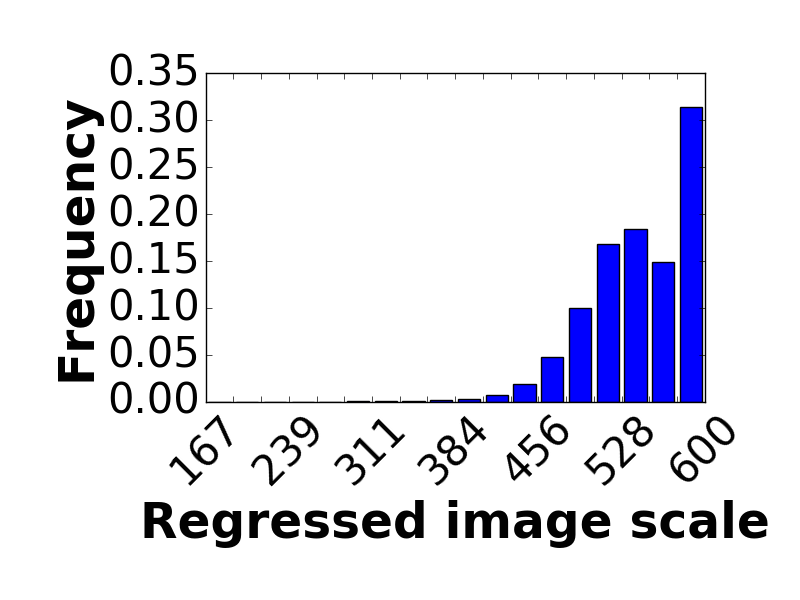}\label{fig:histogram:1}}
    \caption{The regressed scale distribution of AdaScale tested on ImageNet VID validation set. (a)-(d) use different $S_{train}$.}
    \label{fig:histogram}
\end{figure*}

\textit{Training Scales of Object Detector:} To understand how multi-scale training affects the performance of AdaScale, we try different sets of training scales $S_{train}$ and the results are shown in Table~\ref{table:multi-scales} and Fig.~\ref{fig:histogram}. We find that a larger set of $S_{train}$ improves both the mAP and speed of AdaScale. From Fig.~\ref{fig:histogram}(a)-(d), we can also observe higher speed with smaller \textit{training} scales. We postulate that it is due to two reasons: (i) Multi-scale trained object detector is able to generate more meaningful labels for the regressor to learn since it is less biased toward some scales. (ii) The object detector becomes good at multiple scales that could be better exploited by the scale regressor.

\begin{table}[ht]
\vspace{-6pt}
\centering
\caption{mAP and runtime for different regressor architectures.}
\label{table:regressor_arch}
\begin{tabular}{|c||c|c|c|}
    \hline
    kernel size & 1 & 1\&3 & 1\&3\&5\\
    \hline
    mAP (\%) & 75.3 & 75.5 & 75.5\\
    \hline
    runtime (ms) & 51 & 47 & 50\\
    \hline
\end{tabular}
\vspace{-6pt}
\end{table}

\textit{Regressor Architectures:} We try using different sizes of filter for the regressor module and we show the results in Table~\ref{table:regressor_arch}. Interestingly, since the \textit{accuracy} of the regressor directly affects the \textit{speed} of the object detector, both regressor's accuracy and the overhead of the module affect the final overall speed.

\section{Conclusion}
\label{sec:conclusion}

Given the importance of video object detection, we present a thorough study of the possibility of improving both speed and accuracy in video object detection with adaptive scaling. Our contributions are three-fold: (i) to the best of our knowledge, our work is the first work to demonstrate the use of down-sampled images for improving both speed and accuracy for video object detection, (ii) we provide comprehensive empirical results that demonstrate improvement in both ImageNet VID as well as mini YouTube-BB datasets, and (iii) we combine our technique with state-of-the-art video object detection acceleration techniques and further improve the speed by an additional 25\% with the added benefit of slightly higher accuracy.

\section*{Acknowledgements}
This research was supported in part by National Science Foundation CSR Grant No. 1815780 and National Science Foundation CCF Grant No. 1815899. This work used the Extreme Science and Engineering Discovery Environment (XSEDE), which is supported by National Science Foundation grant number ACI-1548562~\cite{Nystrom:2015:BUF:2792745.2792775}. Specifically, it used the Bridges system, which is supported by National Science Foundation award number ACI-1445606, at the Pittsburgh Supercomputing Center (PSC).

\bibliography{main}
\bibliographystyle{sysml2019}

\clearpage

\section{Supplementary Materials}
\begin{figure*}[h!]
\renewcommand*\thesubfigure{\arabic{subfigure}}
    \subfloat[1][airplane]{\includegraphics[width=.33\linewidth]{airplane.png}\label{fig:prcurve_all_airplane}}
    \subfloat[][antelope]{\includegraphics[width=.33\linewidth]{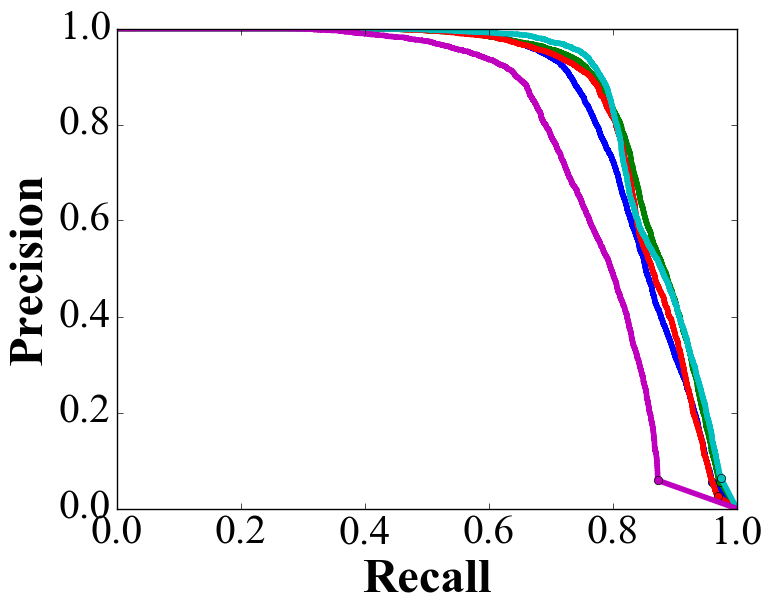}\label{fig:prcurve_all_antelope}}
    \subfloat[][bear]{\includegraphics[width=.33\linewidth]{bear.png}\label{fig:prcurve_all_bear}}\\
    \subfloat[][bicycle]{\includegraphics[width=.33\linewidth]{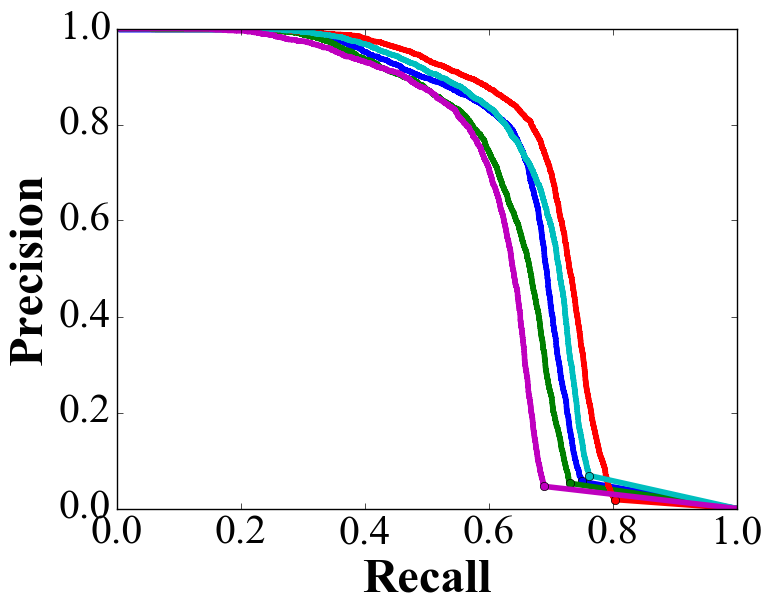}\label{fig:prcurve_all_bicycle}}
    \subfloat[][bird]{\includegraphics[width=.33\linewidth]{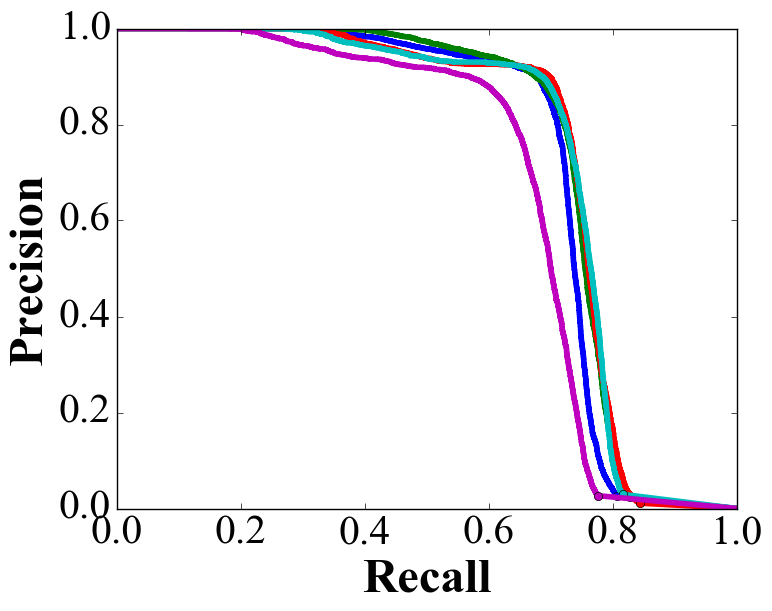}\label{fig:prcurve_all_bird}}
    \subfloat[][bus]{\includegraphics[width=.33\linewidth]{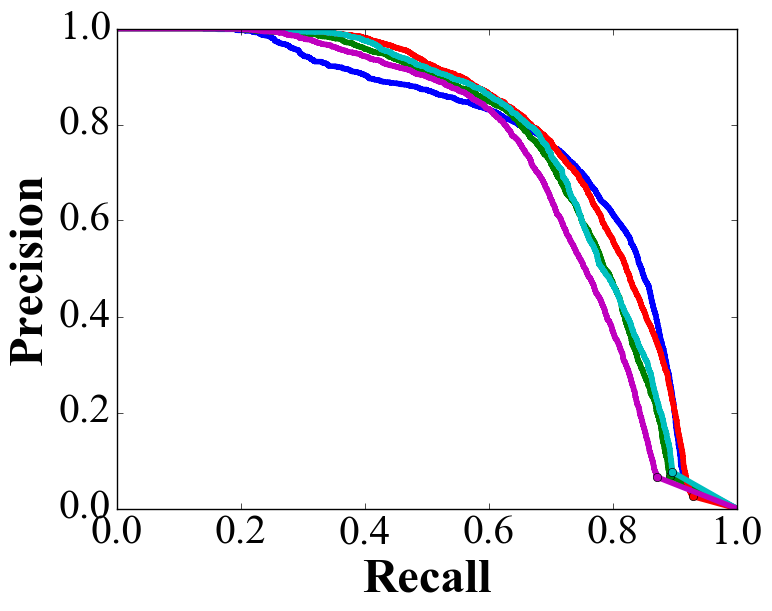}\label{fig:prcurve_all_bus}}\\
    \subfloat[][car]{\includegraphics[width=.33\linewidth]{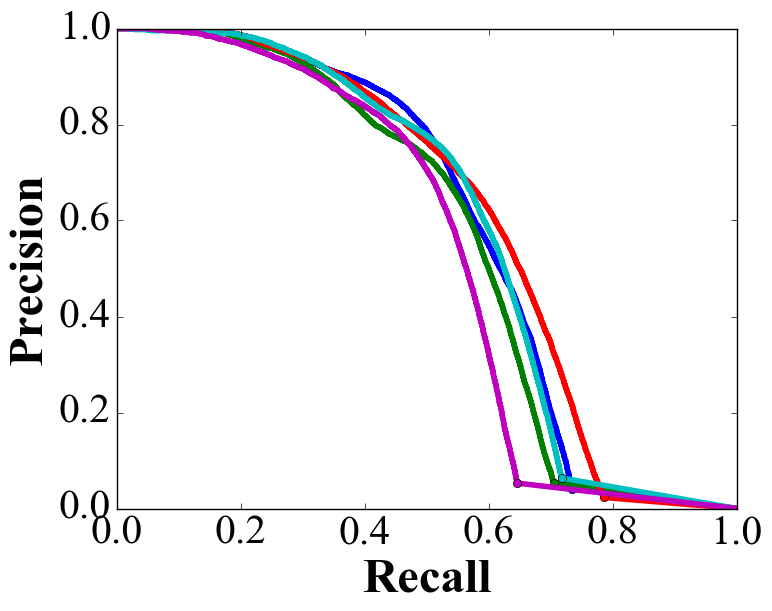}\label{fig:prcurve_all_car}}
    \subfloat[][cattle]{\includegraphics[width=.33\linewidth]{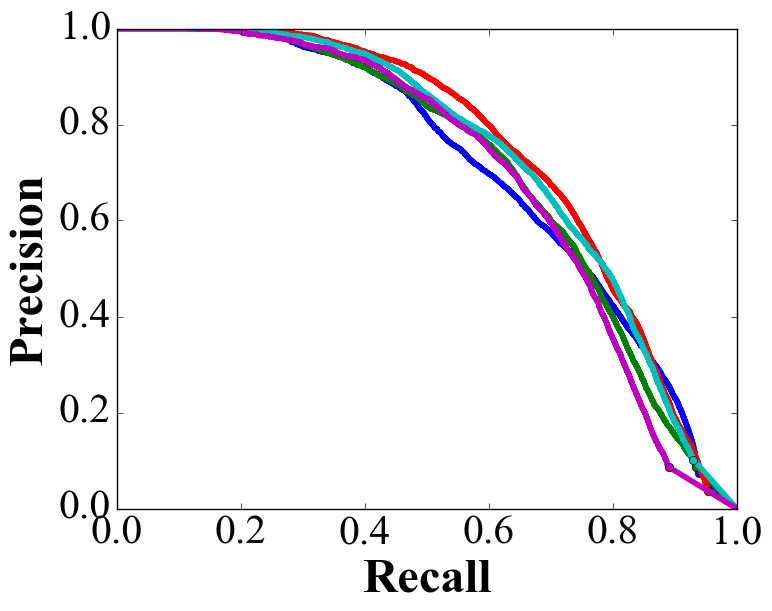}\label{fig:prcurve_all_cattle}}
    \subfloat[][dog]{\includegraphics[width=.33\linewidth]{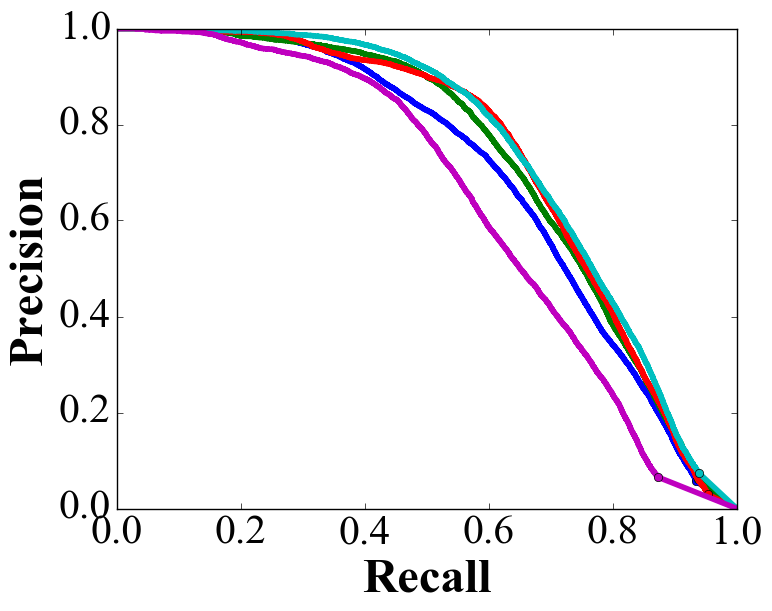}\label{fig:prcurve_all_dog}}\\
\end{figure*}

\begin{figure*}[h]\ContinuedFloat
\renewcommand*\thesubfigure{\arabic{subfigure}}
    \subfloat[][domestic\_cat]{\includegraphics[width=.33\linewidth]{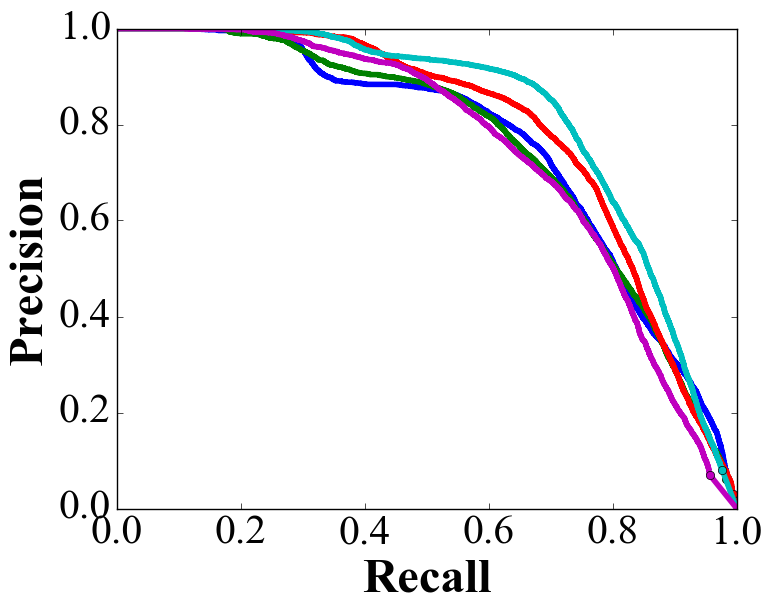}\label{fig:prcurve_all_domestic_cat}}
    \subfloat[][elephant]{\includegraphics[width=.33\linewidth]{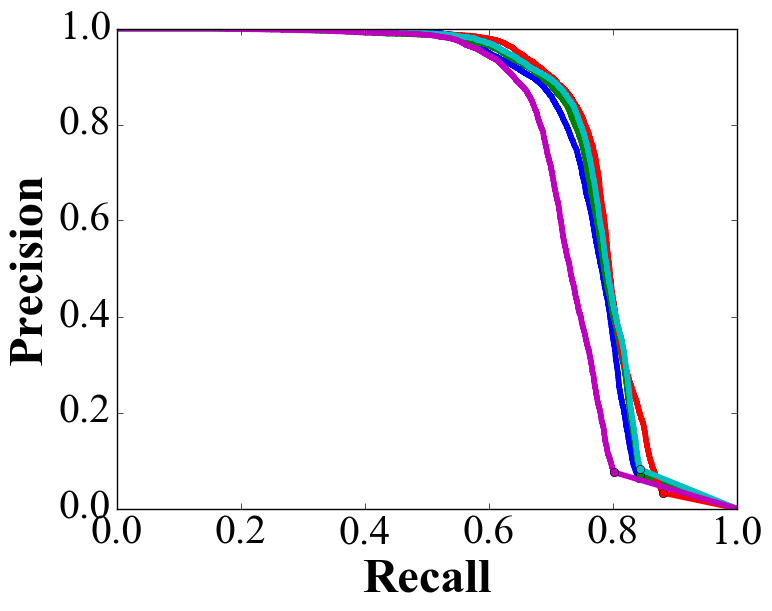}\label{fig:prcurve_all_elephant}}
    \subfloat[][fox]{\includegraphics[width=.33\linewidth]{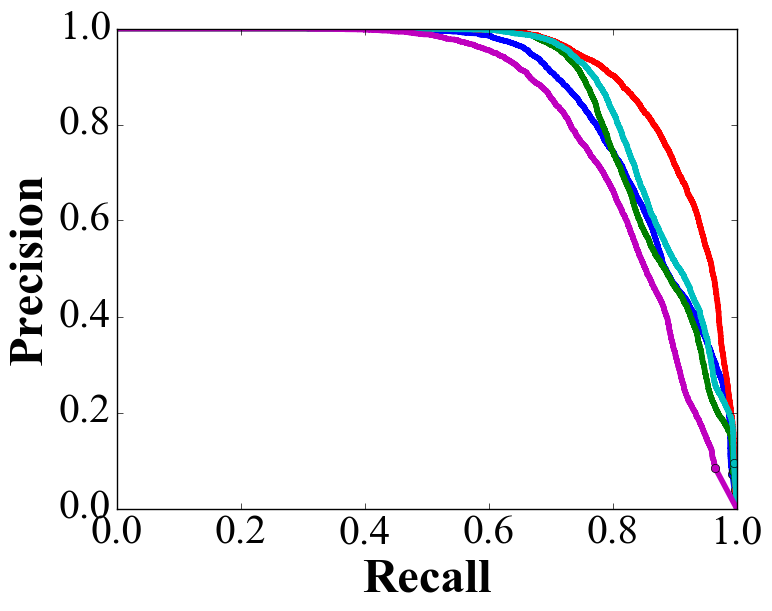}\label{fig:prcurve_all_fox}}\\
    \subfloat[][giant\_panda]{\includegraphics[width=.33\linewidth]{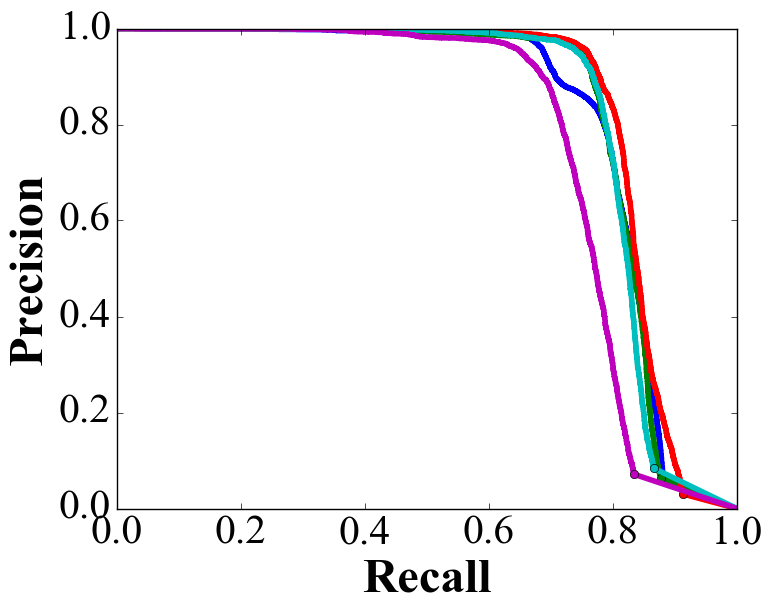}\label{fig:prcurve_all_giant_panda}}
    \subfloat[][hamster]{\includegraphics[width=.33\linewidth]{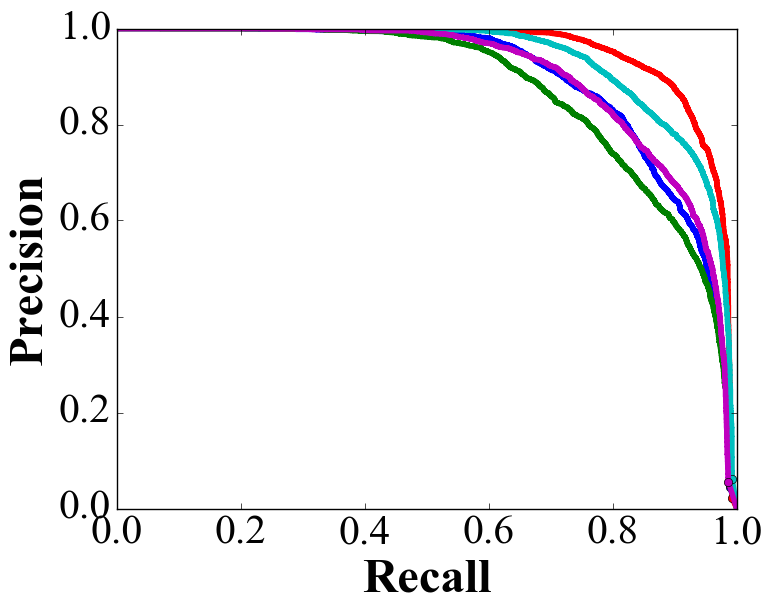}\label{fig:prcurve_all_hamster}}
    \subfloat[][horse]{\includegraphics[width=.33\linewidth]{horse.png}\label{fig:prcurve_all_horse}}\\
    \subfloat[][lion]{\includegraphics[width=.33\linewidth]{lion.png}\label{fig:prcurve_all_lion}}
    \subfloat[][lizard]{\includegraphics[width=.33\linewidth]{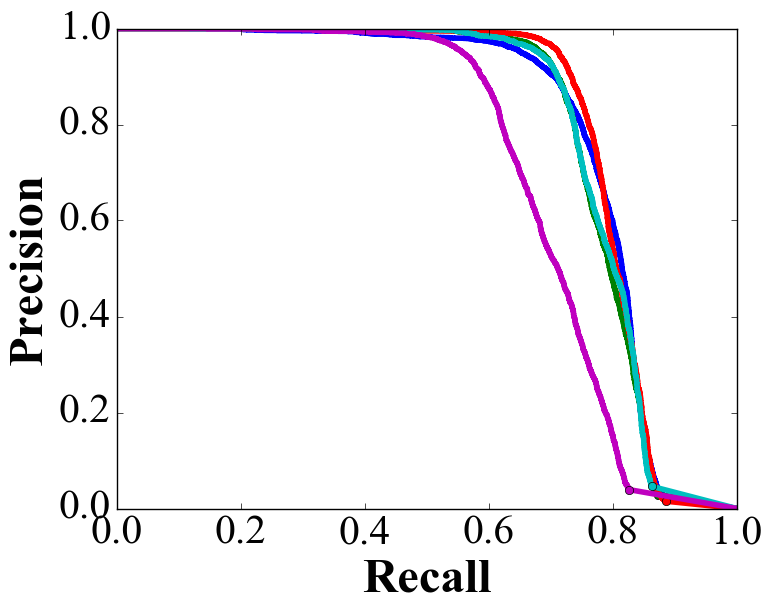}\label{fig:prcurve_all_lizard}}
    \subfloat[][monkey]{\includegraphics[width=.33\linewidth]{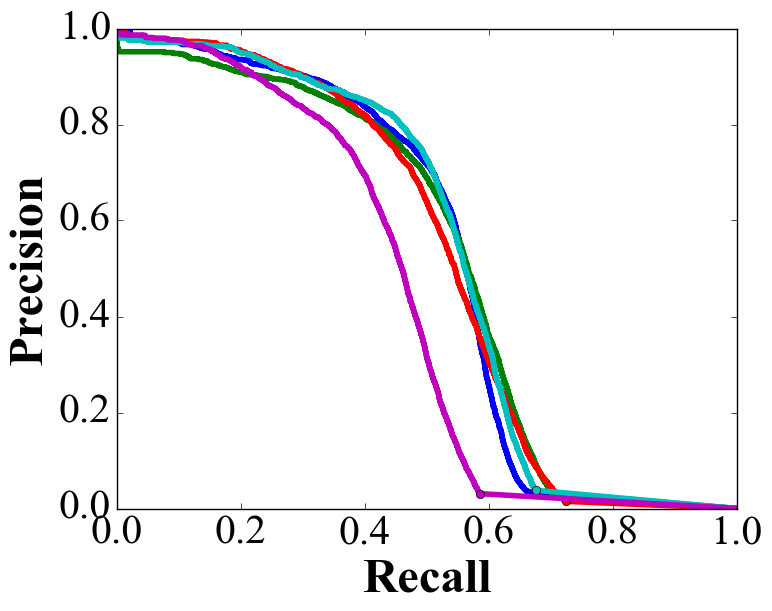}\label{fig:prcurve_all_monkey}}\\
    \subfloat[][motorcycle]{\includegraphics[width=.33\linewidth]{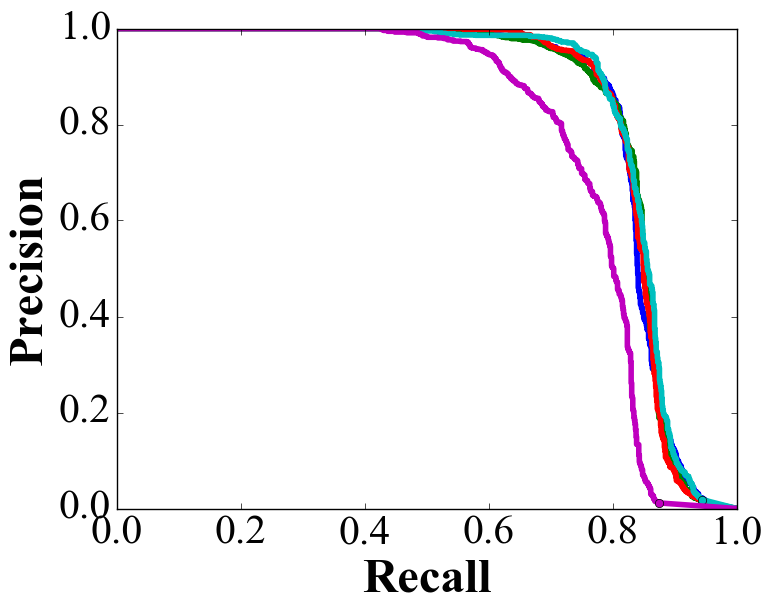}\label{fig:prcurve_all_motorcycle}}
    \subfloat[][rabbit]{\includegraphics[width=.33\linewidth]{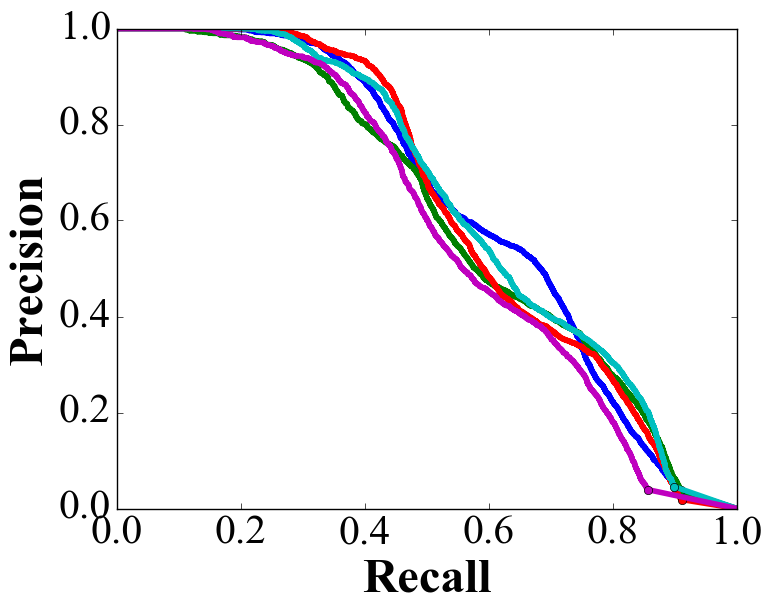}\label{fig:prcurve_all_rabbit}}
    \subfloat[][red\_panda]{\includegraphics[width=.33\linewidth]{red_panda.png}\label{fig:prcurve_all_red_panda}}\\
\end{figure*}

\begin{figure*}[h]\ContinuedFloat
\renewcommand*\thesubfigure{\arabic{subfigure}}
    \subfloat[][sheep]{\includegraphics[width=.33\linewidth]{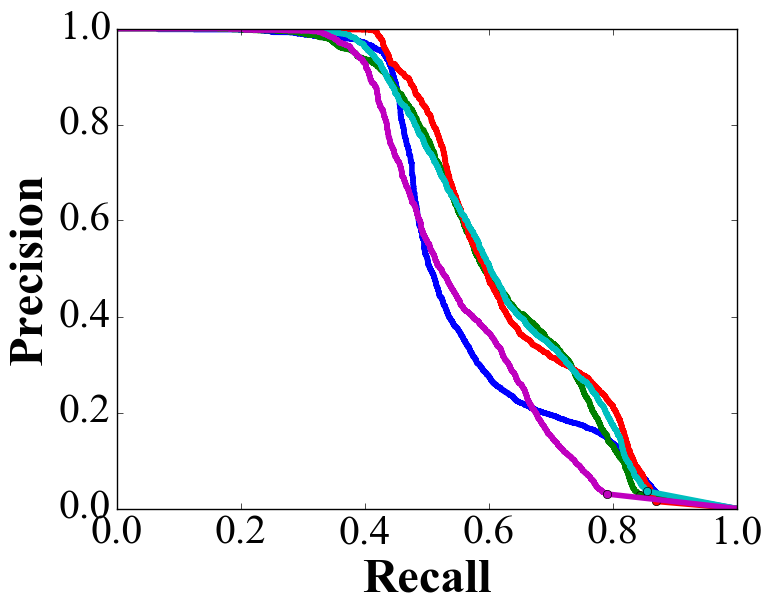}\label{fig:prcurve_all_sheep}}
    \subfloat[][snake]{\includegraphics[width=.33\linewidth]{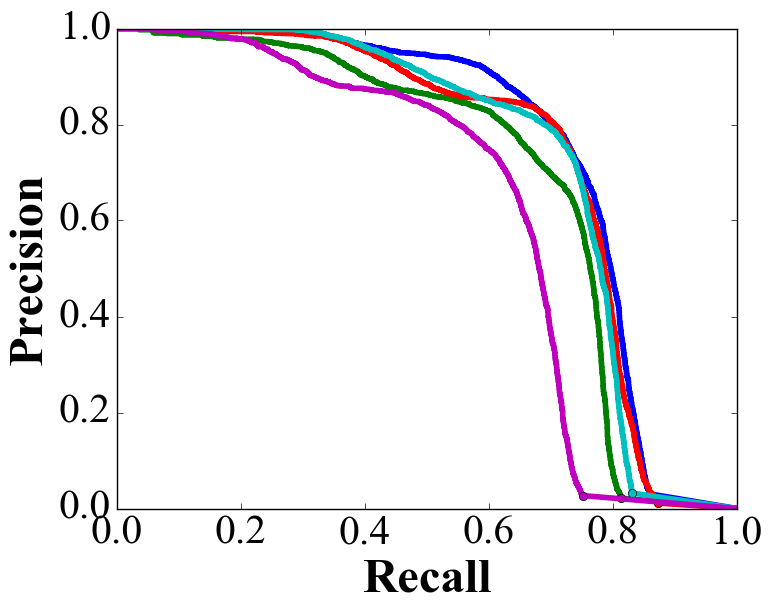}\label{fig:prcurve_all_snake}}
    \subfloat[][squirrel]{\includegraphics[width=.33\linewidth]{squirrel.png}\label{fig:prcurve_all_squirrel}}\\
    \subfloat[][tiger]{\includegraphics[width=.33\linewidth]{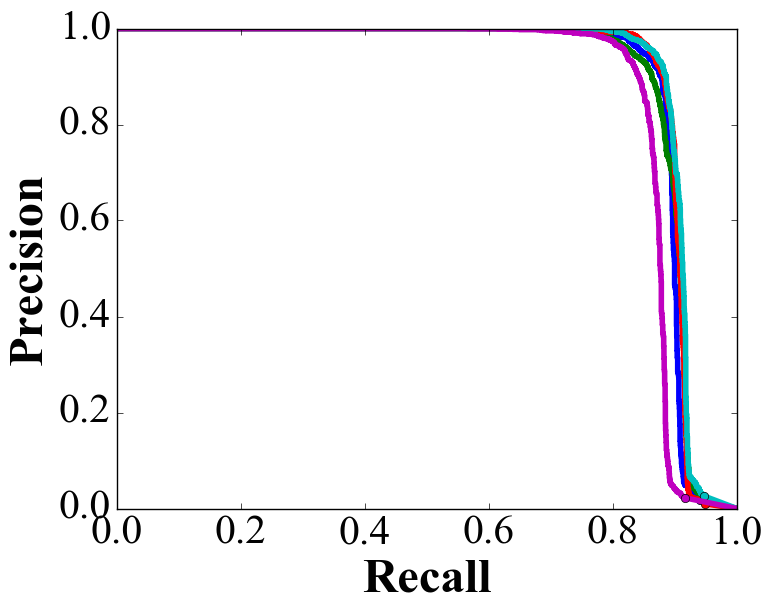}\label{fig:prcurve_all_tiger}}
    \subfloat[][train]{\includegraphics[width=.33\linewidth]{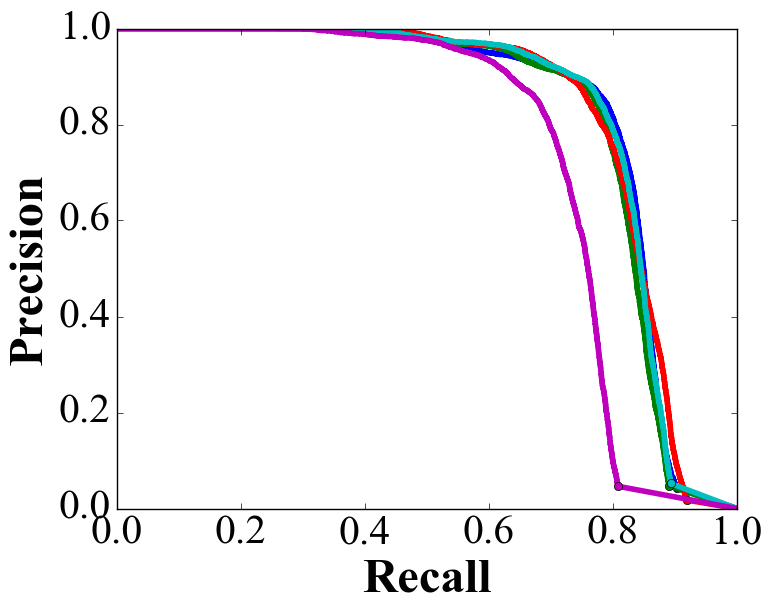}\label{fig:prcurve_all_train}}
    \subfloat[][turtle]{\includegraphics[width=.33\linewidth]{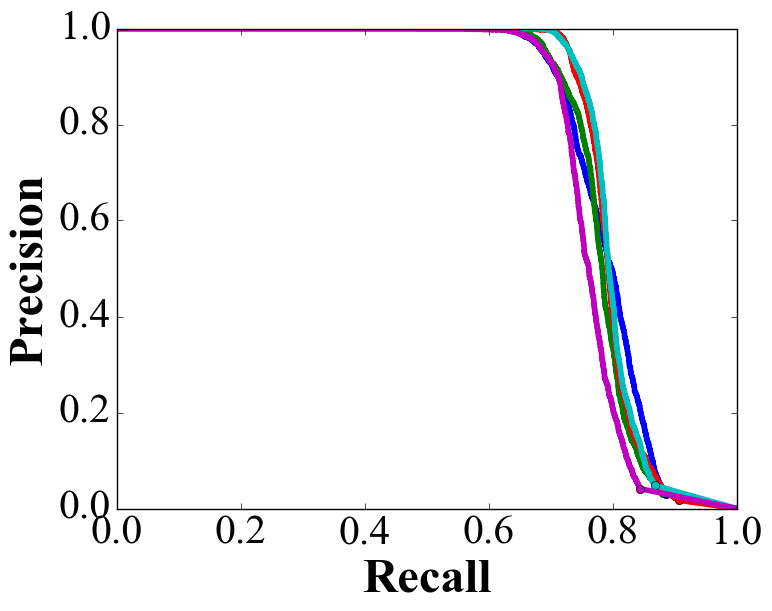}\label{fig:prcurve_all_turtle}}\\
    \subfloat[][watercraft]{\includegraphics[width=.33\linewidth]{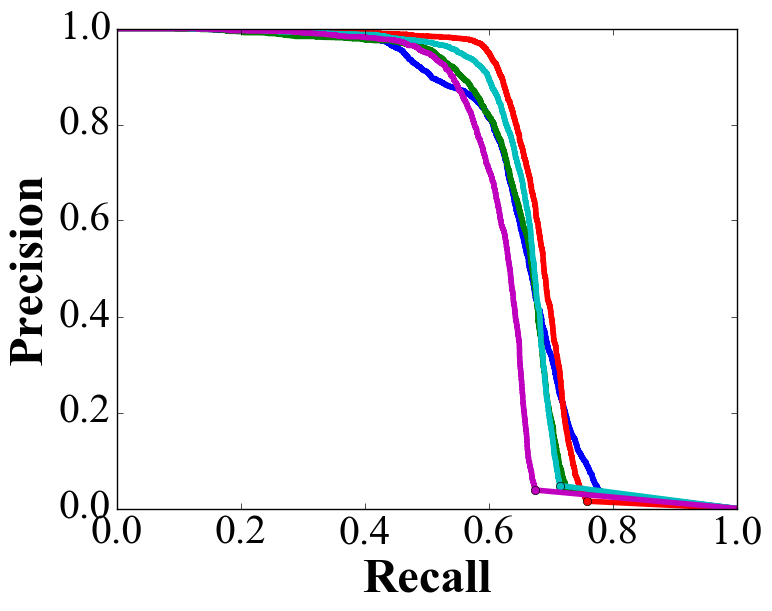}\label{fig:prcurve_all_watercraft}}
    \subfloat[][whale]{\includegraphics[width=.33\linewidth]{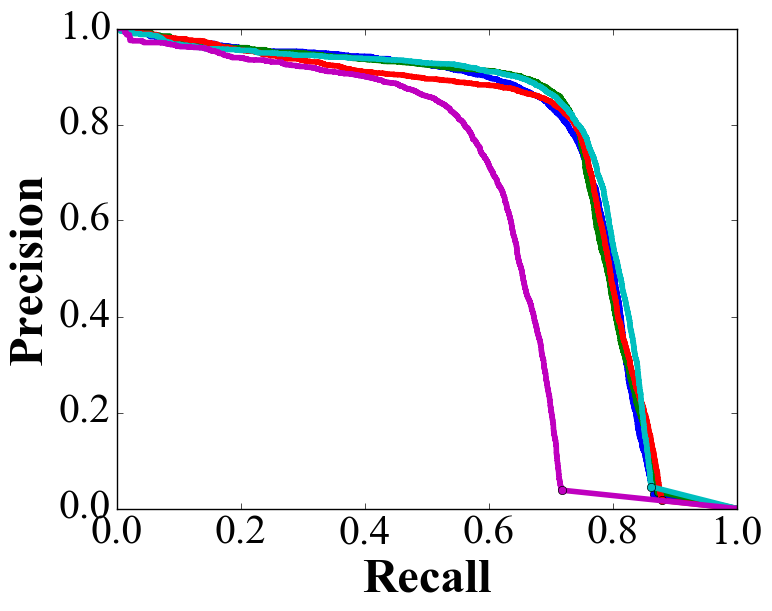}\label{fig:prcurve_all_whale}}
    \subfloat[][zebra]{\includegraphics[width=.33\linewidth]{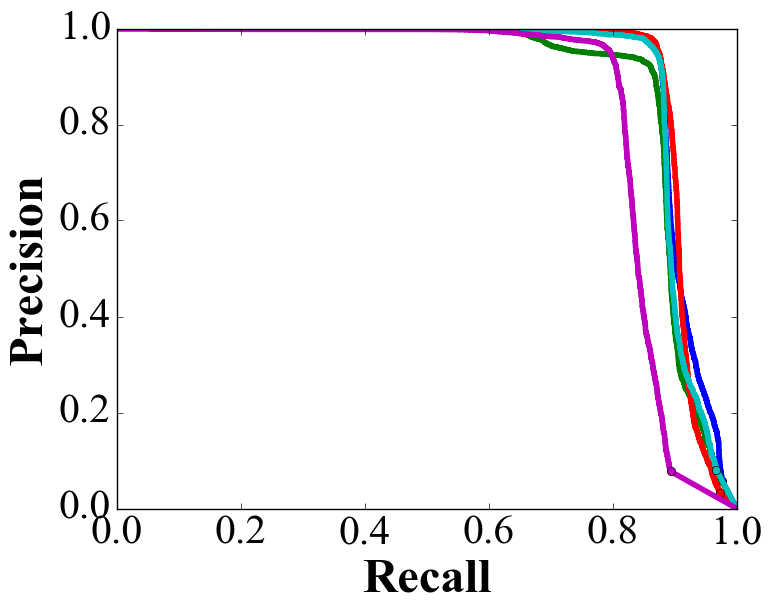}\label{fig:prcurve_all_zebra}}\\
    \begin{center}
    \includegraphics[width=0.7\linewidth]{prcurve.png}
    \vspace{-8pt}
    \end{center}
    \vspace{-8pt}
    \caption{Precision-Recall curves for 30 categories on ImageNet VID dataset}
    \label{fig:prcurve_all}
\end{figure*}

\clearpage
%\setcounter{tocdepth}{3}
%\subsubsection{True Positive \& False Positive}
%\setcounter{secnumdepth}{3}
%\setcounter{tocdepth}{3}

\begin{figure*}[h!]
\renewcommand*\thesubfigure{\arabic{subfigure}}
    \subfloat[][airplane]{\includegraphics[width=.33\linewidth]{tpfp_airplane.png}\label{fig:tpfp_all_airplane}}
    \subfloat[][antelope]{\includegraphics[width=.33\linewidth]{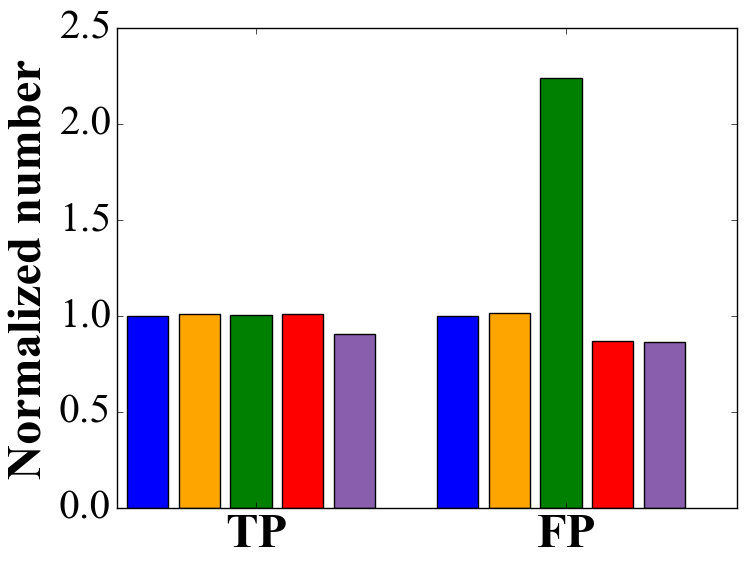}\label{fig:tpfp_all_antelope}}
    \subfloat[][bear]{\includegraphics[width=.33\linewidth]{tpfp_bear.png}\label{fig:tpfp_all_bear}}\\
    \subfloat[][bicycle]{\includegraphics[width=.33\linewidth]{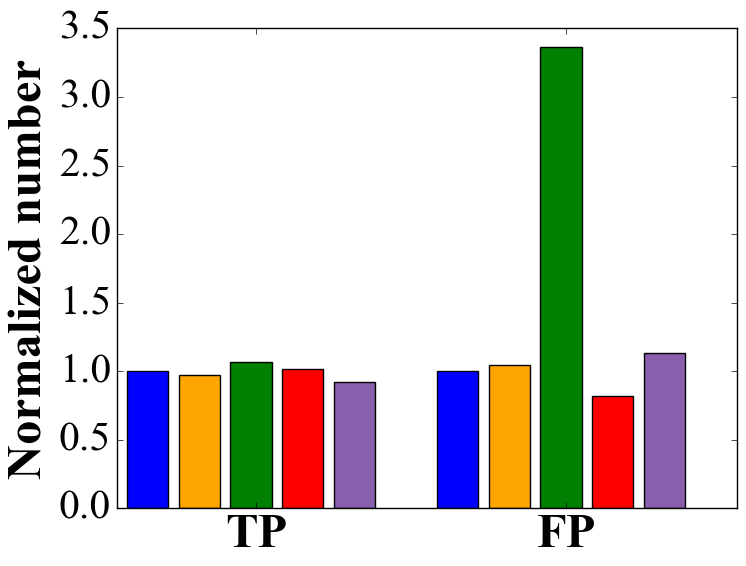}\label{fig:tpfp_all_bicycle}}
    \subfloat[][bird]{\includegraphics[width=.33\linewidth]{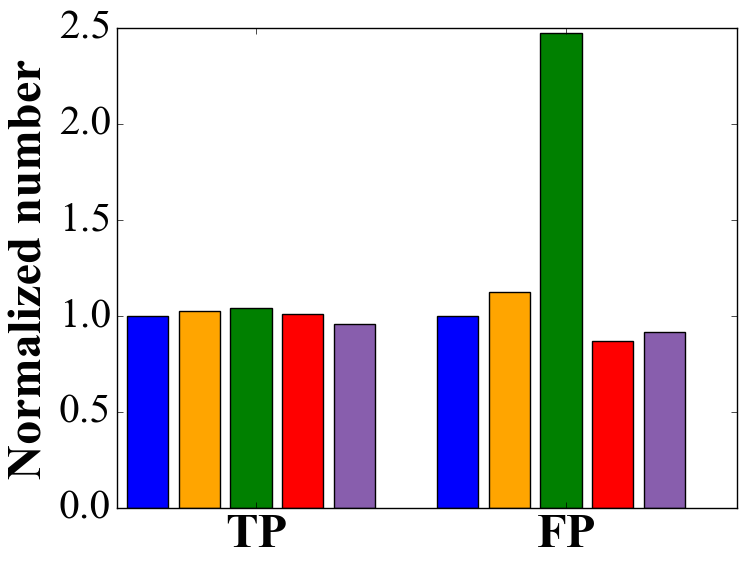}\label{fig:tpfp_all_bird}}
    \subfloat[][bus]{\includegraphics[width=.33\linewidth]{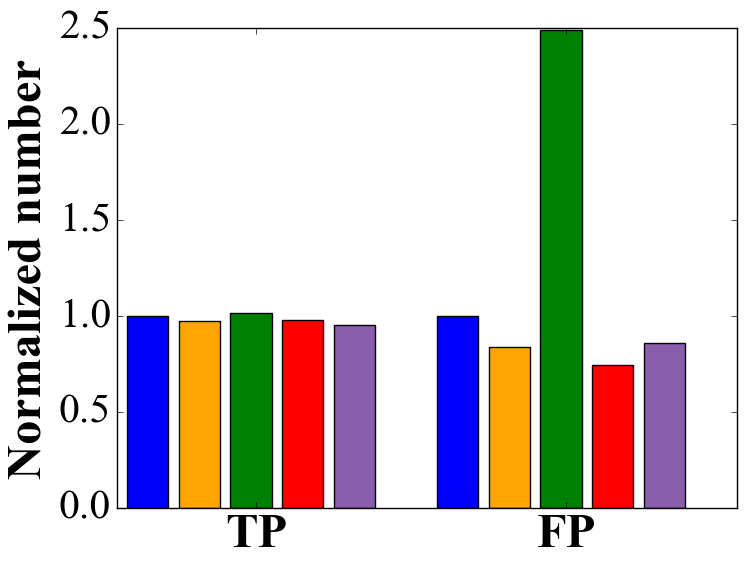}\label{fig:tpfp_all_bus}}\\
    \subfloat[][car]{\includegraphics[width=.33\linewidth]{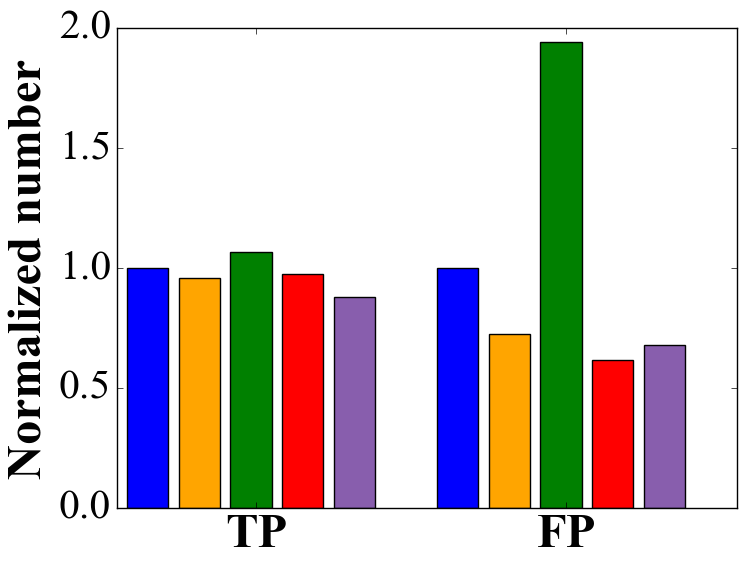}\label{fig:tpfp_all_car}}
    \subfloat[][cattle]{\includegraphics[width=.33\linewidth]{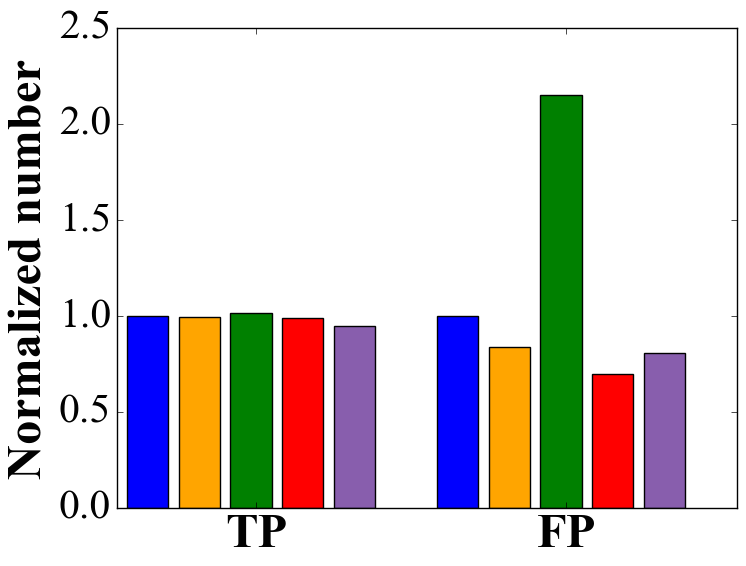}\label{fig:tpfp_all_cattle}}
    \subfloat[][dog]{\includegraphics[width=.33\linewidth]{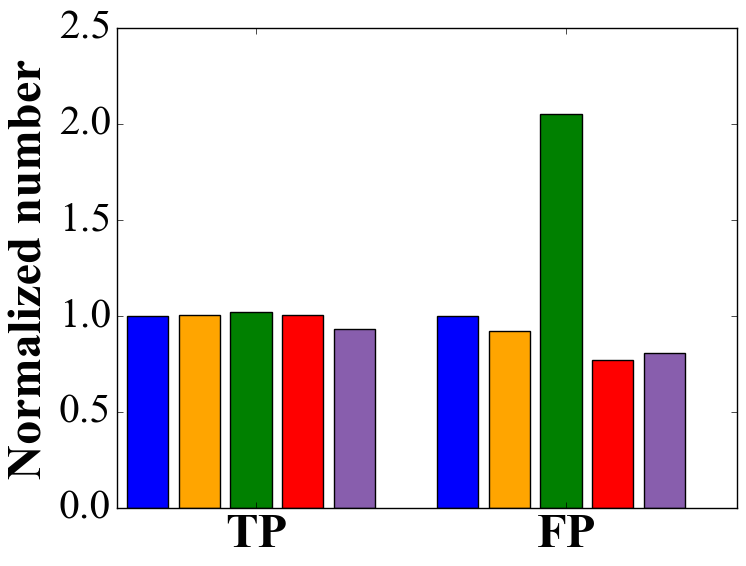}\label{fig:tpfp_all_dog}}\\
\end{figure*}

\begin{figure*}[h]\ContinuedFloat
\renewcommand*\thesubfigure{\arabic{subfigure}}
    \subfloat[][domestic\_cat]{\includegraphics[width=.33\linewidth]{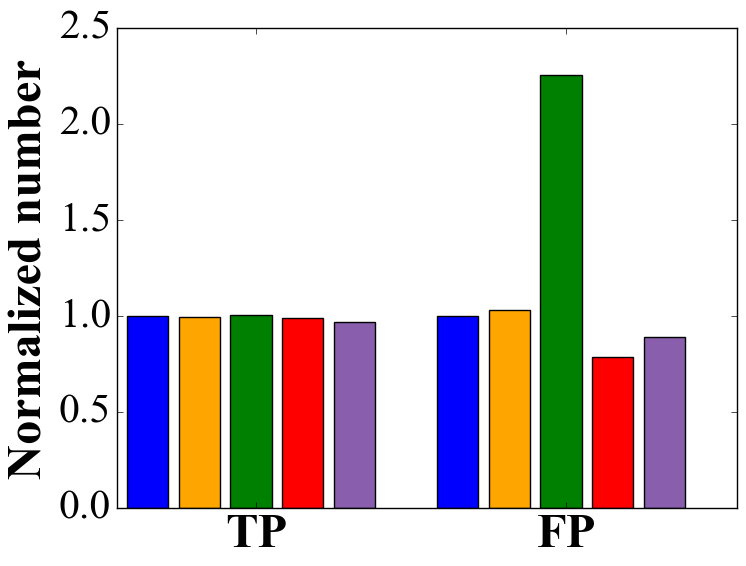}\label{fig:tpfp_all_domestic_cat}}
    \subfloat[][elephant]{\includegraphics[width=.33\linewidth]{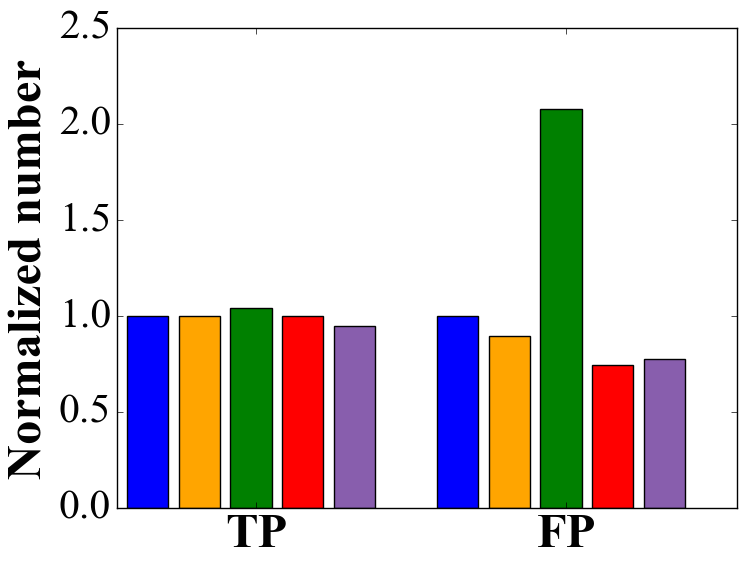}\label{fig:tpfp_all_elephant}}
    \subfloat[][fox]{\includegraphics[width=.33\linewidth]{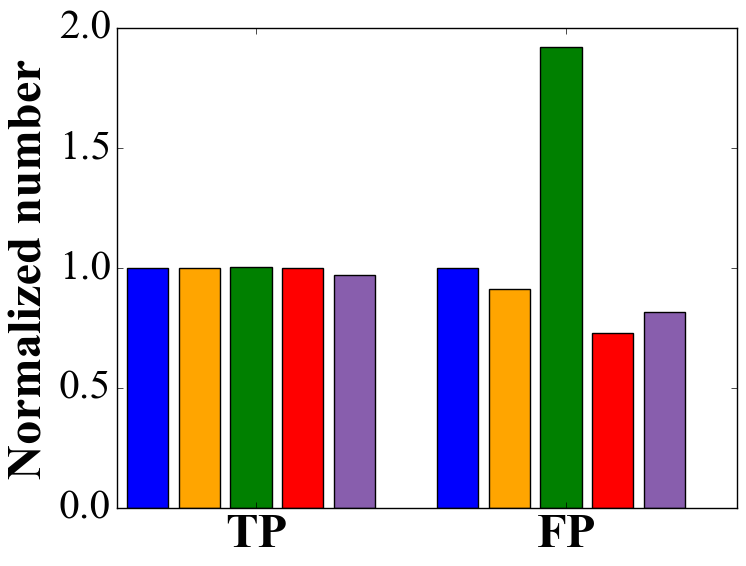}\label{fig:tpfp_all_fox}}\\
    \subfloat[][giant\_panda]{\includegraphics[width=.33\linewidth]{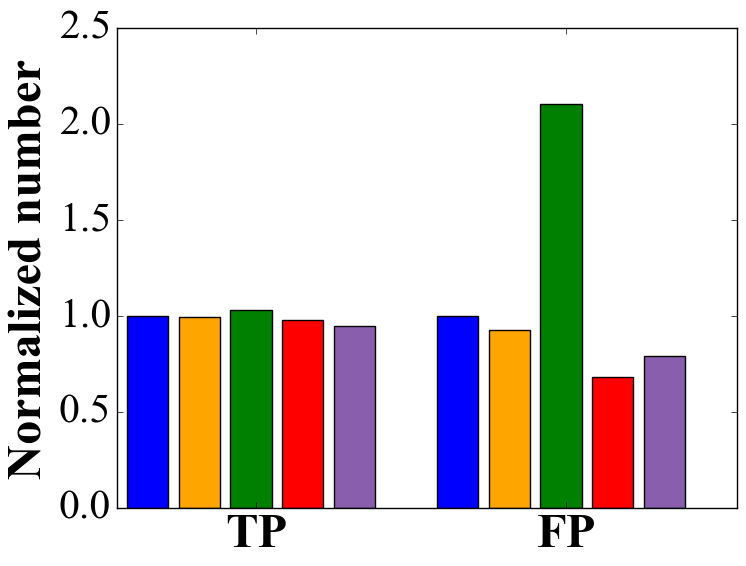}\label{fig:tpfp_all_giant_panda}}
    \subfloat[][hamster]{\includegraphics[width=.33\linewidth]{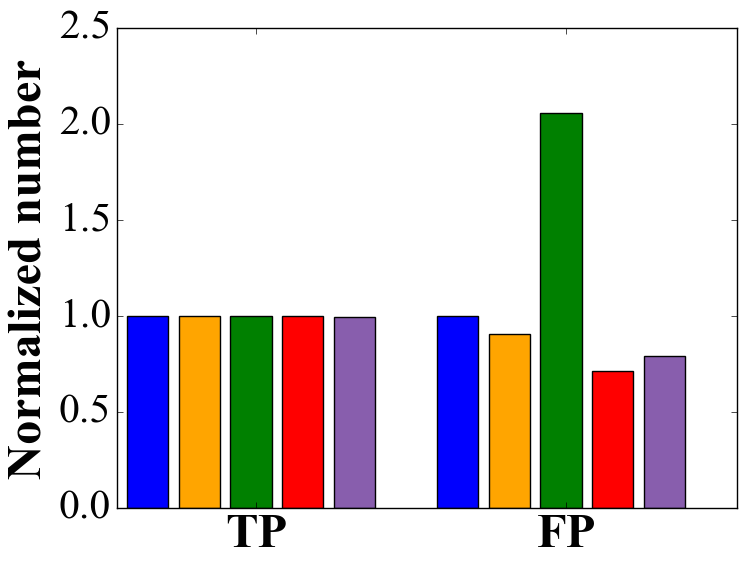}\label{fig:tpfp_all_hamster}}
    \subfloat[][horse]{\includegraphics[width=.33\linewidth]{tpfp_horse.png}\label{fig:tpfp_all_horse}}\\
    \subfloat[][lion]{\includegraphics[width=.33\linewidth]{tpfp_lion.png}\label{fig:tpfp_all_lion}}
    \subfloat[][lizard]{\includegraphics[width=.33\linewidth]{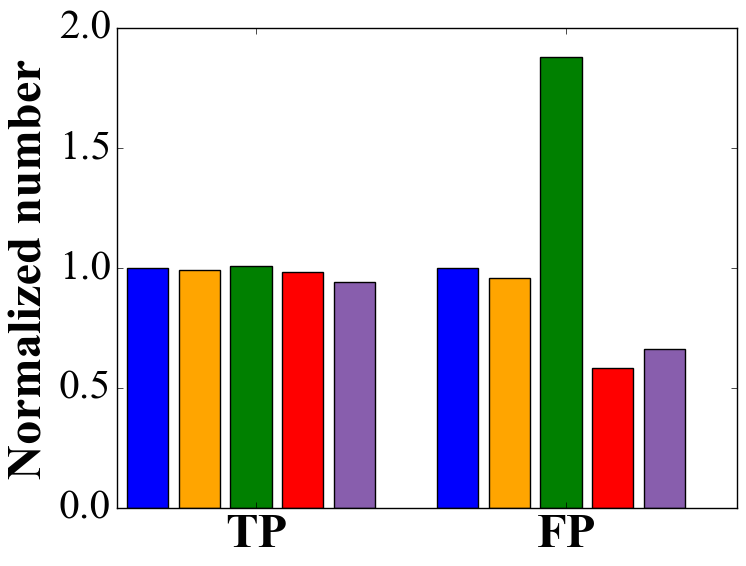}\label{fig:tpfp_all_lizard}}
    \subfloat[][monkey]{\includegraphics[width=.33\linewidth]{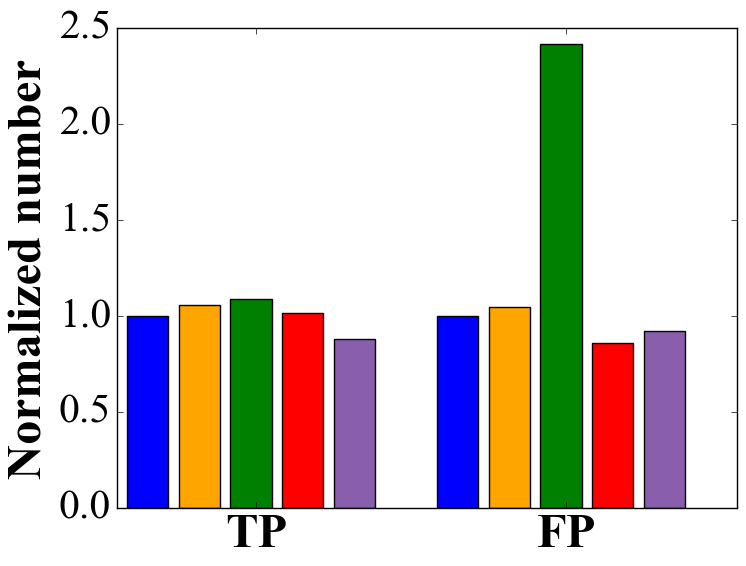}\label{fig:tpfp_all_monkey}}\\
    \subfloat[][motorcycle]{\includegraphics[width=.33\linewidth]{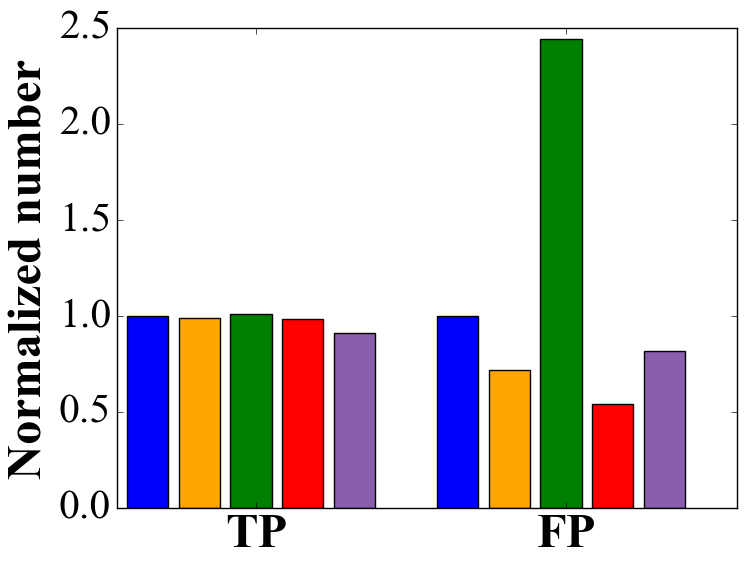}\label{fig:tpfp_all_motorcycle}}
    \subfloat[][rabbit]{\includegraphics[width=.33\linewidth]{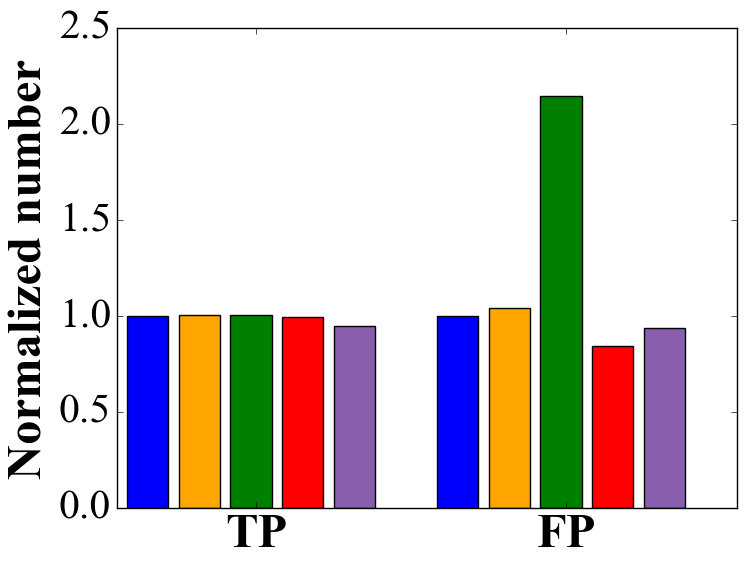}\label{fig:tpfp_all_rabbit}}
    \subfloat[][red\_panda]{\includegraphics[width=.33\linewidth]{tpfp_red_panda.png}\label{fig:tpfp_all_red_panda}}\\
\end{figure*}

\begin{figure*}[h]\ContinuedFloat
\renewcommand*\thesubfigure{\arabic{subfigure}}
    \subfloat[][sheep]{\includegraphics[width=.33\linewidth]{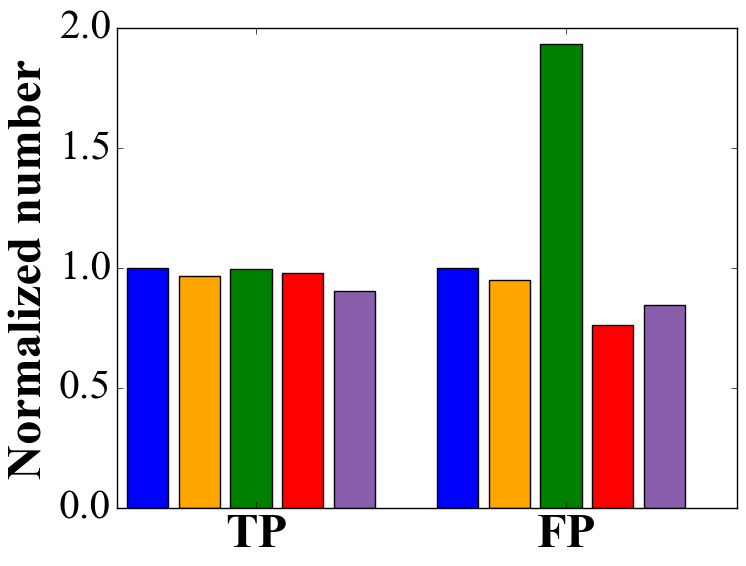}\label{fig:tpfp_all_sheep}}
    \subfloat[][snake]{\includegraphics[width=.33\linewidth]{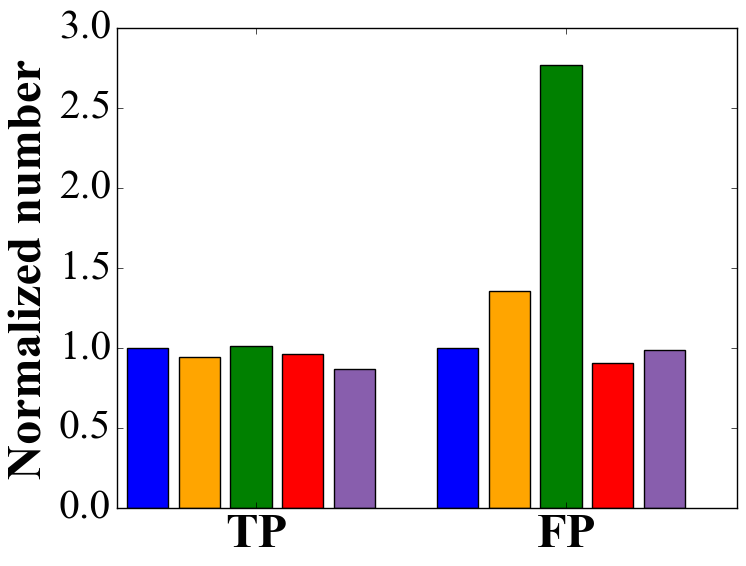}\label{fig:tpfp_all_snake}}
    \subfloat[][squirrel]{\includegraphics[width=.33\linewidth]{tpfp_squirrel.png}\label{fig:tpfp_all_squirrel}}\\
    \subfloat[][tiger]{\includegraphics[width=.33\linewidth]{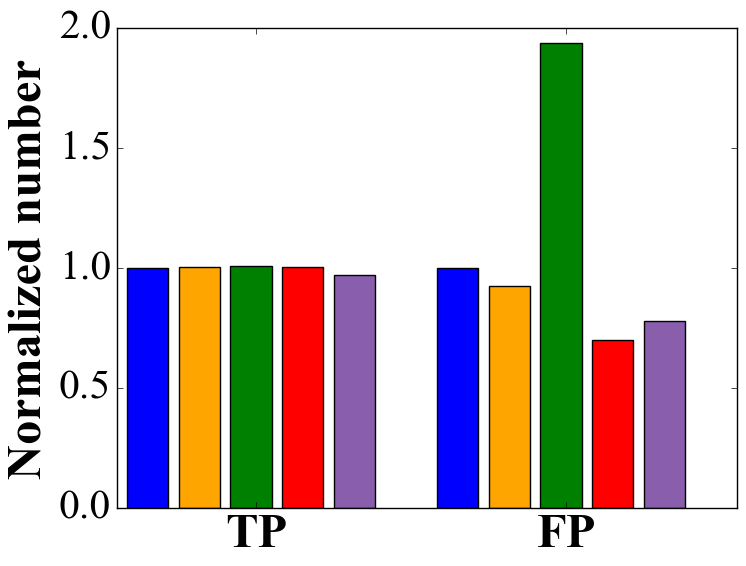}\label{fig:tpfp_all_tiger}}
    \subfloat[][train]{\includegraphics[width=.33\linewidth]{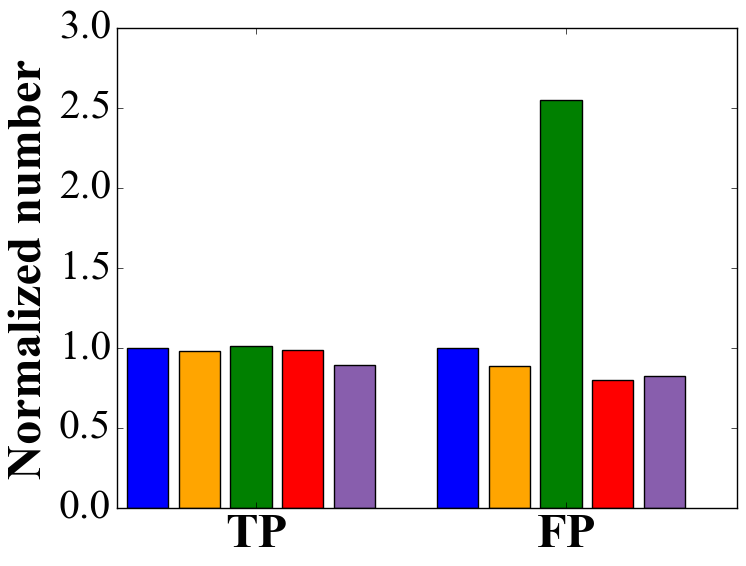}\label{fig:tpfp_all_train}}
    \subfloat[][turtle]{\includegraphics[width=.33\linewidth]{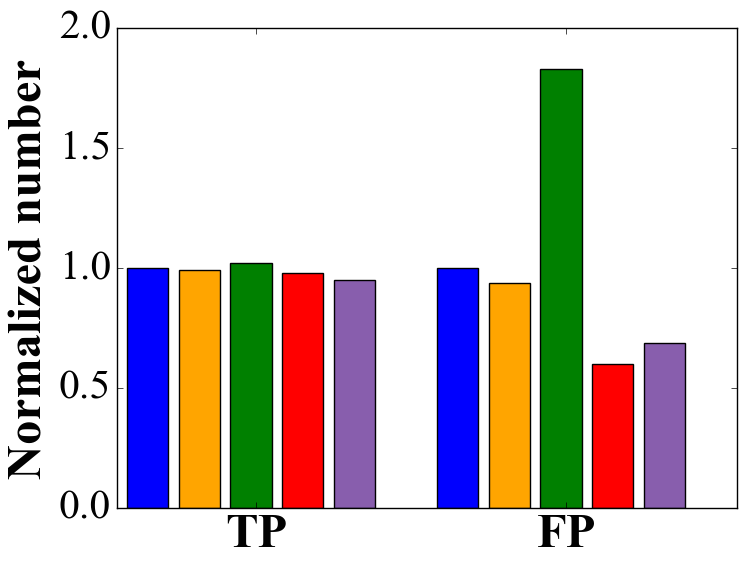}\label{fig:tpfp_all_turtle}}\\
    \subfloat[][watercraft]{\includegraphics[width=.33\linewidth]{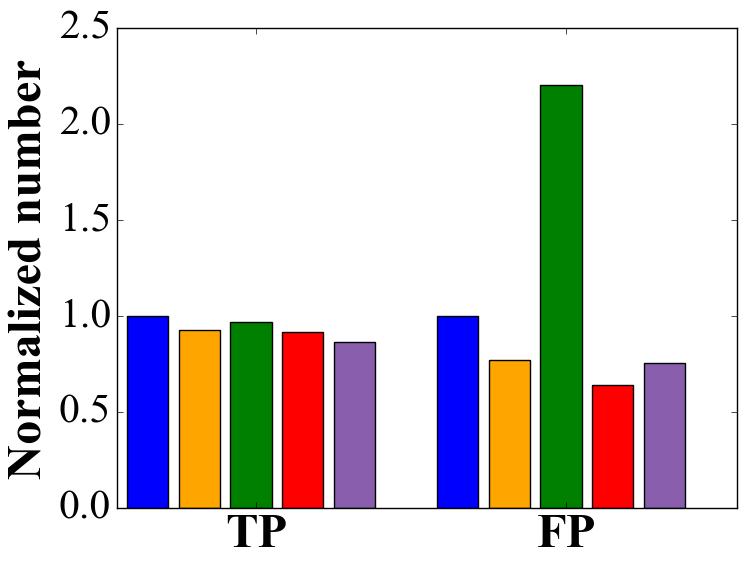}\label{fig:tpfp_all_watercraft}}
    \subfloat[][whale]{\includegraphics[width=.33\linewidth]{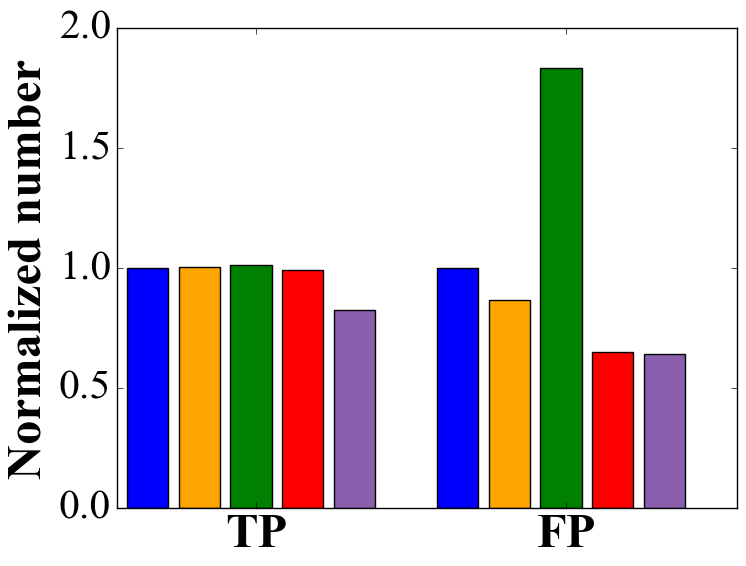}\label{fig:tpfp_all_whale}}
    \subfloat[][zebra]{\includegraphics[width=.33\linewidth]{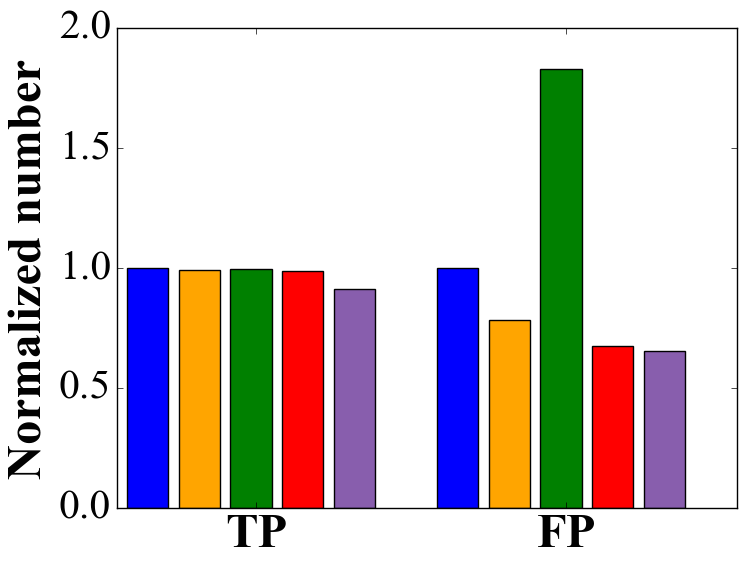}\label{fig:tpfp_all_zebra}}\\
    \begin{center}
    %\vspace{-12pt}
    \includegraphics[width=0.7\linewidth]{tpfp.png}
    \vspace{-6pt}
    \end{center}
    \caption{Normalized true positives and false positives for different methods across all the images in ImageNet VID validation set for 30 categories.}
    \label{fig:tpfp_all}
\end{figure*}

\end{document}